\pdfoutput=1

\documentclass[11pt]{article}

\usepackage[preprint]{acl}

\usepackage{times}
\usepackage{latexsym}

\usepackage[T1]{fontenc}

\usepackage[utf8]{inputenc}

\usepackage{microtype}

\usepackage{inconsolata}
\usepackage{graphicx}
\usepackage{xspace}
\usepackage{amsmath,amsfonts,amssymb}
\usepackage[capitalise]{cleveref}
\usepackage{booktabs}
\usepackage{enumitem}
\usepackage{xcolor}
\usepackage[many]{tcolorbox}
\usepackage{siunitx}
\usepackage{marvosym}
\usepackage{makecell}
\usepackage{multirow}

\usepackage{amsmath,amsfonts,bm}

\def\eqref#1{equation~\ref{#1}}

\def\1{\bm{1}}

\def\vp{{\bm{p}}}

\def\vt{{\bm{t}}}

\def\vx{{\bm{x}}}
\def\vy{{\bm{y}}}

\DeclareMathAlphabet{\mathsfit}{\encodingdefault}{\sfdefault}{m}{sl}
\SetMathAlphabet{\mathsfit}{bold}{\encodingdefault}{\sfdefault}{bx}{n}

\def\gY{{\mathcal{Y}}}

\newcommand{\nl}{NL\xspace}
\definecolor{mygreen}{rgb}{0.82352941,0.90588235,0.72941176}
\definecolor{myblue}{rgb}{0.68235294,0.77254902,0.98039216}
\definecolor{myred}{rgb}{0.90980392,0.65490196,0.74901961}
\definecolor{myyellow}{rgb}{0.98039216,0.89803922,0.68627451}

\newtcolorbox{boxA}{
    fontupper = \bf,
    boxrule = 1.5pt,
    colframe = black 
}

\tcbset{
    sharp corners,
    colback = white,
    before skip = 0.2cm,    
    after skip = 0.5cm      
}

\title{Beyond Natural Language: LLMs Leveraging Alternative Formats for Enhanced Reasoning and Communication}

\author{Weize Chen$^1$\thanks{\ \ Equal Contribution}, Chenfei Yuan$^1$\footnotemark[1], Jiarui Yuan$^1$\footnotemark[1], Yusheng Su$^1$, Chen Qian$^1$, \\\textbf{Cheng Yang}$^3$\thanks{\ \ Corresponding author.}, \textbf{Ruobing Xie}$^2$, \textbf{Zhiyuan Liu}$^1$\footnotemark[2], \textbf{Maosong Sun}$^1$\\
   $^1$ Tsinghua University\\ 
   $^2$ Tencent\\
   $^3$ Beijing University of Posts and Telecommunications\\
  \texttt{\{chenwz21,yuancf21,yuanjr22\}@mails.tsinghua.edu.cn}
  }

\begin{document}
\maketitle
\begin{abstract}
Natural language (NL) has long been the predominant format for human cognition and communication, and by extension, has been similarly pivotal in the development and application of Large Language Models (LLMs). Yet, besides NL, LLMs have seen various non-NL formats during pre-training, such as code and logical expression. NL's status as the optimal format for LLMs, particularly in single-LLM reasoning and multi-agent communication, has not been thoroughly examined.
In this work, we challenge the default use of NL by exploring the utility of non-NL formats in these contexts. We show that allowing LLMs to autonomously select the most suitable format before reasoning or communicating leads to a 3.3 to 5.7\% improvement in reasoning efficiency for different LLMs, and up to a 72.7\% reduction in token usage in multi-agent communication, all while maintaining communicative effectiveness. 
Our comprehensive analysis further reveals that LLMs can devise a format from limited task instructions and that the devised format is effectively transferable across different LLMs. Intriguingly, the structured communication format decided by LLMs exhibits notable parallels with established agent communication languages, suggesting a natural evolution towards efficient, structured communication in agent communication. 
Our code is released at \url{https://github.com/thunlp/AutoForm}.

\end{abstract}

\section{Introduction}
\label{sec:introduction}

\begin{figure}[t!]
    \centering
    \includegraphics[width=0.9\linewidth]{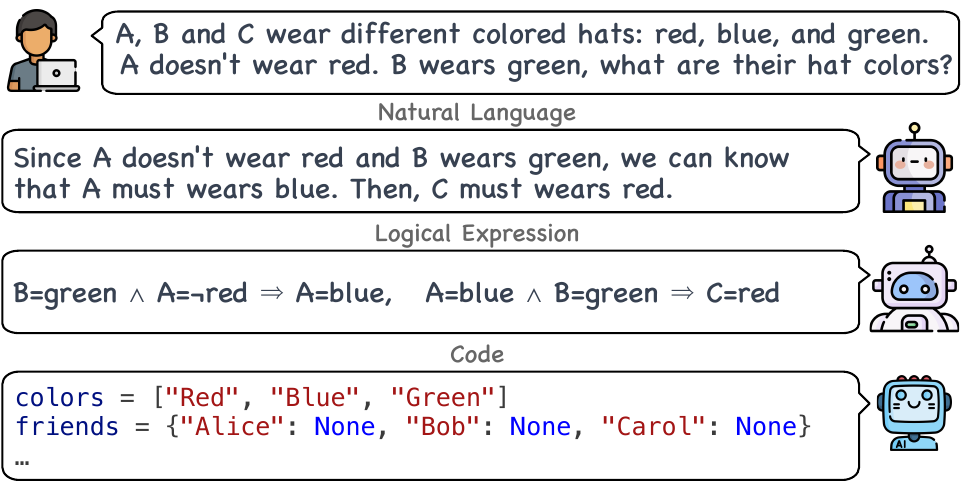}
    \vspace{-0.6em}
    \caption{LLMs may leverage non-NL thought format.}
    \label{fig:intro}
    \vspace{-1.4em}
\end{figure}

Natural language (NL) has long been recognized as a fundamental format for human thought expression and communication, underscored by its pivotal role in the cognitive processes and information exchange of humans~\citep{chomsky2006language,lakoff2008women,whorf2012language}. However, the human mind's capabilities often extend beyond the scope of NL, as suggested by the concept of \textit{mentalese}, a mental language posited by linguists~\citep{fodor1975language,pinker2003language}. Recent advancements in LLMs~\citep{DBLP:journals/corr/abs-2303-08774,team2023gemini,anthropic2023claude} have been remarkable, leading to their increasingly sophisticated application in language agents~\citep{DBLP:conf/iclr/YaoZYDSN023,DBLP:conf/uist/ParkOCMLB23,Significant_Gravitas_AutoGPT}. These advancements, while impressive, predominantly utilize NL for both single-LLM reasoning via Chain-of-Thought (CoT)~\citep{DBLP:conf/nips/Wei0SBIXCLZ22,DBLP:conf/nips/KojimaGRMI22} and multi-agent communication~\citep{DBLP:journals/corr/abs-2308-08155,DBLP:conf/uist/ParkOCMLB23}. Given the human mind's proficiency in transcending NL, critical inquiries emerge: Is NL the optimal format for LLMs in reasoning and inter-agent communication? If not, how should we determine the most suitable format for these applications (\cref{fig:intro})?



Recent research challenges the notion that NL is the ideal intermediate format for LLM reasoning and multi-agent communication. Emerging variants of CoT, such as Program-of-Thought \citep{DBLP:journals/corr/abs-2211-12588,DBLP:conf/icml/GaoMZ00YCN23} and X-of-Thought \citep{DBLP:conf/emnlp/LiuG0HZQZ23} have explored the use of alternative formats like code and mathematical equations, expanding the LLMs' reasoning capabilities. However, these approaches often integrate external tools, where the alternative formats primarily serve as a means to facilitate tool execution (e.g., prompting LLM to generate code and use code interpreter execution result as the answer). This introduces complexity in discerning whether the performance improvements are attributable to the format itself or the accompanying tools. Additionally, while the natural ambiguities and emotions inherent in NL may align well with the nuances of human communication, these may not be desired in agent communication, where precision is more important. Nonetheless, current multi-agent research predominantly utilizes NL~\citep{DBLP:journals/corr/abs-2303-17760,DBLP:journals/corr/abs-2308-08155,DBLP:journals/corr/abs-2308-10848}, with limited exploration of other potentially more accurate and efficient communication formats.

In this study, we implement a straightforward and effective mechanism that prompts the model to favor non-NL formats for single-LLM reasoning and multi-agent communication tasks. By adding an instruction to the original CoT prompt that directs LLMs to explore a non-NL format appropriate for the current input, we showcase the LLMs' capacity for autonomous format decisions. We observe that the LLMs can leverage many non-NL formats such as ordered lists, logical expressions, and markdown tables to reason better. Also, we observe that agents can use more structured language as their communication language to enhance the efficiency of multi-agent collaboration. In particular, our analyses across various single-LLM reasoning tasks demonstrate an average improvement in performance by 3.3-5.7\%. For multi-agent communication, we observe a reduction in token usage by up to 72.7\% without sacrificing effectiveness. These results highlight the considerable potential of non-NL formats in amplifying the reasoning capabilities and communicative efficiency of LLMs.


Our investigation further extends to a comprehensive analysis revealing that LLMs can devise a suitable format from a set of task-specific examples. Using the fixed devised format for the whole task also leads to better answers. Moreover, we show that the formats devised by one LLM are transferable to another LLM. And for the multi-agent communication format, we find that the format adopted by LLMs mirrors those of traditional Agent Communication Languages (ACLs) like KQML~\citep{DBLP:conf/cikm/FininFMM94}, highlighting their clarity, brevity, and structured format for efficient exchanges. Empirically, our approach significantly reduces token usage compared to both ACLs and NL without sacrificing performance. Our work underscores the efficacy of non-NL formats in advancing LLM reasoning and communication.


\section{Related Work}
\label{sec:related-work}
\textbf{LLM Reasoning.}
LLMs have exhibited impressive reasoning performance, especially when employing Chain-of-Thought (CoT) technique~\citep{DBLP:conf/nips/Wei0SBIXCLZ22,DBLP:conf/nips/KojimaGRMI22}. CoT requires LLMs to articulate their reasoning process step-by-step before arriving at a final answer. Building on the CoT framework, variants have been proposed. Program-of-Thought (PoT)~\citep{DBLP:journals/corr/abs-2211-12588,DBLP:conf/icml/GaoMZ00YCN23} prompts models to generate code as thought, and offloads the answer generation to a code interpreter. X-of-Thought~\citep{DBLP:conf/emnlp/LiuG0HZQZ23} integrates CoT, PoT and Equation-of-Thought, dynamically ensembling these methods for improved reasoning. Tree-of-Thought~\citep{DBLP:journals/corr/abs-2305-10601} employs depth and breadth-first search techniques to produce high-quality reasoning chains. 
While some CoT variants explore formats beyond \nl for reasoning, the chosen formats' improvements are obscured by the concurrent use of supplementary tools such as code interpreters, blurring the distinction between format efficacy and tool execution.
We focus on the format itself, investigating whether alternative formats to NL improve the CoT performance.

\paragraph{Multi-Agent Problem Solving.}
Advances in Large Language Models (LLMs) have led to the development of autonomous agents like AutoGPT~\citep{Significant_Gravitas_AutoGPT} and OpenAI Assistant~\citep{openai2023assistant}, demonstrating success in diverse tasks~\citep{DBLP:journals/corr/abs-2303-11366,DBLP:journals/corr/abs-2311-12983,DBLP:journals/corr/abs-2307-13854,DBLP:journals/corr/abs-2304-05332}. Recent research extends this to multi-agent systems for collaborative problem-solving~\citep{DBLP:journals/corr/abs-2305-14325,Osika_gpt-engineer_2023,DBLP:journals/corr/abs-2308-00352,DBLP:journals/corr/abs-2307-07924}. CAMEL~\citep{DBLP:journals/corr/abs-2303-17760} explores collaborative problem-solving between two agents through role-playing. ChatEval~\citep{DBLP:journals/corr/abs-2308-07201} and PRD~\citep{DBLP:journals/corr/abs-2307-02762} assess model responses using multi-agent debates. AgentVerse~\citep{DBLP:journals/corr/abs-2308-10848} introduces a comprehensive framework for multi-agent collaboration, highlighting emergent inter-agent behaviors. However, the alternative formats of multi-agent communication remains underexplored and \nl is directly adopted across various research. \citet{DBLP:journals/corr/abs-2310-06272} explores agent communication with hidden states, but is limited to agents with the same open-source LLM. Our work analyze communication formats for both homogeneous and heterogeneous LLMs.
\section{Method}
\label{sec:method}

\begin{figure*}[t!]
    \centering
    \includegraphics[width=0.85\linewidth]{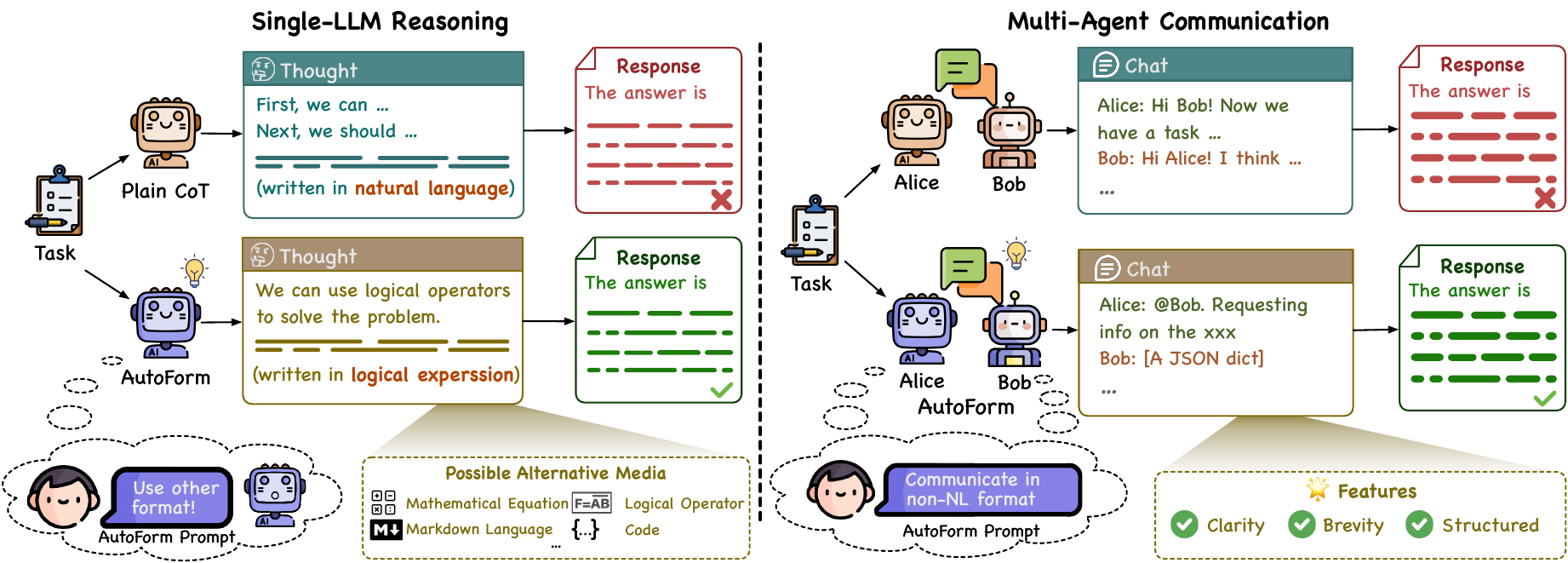}
    \vspace{-0.8em}
    \caption{Overview of single-LLM reasoning and multi-agent communication using plain CoT versus the CoT with AutoForm. The left side depicts the shift from natural language to alternative formats in single-LLM reasoning, while the right side illustrates the enhanced efficiency in multi-agent communication.}
    \label{fig:overview}
\end{figure*}

\subsection{Problem Formulation}
\label{sec:method-preliminaries}
Consider an LLM parameterized by $\theta$, denoted as $p_\theta$. In response to a task description $\vx$ and a prompt $\vp$, CoT prompting initially guides the model to generate thought $\vt=\{t_i\}$ utilizing a thought format $m_t$. While this format is often unspecified and defaults to natural language, alternative formats are feasible. The LLM then formulates an answer $\vy=\{y_j\}$ based on $\vt$. This process is mathematically expressed as sampling from the conditional probability distribution:
\begin{equation}
\small
\begin{aligned}
    p_\theta(\vy, \vt| \vp, \vx, m_t)&=p_\theta(\vy|\vt, \vp, \vx, m_t)p_\theta(\vt| \vp, \vx, m_t),\\
    p_\theta(\vt| \vp, \vx, m_t) &= \prod_{i} p_\theta\left(t_i| \vt_{<i},\vp, \vx, m_t\right),\\
    p_\theta(\vy|\vt, \vp, \vx, m_t)&=\prod_{j} p_\theta\left(y_i| \vy_{<i},\vt, \vp, \vx, m_t\right).
    \label{eq:cot-formula}
\end{aligned}
\end{equation}


In multi-agent scenarios, we extend this formulation to encompass communication among multiple LLMs, each characterized by $\theta_k$. We consider a set of LLMs $\{p_{\theta_k}\}$ collaborating on a task. Communication among these agents utilizes format $m_c$, which can be NL or other alternative formats. This communication is formulated as:
\begin{equation}
\small
\begin{aligned}
    p_{\theta_k}(\vy_k, \vt | \gY, \vp, \vx, m_c, m_t) &=\\p_{\theta_k}(\vy_k|\vt, \gY, \vp, \vx, m_c)&\cdot p_{\theta_k}(\vt| \gY, \vp, \vx, m_t),
    \label{eq:comm-formula}
\end{aligned}
\end{equation}
here we slightly abuse the notation and use $\vy_k$ to denote the response generated by agent $k$, and $\gY$ denote the communication history.

\subsection{Format Choosing for LLMs}
\label{sec:method-medium-choosing}

Building upon the framework delineated in~\cref{sec:method-preliminaries}, our work investigates the effectiveness of allowing the LLMs to decide the thought and communication format before actually starting reasoning or communicating. At the heart of our method is the exploration of alternative formats beyond natural language. We hypothesize that various formats, such as structured data formats (e.g., JSON, markdown tables, lists) or symbolic representations (e.g., logical expressions, mathematical equations), can potentially yield more precise and effective reasoning and streamline communication.

We employ a simple yet effective prompting mechanism, where the LLMs are prompted to select and utilize the format most conducive to the task at hand, which we term as \textbf{AutoForm} (\underline{Auto}nomously-Decided \underline{Form}at). The overview of AutoForm is illustrated at~\cref{fig:overview}. Specifically, for single-LLM reasoning, we add an instruction encouraging the use of non-NL formats to the original CoT prompt. In multi-agent scenarios, a similar instruction for format decision is also added. The specifics of these prompts are detailed in~\cref{sec:appendix-prompts}. In this way, the LLMs implicitly determine and use the thought format $m_t^*=p_\theta(\vx, \vp_t)$ for single-LLM reasoning and the communication format $m_c^*=p_\theta(\vx, \vp_c)$ for multi-agent communication, where $\vp_t$ and $\vp_c$ include instructions for format decision.

\section{Experiments}
\label{sec:experiments}

\begin{table*}[t]
    \centering
    \resizebox{0.8\textwidth}{!}{
    \begin{tabular}{lcccccccc}
        \toprule
        \textbf{Model} & \textbf{Logic Grid} & \textbf{Coin Flip} & \textbf{Info Essen} & \textbf{MM QA} & \textbf{AQuA} 
        & \textbf{Average} \\
        \midrule
        GPT-3.5 CoT     & 46.7$_{\pm 1.6}$ & 23.1$_{\pm 1.0}$ & 32.3$_{\pm 3.2}$ & 24.9$_{\pm 0.8}$ & 60.9$_{\pm 1.2}$ 
        & 41.1$_{\pm 1.8}$ \\
         \ \ \textit{+AutoForm} & \textbf{48.0$_{\pm 3.9}$} & \textbf{39.4$_{\pm 1.1}$} & \textbf{36.7$_{\pm 3.2}$} & \textbf{26.8$_{\pm 0.6}$} & \textbf{63.7$_{\pm 0.7}$} 
         & \textbf{46.0}$_{\pm 2.3}$ \\
        \midrule
        Gemini Pro CoT          & 49.7$_{\pm 0.2}$ & 47.5$_{\pm 0.2}$ & 34.3$_{\pm 0.7}$ & 28.1$_{\pm 0.7}$ & 56.3$_{\pm 0.6}$ 
        & 43.2$_{\pm 0.5}$ \\
         \ \ \textit{+AutoForm}  & \textbf{51.2$_{\pm 0.8}$} & \textbf{57.6$_{\pm 0.7}$}  & \textbf{39.2$_{\pm 1.8}$} & \textbf{31.3$_{\pm 1.1}$} & \textbf{60.0$_{\pm 0.4}$} 
         & \textbf{47.9}$_{\pm 1.1}$ \\
        \midrule
        GPT-4 CoT       & 61.8$_{\pm 1.6}$ & 93.4$_{\pm 1.0}$ & 78.4$_{\pm 2.5}$ & 38.4$_{\pm 1.1}$ & 79.1$_{\pm 0.3}$ 
        & 71.8$_{\pm 1.5}$ \\
         \ \ \textit{+AutoForm}   & \textbf{65.8$_{\pm 2.2}$} & \textbf{98.4$_{\pm 0.2}$} & \textbf{76.9$_{\pm 2.5}$} & \textbf{41.7$_{\pm 0.9}$} & \textbf{80.4$_{\pm 0.8}$} 
         & \textbf{74.1}$_{\pm 1.6}$ \\
        \bottomrule
    \end{tabular}
    }
    \vspace{-0.5em}
    \caption{Comparative performance of single LLM reasoning across various datasets. "Information Essentiality" dataset is abbreviated as "Info Essen," and "Minute Mysteries QA" is referred to as "MM QA" for conciseness.}
    \label{tab:single-exp-results}
    \vspace{-1.3em}
\end{table*}

\subsection{Experimental Settings}
\label{sec:experiments-settings}

\textbf{Single-LLM Reasoning.}
In a preliminary experiment, we prompt various LLMs to use formats other than NL for solving reasoning problems. The results, shown in~\cref{tab:single-specfic-format}, reveal significant variability in performance across different formats. For instance, GPT-3.5 exhibited a 66.8\% performance gap between using an ordered list and a multi-level list, highlighting the intrinsic suitability of specific formats for different tasks.

However, in practical applications, selecting optimal formats for each task may be impractical. To address this, we conduct a comprehensively evaluate the impact of automatically chosen thought formats on LLMs' reasoning performance. We select reasoning benchmarks covering different types of reasoning, including logical reasoning (Logic Grid Puzzle~\citep{DBLP:journals/corr/abs-2206-04615} and Information Essentialy~\citep{DBLP:journals/corr/abs-2206-04615}), mathematical reasoning (AQuA~\citep{DBLP:conf/acl/LingYDB17}), causal reasoning (Minute Mysteries QA~\citep{DBLP:journals/corr/abs-2206-04615}) and symbolic reasoning (Coin Flip~\citep{DBLP:conf/nips/Wei0SBIXCLZ22}). 
For all these tasks, we require the LLMs to generate the answer in a particular format, and we extract the answer with a written regular expression. The average accuracy and the standard deviation of each dataset over 3 runs are reported on most of the datasets. For more details on the experimental settings, please refer to~\cref{sec:appendix-exp-settings}.


\begin{table*}[t]
    \centering
    \resizebox{0.9\textwidth}{!}{
    \setlength{\tabcolsep}{3pt}
    \begin{tabular}{l*{3}{crr}}
        \toprule
        & \multicolumn{3}{c}{\textbf{Wiki Hop}} & \multicolumn{3}{c}{\textbf{Hotpot QA}} & \multicolumn{3}{c}{\textbf{Narrative QA}}\\
        \cmidrule(lr){2-4}\cmidrule(lr){5-7}\cmidrule(lr){8-10}
        \textbf{Model} & \textbf{RougeL} &\textbf{\# Tokens} & $\bm{\Delta}$\textbf{Tokens} & \textbf{RougeL} & \textbf{\# Tokens}  & $\bm{\Delta}$\textbf{Tokens} & \textbf{RougeL}& \textbf{\# Tokens}  & $\bm{\Delta}$\textbf{Tokens} \\
        \midrule
        GPT-4 + GPT-3.5 & \textbf{0.53} & 281.5 & - & 0.63 & 345.5 & - & 0.43 & 178.3 & -\\ 
        \ \ \textit{+AutoForm} & \textbf{0.53} & \textbf{255.0} & -9.4\% & \textbf{0.70} & \textbf{94.3} & -72.7\%  & \textbf{0.48} & \textbf{119.4} & -33.0\%\\
        \midrule
        GPT-4 + GPT-4 & 0.50 & 237.5 & - & 0.69 & 145.2 & -  & \textbf{0.43} & 240.7 & - \\
         \ \ \textit{+AutoForm} & \textbf{0.52} & \textbf{146.2} & -38.4\% & \textbf{0.76} & \textbf{115.0} & -20.8\%  & \textbf{0.43} & \textbf{141.7} & -41.1\% \\
        \bottomrule
    \end{tabular}
    }
    \vspace{-0.5em}
    \caption{Comparative performance in multi-agent communication across various QA datasets. The table highlights RougeL scores, with better performance in different model pairing settings indicated in bold. The $\bm{\Delta}$Tokens column quantifies the token reduction achieved by the AutoForm method.}
    \label{tab:multi-exp-results}
    \vspace{-1.3em}
\end{table*}

\textbf{Multi-Agent Communication.}
To measure whether alternative formats can streamline communication, we consider scenarios where two agents with different knowledge or contexts are tasked with answering a question. The answer to the question should be derived from the knowledge of only one of the agents, or both agents' knowledge collectively, therefore requiring information exchange. The two agents speak in turn to discuss and give the final answer. To create such scenarios, we utilize three existing datasets: Hotpot QA~\citep{DBLP:conf/emnlp/Yang0ZBCSM18}, Wiki Hop~\citep{DBLP:journals/tacl/WelblSR18} and Narrative QA~\citep{DBLP:journals/tacl/KociskySBDHMG18}. Hotpot QA and Wiki Hop are two multi-hop QA datasets, which require multiple sentences or paragraphs to deduce the final answer. We randomly assign the text segments provided in the datasets to two different agents. Communication is thus needed to derive the correct answer. We have also explored assigning each agent part of the supporting facts, thus needing more communication to derive the answer. Narrative QA requires the model to read the whole book and answer a question. The length of a book often exceeds the context limit of the LLMs. We split the books into nearly equal sizes for the two agents, and ask them to answer the question. For evaluation, we use RougeL as the primary metric. More details are elaborated in~\cref{sec:appendix-exp-settings}.

To measure whether alternative formats can streamline communication, we consider scenarios where two agents with different knowledge or contexts are tasked with answering a question. The answer to the question should be derived from the knowledge of only one of the agents, or both agents' knowledge collectively, therefore requiring information exchange. The two agents speak in turn to discuss and give the final answer. To create such scenarios, we utilize three existing datasets: Hotpot QA~\citep{DBLP:conf/emnlp/Yang0ZBCSM18}, Wiki Hop~\citep{DBLP:journals/tacl/WelblSR18}, and Narrative QA~\citep{DBLP:journals/tacl/KociskySBDHMG18}. Hotpot QA and Wiki Hop are two multi-hop QA datasets, which require multiple sentences or paragraphs to deduce the final answer. We randomly assign the text segments provided in the datasets to two different agents. Communication is thus needed to derive the correct answer. We also explore assigning each agent part of the supporting facts, thus needing more communication to derive the answer. Narrative QA requires the model to read the whole book and answer a question. The length of a book often exceeds the context limit of the LLMs. We split the books into nearly equal sizes for the two agents and ask them to answer the question. While supporting facts in Narrative QA are not guaranteed to reside in different segments, this division still introduces critical challenges: agents must identify who possesses the necessary information, necessitating communication. For evaluation, we use RougeL as the primary metric. More details are elaborated in~\cref{sec:appendix-exp-settings}.

\textbf{Research Questions.}
To comprehensively explore the potential of LLMs in selecting suitable formats for both single-agent reasoning and multi-agent communication, we conduct an in-depth analysis addressing six pivotal research questions (RQs). These questions aim to unravel the intricacies of format selection by LLMs and its impact on task performance across various scenarios:
\begin{itemize}[noitemsep,topsep=0pt,parsep=0pt,partopsep=0pt,leftmargin=1em]
    \item \textbf{RQ1}: Can LLMs select the suitable formats autonomously?~(\cref{sec:experiments-results})
    \item \textbf{RQ2}: What formats are chosen in single-LLM reasoning?~(\cref{sec:analysis-chosen-media})
    \item \textbf{RQ3}: Can LLMs devise a general format for a task based on some task inputs?~(\cref{sec:analysis-medium-generalization})
    \item \textbf{RQ4}: Is the decided format transferable between different LLMs?~(\cref{sec:analysis-transferability})
    \item \textbf{RQ5}: What are the features of the formats used in multi-agent communication? ~(\cref{sec:analysis-comm-features})
    \item \textbf{RQ6}: Does the autonomously determined multi-agent communication format align with conventional agent communication languages such as KQML~\citep{DBLP:conf/cikm/FininFMM94}?~(\cref{sec:analysis-comm-acl})
\end{itemize}


\subsection{RQ1: The Capability of LLMs on Selecting Suitable Format}
\label{sec:experiments-results}

\textbf{Single-LLM Reasoning.}
The comparative efficacy of the AutoForm approach over the conventional Chain-of-Thought (CoT) methodology in single-LLM reasoning tasks is encapsulated in~\cref{tab:single-exp-results}. We observe clear performance improvements when employing AutoForm across different datasets, compared to the baseline CoT method. 

For GPT-3.5, the implementation of AutoForm leads to a significant improvement in accuracy, particularly notable in the Coin Flip dataset, where accuracy escalates from 22.2\% to 38.0\%. This substantial increase highlights the model's enhanced ability in symbolic reasoning. Across other datasets, AutoForm yields consistent enhancements, with increases generally ranging between 3\% to 5\%, culminating in an overall average performance boost of 5.4\%. Similarly, for Gemini Pro, AutoForm achieves an average performance enhancement of 5.7\%. GPT-4 also benefits from AutoForm, with an average performance uplift of 3.3\% across all tasks. These consistent improvements across various datasets demonstrate the method's model-agnostic robustness and the efficacy of utilizing alternative formats in reasoning tasks. It also suggests that alternative formats, apart from NL, can aid LLMs in task resolution. It is just that without explicit reminders, the LLMs do not explore alternative formats. For a detailed illustration of how LLM responses are influenced by AutoForm, refer to~\cref{sec:appendix-examples}, which presents selected examples from each LLM for every task.

\textbf{Multi-Agent Communication.}
The outcomes of our multi-agent communication experiments, detailed in~\cref{tab:multi-exp-results}, provide valuable insights into the efficiency and effectiveness of utilizing alternative communication formats in collaborative environments. In this experiment, we experiment with different model pairings to explore this robustness more comprehensively. Since the initial speaking agent often sets the tone for the communication format, we vary the speaking order in pairings of heterogeneous models,  e.g., \textit{GPT-4 + GPT-3.5} in the table indicates GPT-4 is the initiator. Due to page limit, we place results where GPT-3.5 initiates the conversation, and where the supporting facts are split and distributed to two agents at~\cref{sec:appendix-additional-results-multi}.

A key finding from our experiments is the notable efficiency achieved through AutoForm, as evidenced by the substantial reduction in token usage across diverse model pairings and speaking orders. This efficiency, quantified in the $\bm{\Delta}$Tokens column, illustrates the capability of LLMs to move beyond their typical NL to adopt more concise and efficient communication formats. This is particularly evident in the Hotpot QA dataset with the GPT-4 and GPT-3.5 pairing, where we witness a token reduction reaching 72.7\%. On most of the other datasets, AutoForm also obtains substantial token reduction. These findings imply that LLMs, though extensively trained on NL, are capable of exploring and employing alternative formats to enhance communication efficiency. Detailed case studies further elucidating the features of the chosen formats will be presented in~\cref{sec:analysis-comm-features}, and more cases are presented at~\cref{sec:appendix-examples}.

Furthermore, the effectiveness of the multi-agent communication facilitated by AutoForm, as gauged by RougeL scores, is found to be largely comparable to, and occasionally exceeding, that of natural language-based interactions. This is especially true when GPT-4 initiates the conversation, suggesting that more advanced LLMs possess a better ability to select communication formats that strike a balance between conciseness and clarity. Conversely, we show in~\cref{sec:appendix-additional-results-multi} that with GPT-3.5 as the initiator, despite the similar notable efficiency in token usage, the performance is generally akin to or slightly below that achieved with natural language. These observations highlight the intricate challenge of selecting an appropriate communication format, a task that proves demanding even for sophisticated LLMs, and the importance of balancing brevity with the need to preserve the integrity of the communicative content.

\begin{figure*}[t]
    \centering
    \includegraphics[width=0.9\textwidth]{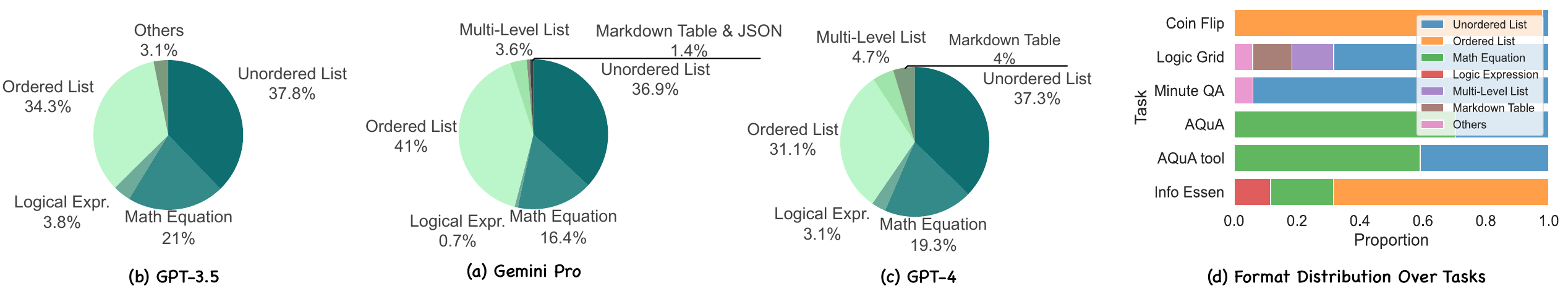}
    \vspace{-0.8em}
    \caption{Format distribution chosen by Gemini Pro (a), GPT-3.5 (b) and GPT-4 (c), and the overall format distribution across tasks from both models (d).}
    \label{fig:media-dist}
\end{figure*}

\subsection{RQ2: Formats Chosen in Single-LLM Reasoning}
\label{sec:analysis-chosen-media}
In addressing RQ1, we investigate the formats selected from LLMs when tasked with reasoning. This analysis is pivotal in understanding how LLMs, when granted the autonomy to choose, navigate away from the default NL format to potentially more efficient alternatives.


We analyze a randomly sampled set of 50 examples from each dataset, investigate the reasoning traces produced with AutoForm, and manually count the number of appeared formats. \cref{fig:media-dist}(a-c) display the distribution of formats chosen by Gemini Pro, GPT-3.5 and GPT-4, and \cref{fig:media-dist}(d) displays their combined preferences across various tasks. The data indicates a notable diversity in format selection by LLMs. A shift towards structured formats, such as lists, logical expressions, and markdown tables is observed. These formats are particularly favored in tasks that demand logical reasoning, offering clearer and more concise data representation, as illustrated in~\cref{fig:media-dist}(d).

\begin{table*}[t]
    \centering
    \vspace{-0.8em}
    \resizebox{0.9\textwidth}{!}{
    \begin{tabular}{lcccccr@{}l}
        \toprule
        \textbf{Model} & \textbf{Logic Grid} & \textbf{Coin Flip} & \textbf{Info Essen} & \textbf{MM QA} & \textbf{AQuA} & \multicolumn{2}{c}{\textbf{Average}} \\
        \midrule
        GPT-3.5 (Instance-Based) & 48.0 & 49.2 & 39.7 & 26.1 & \textbf{66.5} & 45.9 & $_{(+7.5\%)}$ \\
        GPT-3.5 (Task-Based)   & \textbf{51.0} & \textbf{62.8} & \textbf{42.6} & \textbf{28.1} & 65.0 & \textbf{49.9} & $_{(+11.5\%)}$ \\
        \midrule
        Gemini Pro (Instance-Based)   & 39.5 & 44.8 & \textbf{35.3} & 27.1 & \textbf{59.4} & 41.2 & $_{(+2.5\%)}$ \\
        Gemini Pro (Task-Based)     & \textbf{41.5} & \textbf{47.8} & \textbf{35.3} & \textbf{28.6} & \textbf{59.4} & \textbf{42.5} & $_{(+3.8\%)}$ \\
        \midrule
        GPT-4 (Instance-Based)   & \textbf{71.5} & \textbf{100.0}    & \textbf{76.5} & \textbf{41.4} & 78.3          & \textbf{73.5} & $_{(+3.2\%)}$ \\
        GPT-4 (Task-Based)     & 70.0 & 99.8              & 75.0          & \textbf{41.4} & \textbf{79.5} & 73.1 & $_{(+2.8\%)}$ \\
        \midrule
        GPT-4 $\Rightarrow$ GPT-3.5 (Task-Based) & 47.5 & 83.0 & 35.3 & 23.2 & 59.4 & 49.7 & $_{(+11.2\%)}$ \\
        GPT-4 $\Rightarrow$ Gemini Pro (Task-Based) & 40.0 & 74.6 & 25.0 & 25.6 & 47.2 & 42.5 & $_{(+3.8\%)}$  \\
        GPT-3.5 $\Rightarrow$ GPT-4 (Task-Based) & 65.5 & 98.6 & 72.1 & 42.9 & 79.5 & 71.7 & $_{(+1.3\%)}$ \\
        \bottomrule
    \end{tabular}}
    \vspace{-0.7em}
    \caption{Comparative performance of two-step AutoForm with single LLM reasoning across various datasets. The notation $model_1 \Rightarrow model_2$ denotes using $model_1$ for format selection, and $model_2$ for problem-solving. Average performance improvements over CoT results, as presented in~\cref{tab:single-exp-results}, are denoted with a subscript.}
    \label{tab:two-step-automed}
    \vspace{-1.3em}
\end{table*}

\subsection{RQ3: Generalization of Format Selection Based on Task Inputs}
\label{sec:analysis-medium-generalization}

An examination of~\cref{fig:media-dist}(d) reveals a discernible variation in the LLMs' format preferences across different tasks. This variation aligns with the expectation that the optimal format would naturally differ between tasks, each with its unique requirements. In addressing RQ2, we probe whether LLMs are capable of identifying a general format suitable for a given task based on a subset of inputs, and then consistently applying this format for problem-solving. In AutoForm, as delineated in~\cref{sec:method-medium-choosing}, LLMs typically select a format \textit{implicitly} for each instruction on a case-by-case basis. Nonetheless, it stands to reason that certain tasks may be inherently conducive to a specific format. To investigate this hypothesis, we introduce the \textbf{two-step AutoForm}. This approach tasks an LLM with first determining the most efficient format and subsequently utilizing that format in the CoT problem-solving stage. That is, we turn the implicit format decision into an explicit step, which mathematically follows~\cref{eq:cot-formula} instead of merging the two steps.

The two-step AutoForm experiments with two distinct settings: 1) \textit{Instance-Based}, where the LLM selects a format for each instruction, and 2) \textit{Task-Based}, where the LLM deduces a general format for the entire task by analyzing 5 inputs from the task. Note that, unlike few-shot prompting, the Task-Based setting does not provide answers within the inputs, and these inputs are only utilized during the format decision step. 

The results are detailed in \cref{tab:two-step-automed}. The Task-Based setting demonstrates that both GPT-3.5 and Gemini Pro can effectively generalize a thought format from a limited set of inputs within a task, and often outperforming the Instance-Based setting. In contrast, GPT-4 shows similar performance levels in both Task-Based and Instance-Based settings, suggesting that its advanced capabilities may afford it greater flexibility in format usage. These findings indicate that LLMs, particularly less sophisticated ones like GPT-3.5, may benefit from exposure to multiple inputs from a task to better generalize an effective thought format. This ability of LLMs to generalize the format for a task makes the AutoForm approach more practical since the format can be identified only once for a specific task.


\begin{figure*}[t]
    \vspace{1mm}
    \centering
    \includegraphics[width=0.85\textwidth]{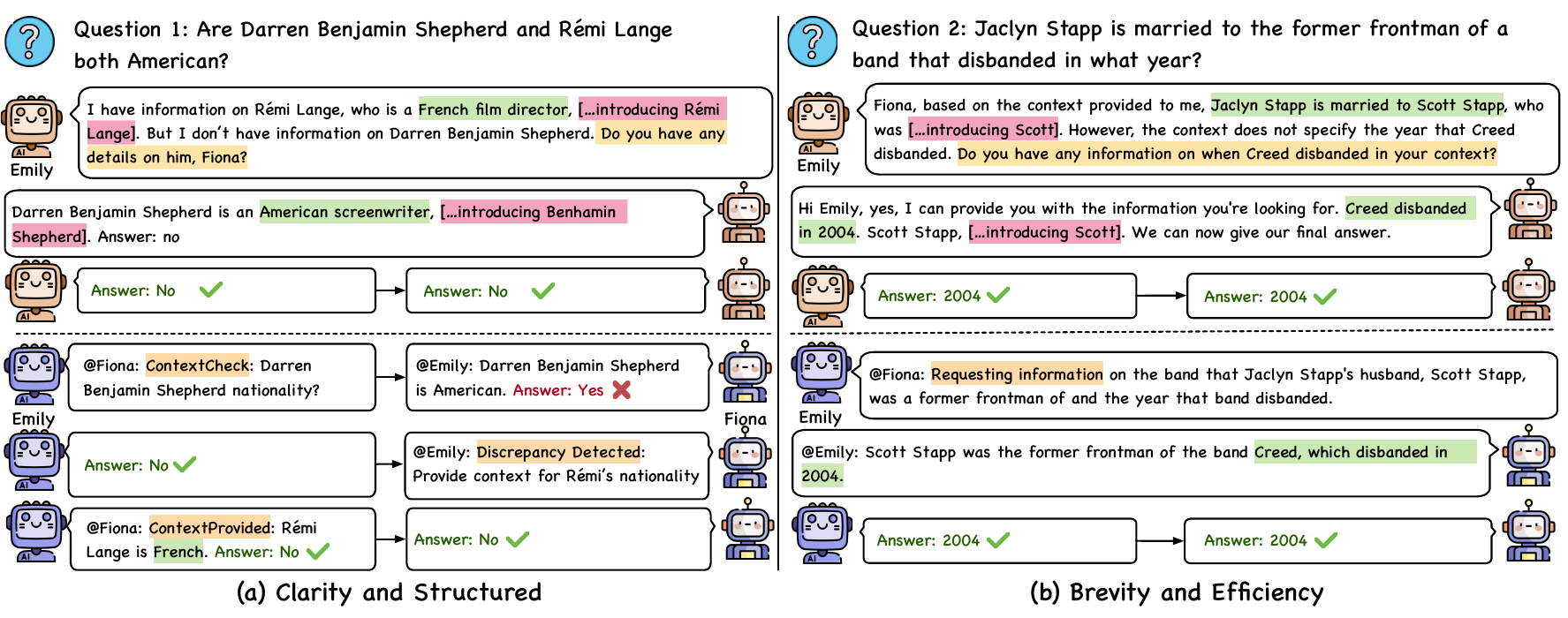}
    \vspace{-0.8em}
    \caption{Multi-agent communication examples. The top panel illustrates a traditional natural language conversation, and the bottom panel shows a conversation using AutoForm. Necessary information related to the question is marked in \colorbox{mygreen}{green}, redundant information is marked in \colorbox{myred}{red}, and speech-act-related phrases are marked in \colorbox{myyellow}{orange}.}
    \label{fig:case}
    \vspace{-1.3em}
\end{figure*}
\subsection{RQ4: Transferability of Format Across Different LLMs}
\label{sec:analysis-transferability}



This subsection delves into RQ3, the transferability of the format across LLMs. The concept of format transferability is crucial in understanding the universality of formats decisions made by LLMs and their applicability across various models. Instead of using the same model for format selection and problem-solving (homogeneous setting) as we have done in~\cref{sec:analysis-medium-generalization}, we explore the heterogeneous setting of the two-step AutoForm, where different models are employed for the two steps.



The results are presented in the last group of~\cref{tab:two-step-automed}. Generally, the format is transferable, but may lead to slightly inferior performances compared to the homogeneous setting. For example, transferring the format decided by GPT-4 to GPT-3.5 or Gemini Pro leads to a decrease compared to the homogeneous setting for the two LLMs in most of the tasks except for Coin Flip. On the other hand, when transferring the format decided by GPT-3.5 to GPT-4, the results are generally comparable to the homogeneous setting for GPT-4. For information essentiality, the format selected by GPT-3.5 is generally less efficient, leading to incomplete problem-solving processes. We give examples in \cref{tab:appendix-info-essen-sep-model-examples-3.5-4,tab:appendix-info-essen-sep-model-examples-4-task-based,tab:appendix-info-essen-sep-model-examples-4-3.5}. For other tasks, the format generated by GPT-3.5 proves adequate and is easily interpreted by GPT-4, resulting in similar performances to the homogeneous setting for GPT-4.

\subsection{RQ5: Features of Communication Format}
\label{sec:analysis-comm-features}
In addressing RQ4, this subsection investigates the characteristics of the formats used by language agents in multi-agent communication scenarios. Our goal is to identify key attributes contributing to efficient communication by examining the formats used during their interactions.

We analyze 50 random interaction logs for each dataset and present some cases in~\cref{fig:case} and~\cref{sec:appendix-examples}. Despite some retained characteristics of NL, the communication formats decided via AutoForm display distinct features:

\textbf{Clarity and Structure.}
An important feature of the selected formats is an emphasis on clarity. LLMs consistently favor formats facilitating unambiguous and straightforward communication, which is vital in our multi-agent scenarios. In these scenarios, agents possess divergent knowledge sets, making the clear exchange of this distinct information indispensable. Structured formats, which provide an organized method of presenting information, are also prevalent. These formats enhance the comprehensibility and accessibility of the content. Contrasting this with the use of NL, we observe that the LLM-decided formats tend to be more direct and clear, effectively reducing redundancy.

\textbf{Brevity and Efficiency.}
Another key feature is the focus on brevity, enhancing communication efficiency. Formats chosen by LLMs often omit elements like pleasantries or emotive expressions, resulting in concise exchanges. This brevity conserves computational resources and concentrates dialogue on the task at hand, optimizing the communication process for faster, more efficient information exchange. This is particularly beneficial in scenarios requiring rapid and effective decision-making.

\subsection{RQ6: Alignment with Conventional Agent Communication Languages}
\label{sec:analysis-comm-acl}
RQ5 probes the extent to which the communication formats determined by LLMs using AutoForm align with traditional Agent Communication Languages (ACLs), such as KQML \citep{DBLP:conf/cikm/FininFMM94} and FIPA-ACL \citep{fipa_acl_messag_struc_specif}. These ACLs have been instrumental in structuring communication between intelligent agents to foster cooperation and coordination. A typical KQML message, as shown below, exemplifies the structured nature of traditional ACLs:

{\small\begin{verbatim}
(ask-one
  :sender joe
  :content (PRICE IBM ?price)
  :receiver stock-server
  :reply-with ibm-stock
  :language LPROLOG
  :ontology NYSE-TICKS)
\end{verbatim}
}

Our examination of the communication patterns emerging from AutoForm reveals an interesting resemblance to these structured elements. As depicted in~\cref{fig:case}, LLMs frequently employ a structured format where "@" denotes the \textit{receiver}, verb phrases such as "ContextCheck" indicate the \textit{performative} (the \textit{"ask-one"} in the above example), and a succinct text string encapsulates the \textit{content}. This structuring mirrors the composition of ACL messages, where each part serves a specific function in the communication process.

Intrigued by this similarity, we conduct an experiment where LLMs are prompted to communicate using a format similar to KQML:

{\small\begin{verbatim}
(performative
  :content ...
  :receiver ...)
\end{verbatim}
}
The results of this experiment on two GPT-4-based agents are presented in~\cref{tab:multi-rusults-conventional-acl}. We have two settings, one prompts the LLMs to use the exact KQML format as presented above, and the other setting uses the JSON version of KQML format, considering that LLMs may be more adept at JSON. While both settings show worse or comparable performance to AutoForm in terms of RougeL, the number of tokens consumed is larger than AutoForm. This finding suggests that while LLMs can indeed emulate the formality of traditional ACL formats, the AutoForm approach optimizes the communication by enhancing clarity and structure, yet concurrently reduces token usage.

\begin{table}[t]
    \centering
    \resizebox{0.95\linewidth}{!}{
    \begin{tabular}{l*{6}{c}}
        \toprule
        & \multicolumn{2}{c}{\textbf{Hotpot QA}} & \multicolumn{2}{c}{\textbf{Wiki Hop}} & \multicolumn{2}{c}{\textbf{Narrative QA}} \\
        \cmidrule(lr){2-3}\cmidrule(lr){4-5}\cmidrule(lr){6-7}
        \textbf{Format} &\textbf{RougeL} & \textbf{\#Tokens} &\textbf{RougeL} & \textbf{\#Tokens} & \textbf{RougeL} & \textbf{\#Tokens}\\
        \midrule
        KQML & \textbf{0.76} &313.8 &\textbf{0.53} &368.1 & 0.28 & 343.3\\
        JSON & 0.71 & 346.0 &\textbf{0.53} & 291.4 &0.22 &385.2\\
        \midrule
        AutoForm & \textbf{0.76} & \textbf{115.0} &0.52 & \textbf{146.2} & \textbf{0.43} & \textbf{141.7}\\
        \bottomrule
    \end{tabular}}
    \vspace{-0.7em}
    \caption{Multi-agent communication performances using conventional ACL format.}
    \label{tab:multi-rusults-conventional-acl}
    \vspace{-1.3em}
\end{table}

These results highlight two key implications. First, AutoForm can generate communication patterns similar to established ACLs. Second, it efficiently distills these traditional formats into a more concise form, conserving computational resources while maintaining communicative effectiveness. This balance of clarity, structure, and brevity makes AutoForm a powerful tool for facilitating intelligent agent communication in various contexts.
\section{Conclusion}
\label{sec:conclusion}



In this work, we demonstrate that LLMs can autonomously determine suitable non-NL formats for reasoning and communication using the AutoForm prompting method. Our analyses address six key research questions, showing that LLMs can generalize a reasoning format from task-specific examples and transfer it across different models. Additionally, the communication formats generated by LLMs resemble traditional ACLs, offering both precision and efficiency. These insights enhance our understanding of LLMs' capabilities beyond NL, improving LLM reasoning and inter-agent communication.
\section*{Limitations}
Despite we have shown many kinds of formats can facilitate LLMs reasoning and communication, the scope of alternative formats explored is still not exhaustive. The potential of numerous other formats and their specific applications to various LLM architectures warrants further investigation.

Moreover, the generalization of chosen formats across tasks, while promising, shows variability in effectiveness depending on the complexity of the task and the specific LLM used. This variability highlights the nuanced nature of format suitability and its impact on task performance, suggesting that further exploration is necessary to fully harness the capabilities of alternative formats.

\bibliography{custom}

\appendix

\clearpage
\section{Experimental Settings}
\label{sec:appendix-exp-settings}
In this section, we introduce the details of our experimental settings. 

\begin{table*}[t]
    \centering
    \resizebox{\textwidth}{!}{
    \begin{tabular}{l r r r r r}
        \toprule
         \textbf{Dataset} & \textbf{\# Examples} & \textbf{Input} & \textbf{Output} & \textbf{Category} & \textbf{License}\\
         \midrule
         \textit{Single-LLM Reasoning}\\
         Logic Grid Puzzle & 200 & Clues + Question & Number & Logical Reasoning & Apache License 2.0\\
         Information Essentiality & 68 & Question + Statement Options & Option Number & Logical Reasoni & Apache License 2.0 \\
         AQuA & 254 & Mathematical Question + Options & Option Number & Mathematical Reasoning & Apache License 2.0 \\
         Minute Mysteries QA & 203 & Story + Question + Options & Option Number & Causal Reasoning & Apache License 2.0 \\
         Coin Flip & 500 & Action Sequence & Yes / No & Symbolic Reasoning & MIT License\\
         \midrule
         \textit{Multi-LLM Communication}\\
         HotPot QA & 100 & Passages + Question & Free Text & Multi-Hop QA & CC BY-SA 4.0\\
         Wiki Hop & 100 & Sentences + Question & Free Text & Multi-Hop QA & CC BY-SA 3.0\\
         Narrative QA & 100 & Book + Question & Free Text & QA & Apache License 2.0\\
         \bottomrule
    \end{tabular}}
    \caption{The datasets we use in our experiments.}
    \label{tab:datasets}
\end{table*}
\subsection{Models}
For OpenAI's models, we use gpt-3.5-turbo-1106 and gpt-4-1106-preview. For Gemini pro, we use the Gemini pro 1.0 in 2024.1.

\subsection{Dataset Pre-Processing}
The statistics of the processed data are presented at~\cref{tab:datasets}. We now elaborate the dataset download and pre-process process.
\paragraph{Single-LLM Reasoning.}
For Logic Grid, Information Essentiality and Minute Mysteries QA that are from Big-Bench, we download the dataset from the official repo\footnote{\url{https://github.com/google/BIG-bench}}. For Coin Flip, we download the dataset from \url{https://huggingface.co/datasets/skrishna/coin_flip}, and use the first 500 examples in the test set. For AQuA, we download the dataset from \url{https://huggingface.co/datasets/aqua_rat/} and use its test set. 

\paragraph{Multi-Agent Communication.}
For the Hotpot QA dataset, we adhere to the methodology outlined by Reflexion~\citep{DBLP:journals/corr/abs-2303-11366}, obtaining the dataset from their repository~\footnote{\url{https://github.com/noahshinn/reflexion/tree/main/hotpotqa_runs/data}}. In the case of Wiki Hop, we acquire it through the Huggingface Datasets platform, from which we randomly selected 100 examples from its validation set for our study. For the Narrative QA dataset, also sourced from Huggingface Datasets, we note inconsistencies in the quality of the e-books included. To ensure higher data quality, we exclusively utilize e-books from Project Gutenberg by checking whether the e-book starts with "Project Gutenberg's". Furthermore, considering the context length limitation of 16k tokens in GPT-3.5, we exclude e-books exceeding 30k tokens. This exclusion is to enable splitting the content into two segments, each fitting within the GPT-3.5 context limit. From this refined dataset, we randomly chose 100 examples for analysis.

\subsection{Metrics}
For single-LLM reasoning, we report the accuracy by comparing the generated answer with the label. For multi-agent communication, we report the RougeL score of the generated answer. The RougeL is calculated using Google's implementation\footnote{\url{https://github.com/google-research/google-research/tree/master/rouge}}.

\begin{table*}[t]
    \centering
    \resizebox{\textwidth}{!}{
    \begin{tabular}{lcccccccc}
        \toprule
        \textbf{Dataset} & \textbf{Model} & \textbf{\makecell[c]{Math\\Equation}} & \textbf{\makecell[c]{Unordered \\ List}} & \textbf{\makecell[c]{Ordered \\ List}} & \textbf{\makecell[c]{Markdown \\ Table}} & \textbf{\makecell[c]{Multi-level \\ List}} & \textbf{\makecell[c]{Logical \\ Expression}} \\
        \midrule
        \multirow{3}{*}{\makecell[c]{Logic\\Grid}} & GPT-3.5 & \textbf{50.0} & 47.0 & 46.0 & 48.5 & 45.0 & 41.0 \\
         & GPT-4 & 58.0 & \textbf{65.0} & 59.0 & 55.0 & 62.0 & 59.0 \\
         & Gemini Pro & 42.0 & \textbf{50.5} & 48.0 & 49.0 & 48.0 & 49.0 \\
        \midrule
        \multirow{3}{*}{\makecell[c]{Coin\\Flip}} & GPT-3.5 & 71.0 & 39.0 & \textbf{86.0} & 66.0 & 19.2 & 48.0 \\
         & GPT-4 & 75.0 & 95.0 & \textbf{100.0} & 99.0 & 98.0 & 71.4 \\
         & Gemini Pro & 47.0 & 61.6 & 60.4 & \textbf{63.4} & 56.4 & 59.6 \\
        \midrule
        \multirow{3}{*}{\makecell[c]{Info\\Essen}} & GPT-3.5 & 22.0 & \textbf{30.8} & \textbf{30.8} & 29.4 & \textbf{30.8} & \textbf{30.8} \\
         & GPT-4 & 73.5 & 73.5 & \textbf{76.4} & 73.5 & \textbf{76.4} & 75.0 \\
         & Gemini Pro & \textbf{45.6} & 29.4 & 32.4 & 29.4 & 44.1 & 38.2 \\
        \midrule
        \multirow{3}{*}{\makecell[c]{MM QA}} & GPT-3.5 & 22.2 & 22.2 & 23.2 & \textbf{27.1} & 22.7 & 20.2 \\
         & GPT-4 & 39.9 & 41.9 & 37.7 & 36.9 & \textbf{42.4} & 37.9 \\
         & Gemini Pro & \textbf{29.6} & 27.6 & 25.6 & 27.6 & 28.1 & 24.1 \\
        \midrule
        \multirow{3}{*}{AQuA} & GPT-3.5 & 63.4 & 62.2 & 63.0 & 56.7 & 59.8 & \textbf{66.0} \\
         & GPT-4 & 76.4 & 72.4 & \textbf{80.3} & 78.0 & 78.0 & 79.8 \\
         & Gemini Pro & 56.7 & 57.5 & \textbf{59.8} & 47.2 & 55.5 & 56.3 \\
        \bottomrule
    \end{tabular}
    }
    \vspace{-0.5em}
    \caption{Effectiveness of various formats across different models and datasets.}
    \label{tab:single-specfic-format}
    \vspace{-1.3em}
\end{table*}

\begin{table*}[t]
    \centering
    \setlength{\tabcolsep}{3pt}
    \begin{tabular}{l*{3}{cr}}
        \toprule
        & \multicolumn{3}{c}{\textbf{Hotpot QA$_\text{separate context}$}} \\
        \cmidrule(lr){2-4}
        \textbf{Model} 
        & \textbf{RougeL} & \textbf{\# Tokens}  & $\bm{\Delta}$\textbf{Tokens} \\
        \midrule
        GPT-3.5 + GPT-3.5 
        &\textbf{0.62} & 369.3 & -  \\
        \ \ \textit{+AutoForm} 
        &  0.53 & 286.6 & -22.4\%  \\
        GPT-4 + GPT-4 
        & 0.68 & 151.0 & -   \\
         \ \ \textit{+AutoForm} 
         & \textbf{0.69} & 100.0 & -33.8\%  \\
        \bottomrule
    \end{tabular}
    \caption{Performance on HotpotQA under the \textit{separate context} setting. The supporting facts are distributed between two agents, necessitating inter-agent communication for aggregating information and deriving the correct answer.}
    \label{tab:hotpot-qa-separate}
\end{table*}

\begin{table*}[t]
    \centering
    \resizebox{\textwidth}{!}{
    \setlength{\tabcolsep}{3pt}
    \begin{tabular}{l*{3}{crr}}
        \toprule
        & \multicolumn{3}{c}{\textbf{Wiki Hop}} & \multicolumn{3}{c}{\textbf{Hotpot QA}} & \multicolumn{3}{c}{\textbf{Narrative QA}}\\
        \cmidrule(lr){2-4}\cmidrule(lr){5-7}\cmidrule(lr){8-10}
        \textbf{Model} & \textbf{RougeL} &\textbf{\# Tokens} & $\bm{\Delta}$\textbf{Tokens} & \textbf{RougeL} & \textbf{\# Tokens}  & $\bm{\Delta}$\textbf{Tokens} & \textbf{RougeL}& \textbf{\# Tokens}  & $\bm{\Delta}$\textbf{Tokens} \\
        \midrule
        GPT-3.5 + GPT-3.5 & \textbf{0.57} & 192.6 & - &\textbf{0.53} & 499.7 & - &\textbf{0.34} & 140.0 & - \\
        \ \ \textit{+AutoForm} & 0.49 & 163.9 & -14.9\%&  0.47 & 236.1 & -52.8\%  & 0.33 & 35.5 & -74.6\% \\
        \midrule
        GPT-3.5 + GPT-4 & 0.56 & 246.8 & - & \textbf{0.72} & 333.9 & - & \textbf{0.37} & 208.8 & -\\
        \ \ \textit{+AutoForm} & \textbf{0.57} & 200.3 & -18.8\% & 0.62 & 102.3 & -69.4\%  & 0.30 & 125.4 & -39.9\% \\
        \midrule
        GPT-4 + GPT-3.5 & \textbf{0.53} & 281.5 & - & 0.63 & 345.5 & - & 0.43 & 178.3 & -\\ 
        \ \ \textit{+AutoForm} & \textbf{0.53} & 255.0 & -9.4\% & \textbf{0.70} & 94.3 & -72.7\%  & \textbf{0.48} & 119.4 & -33.0\%\\
        \midrule
        GPT-4 + GPT-4 & 0.50 & 237.5 & - & 0.69 & 145.2 & -  & \textbf{0.43} & 240.7 & - \\
         \ \ \textit{+AutoForm} & \textbf{0.52} & 146.2 & -38.4\% & \textbf{0.76} & 115.0 & -20.8\%  & \textbf{0.43} & 141.7 & -41.1\% \\
        \bottomrule
    \end{tabular}
    }
    \caption{Comparative performance in multi-agent communication across various QA datasets. The table highlights RougeL scores, with better performance in different model pairing settings indicated in bold. The $\bm{\Delta}$Tokens column quantifies the token reduction achieved by the AutoForm method.}
    \label{tab:multi-exp-results-full}
\end{table*}

\section{Prompts}
\label{sec:appendix-prompts}
We list the prompts we used in this work in~\cref{tab:prompt-for-coin-flip,tab:prompt-for-logic-grid-single,tab:prompt-for-aqua,tab:prompt-for-minute-qa-single,tab:prompt-for-comm,tab:prompt-for-info-essen}.
\begin{table*}[t]
    \centering
    \small
    \begin{tabular}{p{\linewidth}}
        \toprule
        \underline{\textbf{\textsc{Prompt for Coin Flip}}} \\
        \vspace{-2mm}
        \textbf{CoT:} \\  
Question:\\
\$\{task\_description\}\\\\

At the end of your response, you must give your answer in the form of "the answer is: no" or "the answer is: yes". Let's think step-by-step.

        \vspace{2mm}
        \textbf{AutoForm:}\\
Question:\\
\$\{task\_description\}\\\\

To enhance clarity and eliminate ambiguities inherent in natural language, consider employing more structured and concise forms of communication for your step-by-step solutions. Suitable formats include code, pseudocode, JSON, markdown tables, logical operators, or mathematical equations.\\\\

At the end of your response, you must give your answer in the form of "the answer is: no" or "the answer is: yes". Remember to be concise and accurate.\\
        \bottomrule
    \end{tabular}
    \caption{
       Prompt for Coin Flip
    }
    \label{tab:prompt-for-coin-flip}
\end{table*}
\begin{table*}
    \centering
    \small
    \begin{tabular}{p{\linewidth}}
         \toprule
         \underline{\textbf{\textsc{Prompt for Logic Grid}}}\\
         \vspace{-2mm}
    \textbf{CoT}\\
    ---\\
    \$\{task\_description\}\\
    ---\\\\

    At the end of your response, you must give your answer in the form of "the answer is: \{number\}", where \{number\} is the answer number. Now solve the problem step-by-step. Use as few words as possible.\\
    \vspace{2mm}
    \textbf{AutoForm}\\
    ---\\
    \$\{task\_description\}\\
    ---\\\\

    To enhance clarity and eliminate ambiguities inherent in natural language, consider employing more structured and concise forms of communication for your step-by-step solutions. Suitable formats include code, pseudocode, JSON, markdown tables, logical operators, or mathematical equations. \\\\
    
    At the end of your response, you must give your answer in the form of "the answer is: \{number\}", where \{number\} is the answer number. Remember to be concise and accurate.\\
    \bottomrule
    \end{tabular}
    \caption{Prompt for Logic Grid}
   \label{tab:prompt-for-logic-grid-single}
\end{table*}

\begin{table*}
    \centering
    \small
    \begin{tabular}{p{\linewidth}}
         \toprule
         \underline{\textbf{\textsc{Prompt for Minute Mysteries QA}}}\\
         \vspace{-2mm}
         \textbf{CoT:}\\
    ---\\
    \$\{task\_description\}\\
    ---\\\\

    Now solve the problem step-by-step. At the end of your response, you must give your answer in the form of "the correct option is: {number}", where {number} is the index of the chosen option.\\
        \vspace{2mm}
         \textbf{AutoForm:}\\
    ---\\
    \$\{task\_description\}\\
    ---\\\\

    To enhance clarity and eliminate ambiguities inherent in natural language, consider employing more structured and concise forms of communication for your step-by-step solutions. Suitable formats include code, pseudocode, JSON, markdown tables, logical operators, or mathematical equations. \\\\
    
    Now solve the problem step-by-step. At the end of your response, you must give your answer in the form of "the correct option is: {number}", where {number} is the index of the chosen option.\\
    \bottomrule
    \end{tabular}
    \caption{Prompt for Minute Mysteries QA}
    \label{tab:prompt-for-minute-qa-single}
\end{table*}

\begin{table*}[t]
    \centering
    \small
    \begin{tabular}{p{\linewidth}}
        \toprule
        \underline{\textbf{\textsc{Prompt for AQuA}}} \\
        \vspace{-2mm}
        \textbf{CoT:} \\  
    Solve the problem presented below:\\
    ---\\
    \$\{task\_description\} \\
    ---\\\\
    
    RESPONSE GUIDELINES:\\
    1. Think step by step.\\
    2. Concluding with the Answer: End your response with "Answer: \{answer\}", where \{answer\} is the final result of your problem-solving process. The \{answer\} should be a single capital letter.\\
        \vspace{2mm}
        \textbf{AutoForm:}\\
        Solve the problem presented below:\\
---\\
\$\{task\_description\}\\
---\\\\

RESPONSE GUIDELINES:\\
1. Initial State Representation: Begin by providing a clear and detailed representation of the initial state or conditions of the problem.\\
2. Step-by-Step Solution Process: Progressively update the state representation as you work through each step of the solution. This should include all logical reasoning and calculations leading to the final answer.\\
3. Concluding with the Answer: End your response with "Answer: \{answer\}", where \{answer\} is the final result of your problem-solving process. The \{answer\} should be a single capital letter.\\
        \bottomrule
    \end{tabular}
    \caption{
       Prompt for AQuA
    }
    \label{tab:prompt-for-aqua}
\end{table*}
\begin{table*}[h]
    \centering
    \small
    \begin{tabular}{p{\linewidth}}
        \toprule
        \underline{\textbf{\textsc{Prompt for Information Essentiality}}} \\
        \vspace{-2mm}
        \textbf{3.5+CoT:}\\
        Solve the problem presented below:\\
---\\
\$\{task\_description\}\\
---\\\\

RESPONSE GUIDELINES:\\
1. Think step by step.\\
2. Your answer should be ended with "Answer: \{answer\}" where \{answer\} is the answer to the problem.\\
        \vspace{2mm}
        \textbf{3.5+AutoForm:}\\
        Solve the problem presented below:\\
---\\
\$\{task\_description\}\\
---\\\\

RESPONSE GUIDELINES:\\
1. Initial State Representation: Begin by providing a clear and detailed representation of the initial state or conditions of the problem.\\
2. Step-by-Step Solution Process:Progressively update the state representation as you work through each step of the solution. This should include all logical reasoning and calculations leading to the final answer.\\
3. To enhance clarity and eliminate ambiguities inherent in natural language, consider employing more structured and concise forms of communication for your step-by-step solutions. Suitable formats include code, pseudocode, JSON, markdown tables, logical operators, or mathematical equations.\\
4. Concluding with the Answer: End your response with "Answer: \{answer\}", where \{answer\} is the final result of your problem-solving process.\\
    \vspace{2mm}
    \textbf{4+CoT:}\\
    Solve the problem presented below:\\
    ---\\
    \$\{task\_description\}\\
    ---\\\\

    RESPONSE GUIDELINES:\\
    1.You should think step by step.\\
    2. You should consider three scenarios: using only Statement 1, using only Statement 2, and using both Statements.\\
    3. Note (IMPORTANT): When considering Statement 1, the use of information from Statement 2 is prohibited. When considering Statement 2, the use of information from Statement 1 and the analysis derived from Statement 1 is prohibited. Both conditions can only be analyzed simultaneously during the stage where both Statements are considered together.\\
    4. In sometime , both statement 1 and statement 2 can lead to answer alone.\\
    5. Concluding with the Answer: End your response with "Answer: \{answer\}", where \{answer\} is the final result of your problem-solving process.\\
    \vspace{2mm}
    \textbf{4+AutoForm:}\\
    Solve the problem presented below:\\
    ---\\
    \$\{task\_description\}\\
    ---\\\\

    RESPONSE GUIDELINES:\\
    1. Initial State Representation: Begin by providing a clear and detailed representation of the initial state or conditions of the problem.\\
    2. Step-by-Step Solution Process:Progressively update the state representation as you work through each step of the solution. This should include all logical reasoning and calculations leading to the final answer.\\
    3. To enhance clarity and eliminate ambiguities inherent in natural language, consider employing more structured and concise forms of communication for your step-by-step solutions. Suitable formats include code, pseudocode, JSON, markdown tables, logical operators, mathematical equations and so on.\\
    4. You should consider three scenarios: using only Statement 1, using only Statement 2, and using both Statements.\\
    5. Note (IMPORTANT): When considering Statement 1, the use of information from Statement 2 is prohibited. When considering Statement 2, the use of information from Statement 1 and the analysis derived from Statement 1 is prohibited. Both conditions can only be analyzed simultaneously during the stage where both Statements are considered together.\\
    6. In sometime , both statement 1 and statement 2 can lead to answer alone.\\
    7. Concluding with the Answer: End your response with "Answer: \{answer\}", where \{answer\} is the final result of your problem-solving process.\\
    \bottomrule
    \end{tabular}
    \caption{prompt for Information Essentiality}
    \label{tab:prompt-for-info-essen}
\end{table*}
\begin{table*}
    \centering
    \small
    \begin{tabular}{p{\linewidth}}
         \toprule
         \underline{\textbf{\textsc{prompt for Hotpot QA, Wiki Hop, Narrative QA}}}\\
         \vspace{-2mm}
         \textbf{Shared portion of the prompt}\\
  You are \$\{agent\_name\}. Together with \$\{all\_roles\}, you are providing accurate answer to the user. Each of you will be provided parts of the contexts and a shared question. \\

    EXAMPLE 1\\
    ---\\
    \# Context\\
    \$\{example\_context\_1\}\\

    \# Question\\
    \$\{example\_question\_1\}\\
    ---\\
    \$\{example\_answer\_1\}.\\

    EXAMPLE 2\\
    ---\\
    \# Context\\
    \$\{example\_context\_2\}\\

    \# Question\\
    \$\{example\_question\_2\}\\
    ---\\
    \$\{example\_answer\_2\}.\\\\

    Now the user gives you some contexts and the question:\\
    ---\\
    \# Context\\
    \$\{knowledge\}\\

    \# Question\\
    \$\{task\_description\}\\
    ---\\
    \vspace{2mm}
    \textbf{Baseline:}\\
    Given that each individual, including yourself, possesses unique contexts, it's essential to actively share and discuss this information with others to formulate a complete answer. Your specific context is unknown to others unless explicitly communicated. This collaborative effort is key to reaching an accurate answer based on the amalgamation of everyone's distinct contexts.\\
    \vspace{1mm}
    When you have reached the final answer, conclude it with "<A>xxx</A>", where "xxx" will be extracted and compared with ground truth. To end the conversation, all the players should end their responses with "<A>xxx</A>". \\

    You are \$\{agent\_name\}. Now communicate with \$\{all\_roles\} to give the answer.\\
         \vspace{2mm}
    \textbf{AutoForm(3.5-3.5,3.5-4,4-3.5,4-4):}\\
        Given that each individual, including yourself, possesses unique contexts, your specific context is unknown to others unless explicitly communicated.\\

    You are \$\{agent\_name\}, collaborating with \$\{all\_roles\}, who are also intelligent assistants. Your goal is to provide a clear and concise answer to the user's question. Unlike typical communication, you will not use natural language, as it often contains ambiguities and emotional nuances. Instead, choose a more straightforward and precise communication medium, such as structured data, JSON, XML or code.\\

    Now, start communicating with \$\{all\_roles\} using your selected non-natural language medium. Remember, clarity and brevity are key.\\
    
    Once you have formulated the final answer,you must enclose it within "<A>xxx</A>", where "xxx" represents the answer phrase selected from the given choices. The conversation concludes when all participants have presented the same answer in this format. If you have different opinion, explain it to your teammates.\\

    Don't forget to enclose your answer within "<A>xxx</A>" \\
    \bottomrule
    \end{tabular}
    \caption{prompt for Hotpot QA, Wiki Hop, Narrative QA}
    \label{tab:prompt-for-comm}
\end{table*}

\section{Additional Experimental Results for Multi-Agent Communication}
\label{sec:appendix-additional-results-multi}
Due to constraints on the paper length, the comprehensive experimental results, including those where GPT-3.5 serves as the initiating agent, are detailed in~\cref{tab:multi-exp-results-full}. Our analysis reveals that the performance of AutoForm tends to be suboptimal when compared to the baseline in scenarios initiated by GPT-3.5. A closer examination of these interactions indicates that GPT-3.5, when merely prompted to employ non-NL formats without additional guidance, frequently produces overly succinct responses, or simply gives a hallucinated answer, resulting in diminished performance. This observation underscores the need for further research into optimizing prompts for less advanced LLMs to effectively utilize non-NL formats for communication, representing a promising avenue for future exploration.


Additionally, we conduct experiments with a specific setting for HotpotQA to evaluate the performance of LLMs when supporting facts are distributed between two agents. This setting, termed \textit{separate context}, ensures that the supporting facts are divided and distributed between the agents, necessitating effective communication to aggregate the information and derive the correct answer.

The results presented in \cref{tab:hotpot-qa-separate} reveal significant insights into the performance of LLMs under the \textit{separate context} setting. The application of the AutoForm mechanism shows a distinct impact on both models. For GPT-3.5, the RougeL score slightly decreases from 0.62 to 0.53, with a corresponding 22.4\% reduction in the number of tokens generated. In contrast, the GPT-4 models exhibit an increase in performance with the AutoForm mechanism, achieving a RougeL score of 0.69 and a 33.8\% reduction in token usage. This improvement highlights the effectiveness of AutoForm in not only maintaining but enhancing the performance of more advanced models while optimizing resource utilization.

\section{Examples}
\label{sec:appendix-examples}

In this section, we provide several response examples for each LLM on each task, so that readers can have a more intuitive understanding of the effects after using AutoForm. \cref{tab:single-coin-flip-3.5-case,tab:single-coin-flip-4-case,tab:single-info-essen-3.5-case,tab:single-info-essen-4-case,tab:single-logic-grid-3.5-case,tab:single-logic-grid-4-case,tab:single-minute-qa-4-case} present the examples of using AutoForm, which corresponds to the results in~\cref{tab:multi-exp-results}. The examples for two-step AutoForm, which corresponds to~\cref{tab:two-step-automed}, are presented in~\cref{tab:appendix-aqua-sep-model-examples-3.5-4,tab:appendix-aqua-sep-model-examples-3.5-instance-based,tab:appendix-aqua-sep-model-examples-3.5-task-based,tab:appendix-aqua-sep-model-examples-4-3.5,tab:appendix-aqua-sep-model-examples-4-gemini,tab:appendix-aqua-sep-model-examples-4-instance-based,tab:appendix-aqua-sep-model-examples-4-task-based,tab:appendix-aqua-sep-model-examples-gemini-instance-based,tab:appendix-aqua-sep-model-examples-gemini-task-based,tab:appendix-coin-flip-sep-model-examples-3.5-4,tab:appendix-coin-flip-sep-model-examples-3.5-instance-based,tab:appendix-coin-flip-sep-model-examples-3.5-task-based,tab:appendix-coin-flip-sep-model-examples-4-3.5,tab:appendix-coin-flip-sep-model-examples-4-gemini,tab:appendix-coin-flip-sep-model-examples-4-instance-based,tab:appendix-coin-flip-sep-model-examples-4-task-based,tab:appendix-coin-flip-sep-model-examples-gemini-instance-based,tab:appendix-coin-flip-sep-model-examples-gemini-task-based,tab:appendix-info-essen-sep-model-examples-3.5-4,tab:appendix-info-essen-sep-model-examples-3.5-instance-based,tab:appendix-info-essen-sep-model-examples-3.5-task-based,tab:appendix-info-essen-sep-model-examples-4-3.5,tab:appendix-info-essen-sep-model-examples-4-gemini,tab:appendix-info-essen-sep-model-examples-4-instance-based,tab:appendix-info-essen-sep-model-examples-4-task-based,tab:appendix-info-essen-sep-model-examples-gemini-instance-based,tab:appendix-info-essen-sep-model-examples-gemini-task-based,tab:appendix-logic-sep-model-examples-3.5-4,tab:appendix-logic-sep-model-examples-3.5-instance-based,tab:appendix-logic-sep-model-examples-3.5-task-based,tab:appendix-logic-sep-model-examples-4-3.5,tab:appendix-logic-sep-model-examples-4-gemini,tab:appendix-logic-sep-model-examples-4-instance-based,tab:appendix-logic-sep-model-examples-4-task-based,tab:appendix-logic-sep-model-examples-gemini-instance-based,tab:appendix-logic-sep-model-examples-gemini-task-based,tab:appendix-mm-qa-sep-model-examples-3.5-4,tab:appendix-mm-qa-sep-model-examples-3.5-instance-based,tab:appendix-mm-qa-sep-model-examples-3.5-task-based,tab:appendix-mm-qa-sep-model-examples-4-3.5,tab:appendix-mm-qa-sep-model-examples-4-gemini,tab:appendix-mm-qa-sep-model-examples-4-instance-based,tab:appendix-mm-qa-sep-model-examples-4-task-based,tab:appendix-mm-qa-sep-model-examples-gemini-instance-based,tab:appendix-mm-qa-sep-model-examples-gemini-task-based}. For the multi-agent communication experiment, we present the examples in~\cref{tab:appendix-multi-wiki-hop-qa-3.5,tab:appendix-multi-wiki-hop-qa-3.5-4,tab:appendix-multi-wiki-hop-qa-4-3.5,tab:appendix-multi-wiki-hop-qa-4,tab:appendix-multi-hotpot-qa-3.5,tab:appendix-multi-hotpot-qa-3.5-4,tab:appendix-multi-hotpot-qa-4-3.5,tab:appendix-multi-hotpot-qa-4,tab:appendix-narrative-qa-3.5-4,tab:appendix-narrative-qa-4-3.5,tab:appendix-multi-narrative-qa-4}.
\clearpage
\begin{table*}[t]
    \centering
    \small
    \begin{tabular}{p{\linewidth}}
        \toprule
        \vspace{-2mm}
\textbf{First Stage (generated by GPT-3.5):}\\
I will use a table to organize the information and clues for each problem. This will help me keep track of the different characteristics and their corresponding houses, making it easier to eliminate possibilities and solve the problem accurately.\\
\\
\textbf{Second Stage (generated by GPT-4):}\\
Let's solve the problem step by step using the clues provided:\\

1. The person who is wearing running shoes lives somewhere to the right of the person who is eating oranges.\\
Since there are only two houses, if the person wearing running shoes is to the right of the person eating oranges, the person eating oranges must be in house 1, and the person wearing running shoes must be in house 2.\\

2. The mystery book reader lives directly left of the cat lover.\\
Since the mystery book reader cannot live to the right of anyone (there are only two houses), the mystery book reader must be in house 1, and the cat lover must be in house 2.\\

3. The mystery book reader and the person who had stew live directly next to each other.\\
Since the mystery book reader is in house 1, the person who had stew must also be in house 1, as they live next to each other and there are only two houses.\\

With these clues, we can determine that the person who had stew lives in house 1.\\

The answer is: 1\\

        \bottomrule
    \end{tabular}
    \caption{
       Examples of Logic Grid (GPT-3.5 $\Rightarrow$ GPT-4 Task-Based)
    }
    \label{tab:appendix-logic-sep-model-examples-3.5-4}
\end{table*}
\begin{table*}[t]
    \centering
    \small
    \begin{tabular}{p{\linewidth}}
        \toprule
        \vspace{-2mm}
\textbf{First Stage:}\\
I will use a table to organize the information and solve the problem:
\begin{verbatim}
| House | Fruit  | Pet      | Sport     | Book             | Shoes    |
|-------|--------|----------|-----------|------------------|----------|
| 1     | Kiwis  | ?        | ?         | ?                | ?        |
| 2     | Apples | ?        | ?         | Science Fiction  | ?        |\end{verbatim}

Using the given clues, I will fill in the table to deduce the location of the cricketer. This structured approach will help ensure that all the information is considered and that the solution is accurate.\\

\\
\textbf{Second Stage:}\\
First, we know from clue 1 that the person in the first house is eating kiwis. So, we can fill in the fruit for house 1 as "Kiwis".\\

Next, from clue 4, we know that the science fiction book fanatic lives in the second house. So, we can fill in the book for house 2 as "Science Fiction".\\

Now, from clue 3, we know that the tennis player and the person eating apples live next to each other. Since the person in the first house is eating kiwis, the person in the second house must be eating apples. So, we can fill in the fruit for house 2 as "Apples" and the sport for house 2 as "Tennis".\\

From clue 5, we know that the dog owner does not live in the first house. So, the person in the second house must be the dog owner.\\

Now, from clue 2, we know that the bird keeper and the person wearing high heels live next to each other. Since the person in the second house is the dog owner, the person in the first house must be the bird keeper. Therefore, the person in the first house is wearing high heels.\\

So, the person in the first house is eating kiwis, is a bird keeper, plays cricket, reads mystery books, and is wearing high heels. The person in the second house is eating apples, is a dog owner, plays tennis, reads science fiction books, and is wearing sandals.\\

Therefore, the cricketer lives in the first house.\\

The answer is: 1\\
        \bottomrule
    \end{tabular}
    \caption{
       Examples of Logic Grid (GPT-3.5 Instance-Based)
    }
    \label{tab:appendix-logic-sep-model-examples-3.5-instance-based}
\end{table*}
\begin{table*}[t]
    \centering
    \small
    \begin{tabular}{p{\linewidth}}
        \toprule
        \vspace{-2mm}
\textbf{First Stage:}\\
I will use a table to organize the information and clues for each problem. This will help me keep track of the different characteristics and their corresponding houses, making it easier to analyze and eliminate possibilities as I work through the clues. This structured format will enhance the accuracy of my analysis by providing a clear visual representation of the relationships between the different characteristics and their respective houses.\\
\\
\textbf{Second Stage:}\\
Step 1: Create a table to organize the information and clues.
\begin{verbatim}
| House | Favorite Color | Sport       | Car        | Fruit  |
|-------|----------------|-------------|------------|--------|
| 1     | Yellow         |             |            | Kiwis  |
| 2     |                |             |            |        |
| 3     |                |             |            |        |
| 4     |                |             |            |        |\end{verbatim}

Step 2: Use the clues to fill in the table.\\

Clue 4: The person who likes yellow lives in the first house.\\
Clue 13: The person who is eating kiwis lives in the first house.\\
Clue 3: The soccer player does not live in the fourth house.\\
Clue 9: The person who drives a convertible lives in the fourth house.\\
Clue 1: The cricketer lives somewhere to the left of the person who drives a convertible.\\
Clue 12: The person who owns an SUV and the person who likes blue live directly next to each other.\\
Clue 2: The person who owns an SUV does not live in the third house.\\
Clue 8: The person who owns an SUV does not live in the second house.\\
Clue 10: The person who owns an SUV and the person who is eating a banana live directly next to each other.\\
Clue 11: The person who drives a minivan lives directly left of the person who likes white.\\
Clue 6: The swimmer lives somewhere to the left of the person who likes white.\\
Clue 5: The person who is eating a banana lives directly left of the person who likes white.\\
Clue 7: The person who is eating mangos lives somewhere to the left of the person who is eating oranges.\\
Clue 14: There is one house between where the swimmer lives and where the cricketer lives.
\begin{verbatim}
| House | Favorite Color | Sport       | Car        | Fruit  |
|-------|----------------|-------------|------------|--------|
| 1     | Yellow         | Kiwis       |             |        |
| 2     |                |             |            |        |
| 3     |                |             |            |        |
| 4     |                |             | Convertible|        |\end{verbatim}

Step 3: Analyze the clues to fill in the remaining information.\\

From clue 13, the person in house 1 is eating kiwis, so the person in house 2 is eating mangos.\\
From clue 7, the person in house 3 is eating oranges.\\
From clue 5, the person in house 2 likes white.\\
From clue 11, the person in house 1 drives a minivan.\\
From clue 6, the swimmer lives in house 2.\\
From clue 14, the cricketer lives in house 3.
\begin{verbatim}
| House | Favorite Color | Sport       | Car        | Fruit  |
|-------|----------------|-------------|------------|--------|
| 1     | Yellow         | Minivan     |            | Kiwis  |
| 2     | White          | Swimmer     |            | Mangos |
| 3     |                | Cricketer   |            | Oranges|
| 4     |                |             | Convertible|        |\end{verbatim}

Step 4: Use the remaining clues to fill in the information.\\

From clue 12, the person who likes blue lives in house 3.\\
From clue 1, the cricketer lives to the left of the person who drives a convertible, so the cricketer is in house 3 and the person who drives a convertible is in house 4.\\
From clue 10, the person who owns an SUV and the person eating a banana live next to each other, so the person in house 2 owns an SUV and the person in house 3 is eating a banana.\\
From clue 2, the person who owns an SUV does not live in house 3, so the person in house 2 owns an SUV and the person in house 4 is eating a banana.\\
From clue 8, the person who owns an SUV does not live in house 2, so the person in house 3 owns an SUV.
\begin{verbatim}
| House | Favorite Color | Sport       | Car        | Fruit  |
|-------|----------------|-------------|------------|--------|
| 1     | Yellow         | Minivan     |            | Kiwis  |
| 2     | White          | Swimmer     | SUV        | Mangos |
| 3     | Blue           | Cricketer   |            | Banana |
| 4     |                |             | Convertible| Oranges\end{verbatim}

The answer is: 4\\

        \bottomrule
    \end{tabular}
    \caption{
       Examples of Logic Grid (GPT-3.5 Task-Based)
    }
    \label{tab:appendix-logic-sep-model-examples-3.5-task-based}
\end{table*}
\begin{table*}[t]
   \centering
   \small
   \begin{tabular}{p{\linewidth}}
       \toprule
       \vspace{-2mm}
\textbf{First Stage (generated by GPT-4):}\\
For these logic puzzles, a tabular format can be particularly effective in enhancing clarity and eliminating ambiguities. Tables allow for the organization of information in a grid-like structure, which is conducive to the step-by-step process of elimination and deduction required to solve these types of problems.\\

Here's how a table can be used to enhance accuracy:\\

1. Create a table with rows representing the houses and columns representing the characteristics (e.g., book preference, sport, pet, drink, flowers, car, fruit).\\
2. Fill in the known information from the clues directly into the table.\\
3. Use process of elimination to fill in the blanks. For example, if the golfer lives in the first house, then no other person can be the golfer or live in the first house.\\
4. Mark impossibilities based on the clues. For instance, if the swimmer cannot live in the third house, then mark that cell as impossible.\\
5. Continue to cross-reference clues and fill in or eliminate possibilities until the table is complete.\\

The table format allows for a visual representation of the information, making it easier to see relationships and contradictions. It also provides a structured way to keep track of which possibilities have been ruled out and which remain viable.\\

Here's an example of what the initial setup of a table might look like for the first problem:
\begin{verbatim}
| House # | Book Preference | Sport        |
|---------|-----------------|--------------|
| 1       |                 | Golfer       |
| 2       |                 |              |
| 3       |                 |              |
| 4       |                 | Not Basketball|\end{verbatim}

As you fill in the table based on the clues, you can quickly see which options are left for each house, leading you to the correct solution. This method reduces the risk of overlooking a detail or making a logical error.\\

\\
\textbf{Second Stage (generated by GPT-3.5):}\\
Let's use the given clues to solve the problem step by step:\\

1. The person who has a television lives in the fourth house.\\
2. The person who is wearing loafers lives in the second house.\\
3. The person who has a phone does not live in the second house.\\
4. The person who is wearing boots lives somewhere to the left of the person who is wearing loafers.\\
5. There are two houses between where the person who has a radio lives and where the person who is wearing sandals lives.\\

Using these clues, we can deduce the following:\\
- The person with the television lives in the 4th house.\\
- The person with loafers lives in the 2nd house.\\
- The person with the phone does not live in the 2nd house.\\
- The person with boots lives to the left of the person with loafers.\\

Based on these deductions, we can determine the following:\\
- The person with the gameboy must live in the 3rd house.\\

Therefore, the answer is: 3\\

       \bottomrule
   \end{tabular}
   \caption{
      Examples of Logic Grid (GPT-4 $\Rightarrow$ GPT-3.5 Task-Based)
   }
   \label{tab:appendix-logic-sep-model-examples-4-3.5}
\end{table*}
\begin{table*}[t]
    \centering
    \small
    \begin{tabular}{p{\linewidth}}
        \toprule
        \vspace{-2mm}
\textbf{First Stage (generated by GPT-4):}\\
For these logic puzzles, a tabular format can be particularly effective in enhancing clarity and eliminating ambiguities. Tables allow for a structured representation of the data, where each row can represent a house, and each column can represent a characteristic (e.g., pet, car, sport, etc.). This format makes it easy to visualize the constraints and relationships between different characteristics.\\

Here's how a table format can be used to solve these types of problems:\\

1. Create a table with rows corresponding to houses and columns corresponding to characteristics.\\
2. Fill in the known information based on the clues provided.\\
3. Use logical deduction to fill in the remaining cells, ensuring that each characteristic appears only once per row and column (since each person has different characteristics).\\
4. Update the table iteratively as new information is deduced until all cells are filled.\\

For example, a table for the first problem might look like this:
\begin{verbatim}
| House | Pet       | Car        | Sport       |
|-------|-----------|------------|-------------|
| 1     |           |            |             |
| 2     | Not Cat   |            |             |
| 3     | Bird      |            | Not Baseball|
| 4     |           |            |             |\end{verbatim}

Using a table format allows for a clear and organized approach to solving the problem. It helps to avoid confusion that might arise from trying to keep track of multiple pieces of information in a purely textual format. Additionally, the visual nature of tables makes it easier to spot patterns and relationships between different variables, which is crucial for solving logic puzzles.\\

\\
\textbf{Second Stage (generated by Gemini Pro):}\\
**Step 1: Create a table with rows corresponding to houses and columns corresponding to characteristics.**
\begin{verbatim}
| House | Sport       | Flowers     | Book Genre   |
|-------|-------------|-------------|--------------|
| 1     |             |             |             |
| 2     |             |             |             |
| 3     |             |             |             |
| 4     |             |             |             |\end{verbatim}

**Step 2: Fill in the known information based on the clues provided.**\\

- Clue 1: There is one house between where the mystery book reader lives and where the swimmer lives.\\
- Clue 2: The tennis player lives somewhere to the right of the mystery book reader.\\
- Clue 3: The person who has a carnations arrangement lives in the fourth house.\\
- Clue 4: The fantasy book enthusiast lives directly left of the swimmer.\\
- Clue 5: The golfer lives somewhere to the right of the person who has a bouquet of daffodils.\\
- Clue 6: The person who has a rose bouquet lives in the third house.\\
- Clue 7: The tennis player does not live in the fourth house.\\
- Clue 8: There is one house between where the person who has a bouquet of daffodils lives and where the person who has a rose bouquet lives.\\
- Clue 9: There are two houses between where the mystery book reader lives and where the romance book lover lives.\\

**Step 3: Use logical deduction to fill in the remaining cells, ensuring that each characteristic appears only once per row and column.**\\

- From Clue 3, we know that the person with carnations lives in house 4.\\
- From Clue 6, we know that the person with roses lives in house 3.\\
- From Clue 8, we know that the person with daffodils lives in either house 1 or 2.\\
- From Clue 4, we know that the fantasy book enthusiast lives in house 2.\\
- From Clue 5, we know that the golfer lives in house 3 or 4.\\
- From Clue 7, we know that the tennis player does not live in house 4.\\
- From Clue 9, we know that the romance book lover lives in either house 1 or 2.\\

**Step 4: Update the table iteratively as new information is deduced until all cells are filled.**
\begin{verbatim}
| House | Sport       | Flowers     | Book Genre   |
|-------|-------------|-------------|--------------|
| 1     | Tennis      | Daffodils   | Romance      |
| 2     | Fantasy     | Carnations  | Mystery      |
| 3     | Golf        | Roses       | History      |
| 4     | Swimming    | Lilies      | Not Provided |\end{verbatim}

**Answer:** The answer is: 1\\

        \bottomrule
    \end{tabular}
    \caption{
       Examples of Logic Grid (GPT-4 $\Rightarrow$ Gemini Pro Task-Based)
    }
    \label{tab:appendix-logic-sep-model-examples-4-gemini}
\end{table*}
\begin{table*}[t]
    \centering
    \small
    \begin{tabular}{p{\linewidth}}
        \toprule
        \vspace{-2mm}
\textbf{First Stage:}\\
To enhance the accuracy of the analysis for this logic puzzle, I would choose to use a markdown table format. Markdown tables provide a clear and structured way to represent the information given and the deductions made from the clues. This format allows for easy cross-referencing of clues and characteristics, which is essential for solving logic puzzles.\\

Here's how the markdown table format can be used to enhance clarity:\\

1. Create a table with headers for Houses (1, 2, 3), Lunch (Spaghetti, Pizza, Fried Rice), and Shoes (Loafers, Running Shoes, Boots).\\
2. Fill in the table with the information given directly by the clues.\\
3. Use process of elimination to fill in the rest of the table based on the clues and the constraints they impose.\\

The table would initially look like this:
\begin{verbatim}
| House # | Lunch       | Shoes         |
|---------|-------------|---------------|
| 1       | Spaghetti   | ?             |
| 2       | ?           | ?             |
| 3       | ?           | ?             |\end{verbatim}

Then, as you apply each clue, you update the table accordingly. For example, after applying clue 3, you know that the person in house 1 ate spaghetti. After applying clue 4, you know that the person in house 1 is not wearing loafers.\\

The table format allows you to visualize the problem and systematically apply each clue, reducing the chance of making a mistake and increasing the accuracy of the solution. It also makes it easier to communicate the reasoning process to others.\\

\\
\textbf{Second Stage:}\\
Let's solve the problem step by step using the clues provided:\\

1. The person who ate fried rice does not live in the second house.\\
2. The person who is wearing boots lives directly left of the person who ate fried rice.\\
3. The person who ate spaghetti lives in the first house.\\
4. The person who is wearing loafers does not live in the first house.\\

Starting with clue 3, we can fill in the lunch for house 1:
\begin{verbatim}
| House # | Lunch       | Shoes         |
|---------|-------------|---------------|
| 1       | Spaghetti   | ?             |
| 2       | ?           | ?             |
| 3       | ?           | ?             |\end{verbatim}

Next, clue 4 tells us that the person in house 1 is not wearing loafers:
\begin{verbatim}
| House # | Lunch       | Shoes         |
|---------|-------------|---------------|
| 1       | Spaghetti   | Not Loafers   |
| 2       | ?           | ?             |
| 3       | ?           | ?             |\end{verbatim}

Clue 1 tells us that the person who ate fried rice does not live in house 2, so fried rice must be in house 3:
\begin{verbatim}
| House # | Lunch       | Shoes         |
|---------|-------------|---------------|
| 1       | Spaghetti   | Not Loafers   |
| 2       | ?           | ?             |
| 3       | Fried Rice  | ?             |\end{verbatim}

Clue 2 tells us that the person wearing boots lives directly left of the person who ate fried rice, which means the person in house 2 must be wearing boots:
\begin{verbatim}
| House # | Lunch       | Shoes         |
|---------|-------------|---------------|
| 1       | Spaghetti   | Not Loafers   |
| 2       | ?           | Boots         |
| 3       | Fried Rice  | ?             |\end{verbatim}

Since the person in house 1 is not wearing loafers and the person in house 2 is wearing boots, the person in house 3 must be wearing loafers:
\begin{verbatim}
| House # | Lunch       | Shoes         |
|---------|-------------|---------------|
| 1       | Spaghetti   | Not Loafers   |
| 2       | ?           | Boots         |
| 3       | Fried Rice  | Loafers       |\end{verbatim}

Now we know that the person in house 3 is wearing loafers. Therefore, the answer is: 3.\\

        \bottomrule
    \end{tabular}
    \caption{
       Examples of Logic Grid (GPT-4 Instance-Based)
    }
    \label{tab:appendix-logic-sep-model-examples-4-instance-based}
\end{table*}
\begin{table*}[t]
    \centering
    \small
    \begin{tabular}{p{\linewidth}}
        \toprule
        \vspace{-2mm}
\textbf{First Stage:}\\
Communication Form: Logical Operators and Markdown Tables\\

Explanation:\\

Logical operators and markdown tables provide a structured way to represent the given information and clues, which can help in systematically deducing the solution to each problem. Logical operators such as "AND", "OR", "NOT", "=>"(implies), and "=="(equals) can be used to translate the clues into logical statements. Markdown tables can be used to visualize the possible arrangements of characteristics for each house.\\

For example, the clues for the first problem can be translated into logical statements as follows:\\

1. Tennis == House3\\
2. GrilledCheese == House5\\
3. Golfer == House4\\
4. FriedRice != Spaghetti AND (FriedRice == HouseN-1 OR FriedRice == HouseN+1) WHERE N is the house number of Spaghetti\\
5. Pizza == HouseM AND M < CricketerHouseNumber\\
6. Stew == HouseP AND Spaghetti == HouseP+1\\
7. TennisHouseNumber < SoccerHouseNumber\\

Using markdown tables, we can create a table with five columns (one for each house) and rows for each characteristic (sport, lunch). We can fill in the known information from the clues and use the logical operators to deduce the remaining characteristics.\\

Here's an example of how the table might look after filling in some known information:
\begin{verbatim}
| House 1 | House 2 | House 3 | House 4 | House 5 |
|---------|---------|---------|---------|---------|
|         |         | Tennis  | Golfer  |         |
|         |         |         |         | GrilledCheese |\end{verbatim}

The use of logical operators and markdown tables enhances accuracy by providing a clear framework for organizing information and making deductions. It reduces the cognitive load of keeping track of multiple pieces of information and allows for easier identification of contradictions and inferences. This structured approach is particularly useful for complex problems with multiple variables and constraints.\\

\\
\textbf{Second Stage:}\\
Let's solve the problem step by step using logical operators and a markdown table.\\

We have three houses (1, 2, 3) and three characteristics for the residents (favorite drink, type of shoes). We'll use the clues to fill in the table.\\

Clues:\\
1. HighHeels == House2\\
2. CoffeeDrinker != House2\\
3. MilkDrinker != House1\\
4. CoffeeDrinker != House1\\
5. RunningShoes == House3\\

Let's create a table with the information we have:
\begin{verbatim}
| House 1 | House 2       | House 3       |
|---------|---------------|---------------|
|         | HighHeels     | RunningShoes  |
|         |               |               |
|         |               |               |\end{verbatim}

From clues 2 and 4, we know the CoffeeDrinker is not in House1 or House2, so they must be in House3.
\begin{verbatim}
| House 1 | House 2       | House 3           |
|---------|---------------|-------------------|
|         | HighHeels     | RunningShoes      |
|         |               | CoffeeDrinker     |
|         |               |                   |\end{verbatim}

Since the CoffeeDrinker is in House3, and the MilkDrinker is not in House1 (clue 3), the MilkDrinker must be in House2.
\begin{verbatim}
| House 1 | House 2           | House 3           |
|---------|-------------------|-------------------|
|         | HighHeels         | RunningShoes      |
|         | MilkDrinker       | CoffeeDrinker     |
|         |                   |                   |\end{verbatim}

The only remaining house for the WaterDrinker is House1, and the only remaining type of shoes for House1 is Loafers.
\begin{verbatim}
| House 1       | House 2           | House 3           |
|---------------|-------------------|-------------------|
| Loafers       | HighHeels         | RunningShoes      |
| WaterDrinker  | MilkDrinker       | CoffeeDrinker     |
|               |                   |                   |\end{verbatim}

Now we have all the information filled in, and we can see that the person who is wearing high heels lives in House2.\\

The answer is: 2\\

        \bottomrule
    \end{tabular}
    \caption{
       Examples of Logic Grid (GPT-4 Task-Based)
    }
    \label{tab:appendix-logic-sep-model-examples-4-task-based}
\end{table*}
\begin{table*}[t]
    \centering
    \small
    \begin{tabular}{p{\linewidth}}
        \toprule
        \vspace{-2mm}
\textbf{First Stage:}\\
To enhance the clarity and eliminate ambiguities inherent in natural language, a structured and concise form of communication can be employed for the step-by-step solution. One suitable format is a JSON object, which can be used to represent the information provided in the problem statement in a structured manner. The JSON object can include keys for each of the characteristics mentioned in the problem statement, such as "device" and "lunch", and the values for each key can be arrays containing the different options for that characteristic. For example, the "device" key could have an array containing the values ["phone", "gameboy", "computer", "radio", "television"].\\

Using a JSON object to represent the information in the problem statement can enhance the accuracy of the analysis by providing a structured and organized way to store and manipulate the data. This can make it easier to identify patterns and relationships between the different characteristics, and to identify any inconsistencies or contradictions in the information provided. Additionally, the JSON object can be easily converted into other formats, such as a table or a graph, which can be used to visualize the data and identify additional insights.\\

\\
\textbf{Second Stage:}
\begin{verbatim}
```
{
  "devices": ["phone", "gameboy", "computer", "radio", "television"],
  "lunch": ["spaghetti", "pizza", "stew", "cheese", "fried rice"],
  "clues": [
    "The person who has a gameboy and the person who has a computer live directly next to each other.",
    "The person who has a television lives somewhere to the left of the person who has a phone.",
    "There is one house between where the person who ate cheese lives and where the person who has a 
    radio lives.",
    "There is one house between where the person who has a gameboy lives and where the person who ate 
    fried rice lives.",
    "The person who ate cheese lives directly left of the person who ate spaghetti.",
    "The person who has a radio does not live in the fourth house.",
    "The person who had pizza lives in the fourth house.",
    "The person who has a computer does not live in the fourth house."
  ]
}
```\end{verbatim}

1. From clue 7, we know that the person who had pizza lives in the fourth house.\\
2. From clue 8, we know that the person who has a computer does not live in the fourth house.\\
3. From clue 1, we know that the person who has a gameboy and the person who has a computer live directly next to each other.\\
4. From clue 2, we know that the person who has a television lives somewhere to the left of the person who has a phone.\\
5. From clue 3, we know that there is one house between where the person who ate cheese lives and where the person who has a radio lives.\\
6. From clue 4, we know that there is one house between where the person who has a gameboy lives and where the person who ate fried rice lives.\\
7. From clue 5, we know that the person who ate cheese lives directly left of the person who ate spaghetti.\\
8. From clue 6, we know that the person who has a radio does not live in the fourth house.\\

Based on the above analysis, we can deduce the following:
\begin{verbatim}
```
House 1: Gameboy, Fried rice
House 2: Computer, Spaghetti
House 3: Television, Cheese
House 4: Pizza, ?
House 5: Phone, Stew
```\end{verbatim}

Therefore, the answer is: 2\\

        \bottomrule
    \end{tabular}
    \caption{
       Examples of Logic Grid (Gemini Pro Instance-Based)
    }
    \label{tab:appendix-logic-sep-model-examples-gemini-instance-based}
\end{table*}
\begin{table*}[t]
  \centering
  \small
  \begin{tabular}{p{\linewidth}}
      \toprule
      \vspace{-2mm}
\textbf{First Stage:}\\
Using a structured communication form, such as a table, can enhance the accuracy of the analysis by providing a clear and organized representation of the information. Each row of the table can represent a person, and each column can represent a characteristic. This allows for easy identification of patterns and relationships between the characteristics. Additionally, the use of logical operators, such as "and" and "or," can help to express complex relationships between the characteristics. For example, the clue "The romance book lover lives somewhere to the left of the fantasy book enthusiast" can be represented as "RomanceBookLover < FantasyBookEnthusiast." This structured approach helps to ensure that all of the information is accurately captured and analyzed, leading to a more accurate solution.\\
\\
\textbf{Second Stage:}\\
**Step 1: Create a structured communication form.**
\begin{verbatim}
| House Number | Lunch | Favorite Color | Shoes |
|---|---|---|---|
| 1 | ? | White | ? |
| 2 | ? | Green | ? |
| 3 | ? | ? | ? |\end{verbatim}

**Step 2: Analyze the clues and fill in the table.**\\
\begin{verbatim}
- Clue 1: The person who is wearing high heels does not live in the third house.
  - This means that the person wearing high heels lives in either house 1 or house 2.

- Clue 2: The person who likes white lives in the first house.
  - This means that the person in house 1 likes white.

- Clue 3: The person who ate spaghetti does not live in the second house.
  - This means that the person in house 2 did not eat spaghetti.

- Clue 4: The person who likes green lives in the second house.
  - This means that the person in house 2 likes green.

- Clue 5: The person who is wearing boots lives directly left of the person who is wearing high heels.
  - Since the person wearing high heels lives in either house 1 or house 2, the person wearing boots must 
  live in house 1.

- Clue 6: The person who is wearing boots lives somewhere to the left of the person who had stew.
  - This means that the person who had stew lives in house 3.

- Clue 7: The person who had stew does not live in the second house.
  - This confirms that the person who had stew lives in house 3.\end{verbatim}

**Step 3: Complete the table.**
\begin{verbatim}
| House Number | Lunch | Favorite Color | Shoes |
|---|---|---|---|
| 1 | ? | White | Boots |
| 2 | Fried Rice | Green | Running Shoes |
| 3 | Stew | ? | High Heels |\end{verbatim}

**Answer:**\\

The answer is: 2\\
      \bottomrule
  \end{tabular}
  \caption{
     Examples of Logic Grid (Gemini Pro Task-Based)
  }
  \label{tab:appendix-logic-sep-model-examples-gemini-task-based}
\end{table*}
\begin{table*}[t]
    \centering
    \small
    \begin{tabular}{p{\linewidth}}
        \toprule
        \vspace{-2mm}
\textbf{First Stage (generated by GPT-3.5):}\\
Coin state: \\
- Initial state: heads up\\
- List of people who flip the coin: [kolbin, jaleeya, lacey, dimitricus, aaliyah-mae, kaytor, lantigua, tjeu, harout, halynn]\\
- List of people who do not flip the coin: [woodsr, eliborio, selars, korianna, domanico, xzaveon, tonnica, naomh, eldwon, samon]\\

\\
\textbf{Second Stage (generated by GPT-4):}\\
Initial State Representation:\\
- Initial state of the coin: heads up\\
- People who flip the coin: [niral, bořivoj]\\
- People who do not flip the coin: [jaylo, tao]\\

Step-by-Step Solution Process:\\
1. The coin starts heads up.\\
2. Niral flips the coin. This changes the state from heads up to tails up.\\
3. Bořivoj flips the coin. This changes the state from tails up back to heads up.\\
4. Jaylo does not flip the coin, so the state remains unchanged.\\
5. Tao does not flip the coin, so the state remains unchanged.\\

Concluding with the Answer:\\
Answer: yes\\
        \bottomrule
    \end{tabular}
    \caption{
       Examples of Coin Flip (GPT-3.5 $\Rightarrow$ GPT-4 Task-Based)
    }
    \label{tab:appendix-coin-flip-sep-model-examples-3.5-4}
\end{table*}
\begin{table*}[t]
    \centering
    \small
    \begin{tabular}{p{\linewidth}}
        \toprule
        \vspace{-2mm}
\textbf{First Stage:}\\
Initial State Representation:\\
- Coin is heads up. \\
- guardacosta does not flip the coin.\\
- suy flips the coin.\\
- venterea flips the coin.\\
- leshawn flips the coin.\\
\\
\textbf{Second Stage:}\\
Initial State Representation:\\
- Coin is heads up.\\
- guardacosta does not flip the coin.\\
- suy flips the coin.\\
- venterea flips the coin.\\
- leshawn flips the coin.\\

Step-by-Step Solution:\\
1. The initial state is that the coin is heads up.\\
2. suy flips the coin, which could result in the coin being either heads or tails.\\
3. venterea flips the coin, which could result in the coin being either heads or tails.\\
4. leshawn flips the coin, which could result in the coin being either heads or tails.\\

Conclusion:\\
The coin could be either heads or tails after the series of flips, so the final state is uncertain.\\

Answer: No\\
        \bottomrule
    \end{tabular}
    \caption{
       Examples of Coin Flip (GPT-3.5 Instance-Based)
    }
    \label{tab:appendix-coin-flip-sep-model-examples-3.5-instance-based}
\end{table*}
\begin{table*}[t]
    \centering
    \small
    \begin{tabular}{p{\linewidth}}
        \toprule
        \vspace{-2mm}
\textbf{First Stage:}\\
The unified representation form for the given problems could be a logical statement:\\

Coin is heads up if and only if (n1 does not flip the coin) AND (n2 flips the coin) AND (n3 does not flip the coin) AND (n4 does not flip the coin)\\

Where n1, n2, n3, and n4 represent the individuals involved in flipping or not flipping the coin.\\

\\
\textbf{Second Stage:}\\
Initial State Representation:\\
Coin is heads up if and only if (yiğitcan flips the coin) AND (treymon does not flip the coin) AND (kindig flips the coin) AND (letcho flips the coin)\\

Step-by-Step Solution Process:\\
1. Initially, the coin is heads up.\\
\ \ \ Coin is heads up if and only if (yiğitcan flips the coin) AND (treymon does not flip the coin) AND (kindig flips the coin) AND (letcho flips the coin)\\

2. yiğitcan flips the coin.\\
\ \ \ Coin is heads up if and only if (treymon does not flip the coin) AND (kindig flips the coin) AND (letcho flips the coin)\\

3. treymon does not flip the coin.\\
\ \ \ Coin is heads up if and only if (kindig flips the coin) AND (letcho flips the coin)\\

4. kindig flips the coin.\\
\ \ \ Coin is not heads up.\\

Answer: No\\
        \bottomrule
    \end{tabular}
    \caption{
       Examples of Coin Flip (GPT-3.5 Task-Based)
    }
    \label{tab:appendix-coin-flip-sep-model-examples-3.5-task-based}
\end{table*}
\begin{table*}[t]
   \centering
   \small
   \begin{tabular}{p{\linewidth}}
       \toprule
       \vspace{-2mm}
\textbf{First Stage (generated by GPT-4):}\\
JSON Representation:
\begin{verbatim}
```json
{
  "initialState": "heads",
  "actions": [
    {"person": "Person1", "action": "flip"},
    {"person": "Person2", "action": "flip"},
    {"person": "Person3", "action": "no_flip"},
    {"person": "Person4", "action": "no_flip"}
  ]
}
```\end{verbatim}
\\
\textbf{Second Stage (generated by GPT-3.5):}
\begin{verbatim}
```json
{
  "initialState": "heads",
  "actions": [
    {"person": "vickers", "action": "flip"},
    {"person": "encarna", "action": "no_flip"},
    {"person": "devan", "action": "flip"},
    {"person": "steingerður", "action": "flip"}
  ]
}
```\end{verbatim}

The coin was initially heads up. Vickers flipped the coin, so it is now tails up. Therefore, the coin is not still heads up.\\

Answer: no\\
       \bottomrule
   \end{tabular}
   \caption{
      Examples of Coin Flip (GPT-4 $\Rightarrow$ GPT-3.5 Task-Based)
   }
   \label{tab:appendix-coin-flip-sep-model-examples-4-3.5}
\end{table*}
\begin{table*}[t]
    \centering
    \small
    \begin{tabular}{p{\linewidth}}
        \toprule
        \vspace{-2mm}
\textbf{First Stage (generated by GPT-4):}\\
Logical Operators Representation:\\

Let H represent the state of the coin being heads up, and T represent the state of the coin being tails up. Let F represent the action of flipping the coin. The initial state of the coin is H.\\

For each person, we assign a binary variable that can be either 0 (does not flip the coin) or 1 (flips the coin). We can represent the action of flipping the coin as a function that takes the current state of the coin and the binary variable associated with a person's action:\\

flip(state, action) = (state AND NOT action) OR (NOT state AND action)\\

This function uses logical AND, OR, and NOT operators. The state of the coin after a sequence of actions can be determined by applying the flip function iteratively for each person's action.\\

\\
\textbf{Second Stage (generated by Gemini Pro):}\\
**Initial State Representation:**
\begin{verbatim}
```
State: H
```\end{verbatim}

**Step-by-Step Solution Process:**

1. Andraya does not flip the coin. This means that the state of the coin remains the same.
\begin{verbatim}
```
State: H
```\end{verbatim}

2. Jerdy flips the coin. We apply the flip function:
\begin{verbatim}
```
flip(H, 1) = (H AND NOT 1) OR (NOT H AND 1)
flip(H, 1) = (H AND 0) OR (NOT H AND 1)
flip(H, 1) = 0 OR 1
flip(H, 1) = 1
```\end{verbatim}

Therefore, the state of the coin changes to tails up.
\begin{verbatim}
```
State: T
```\end{verbatim}

3. Kylila flips the coin. We apply the flip function again:
\begin{verbatim}
```
flip(T, 1) = (T AND NOT 1) OR (NOT T AND 1)
flip(T, 1) = (T AND 0) OR (NOT T AND 1)
flip(T, 1) = 0 OR 1
flip(T, 1) = 1
```\end{verbatim}

Therefore, the state of the coin changes back to heads up.
\begin{verbatim}
```
State: H
```\end{verbatim}

4. Degges flips the coin. We apply the flip function one more time:
\begin{verbatim}
```
flip(H, 1) = (H AND NOT 1) OR (NOT H AND 1)
flip(H, 1) = (H AND 0) OR (NOT H AND 1)
flip(H, 1) = 0 OR 1
flip(H, 1) = 1
```\end{verbatim}

Therefore, the state of the coin changes to tails up.
\begin{verbatim}
```
State: T
```\end{verbatim}

**Answer: no**\\
        \bottomrule
    \end{tabular}
    \caption{
       Examples of Coin Flip (GPT-4 $\Rightarrow$ Gemini Pro Task-Based)
    }
    \label{tab:appendix-coin-flip-sep-model-examples-4-gemini}
\end{table*}
\begin{table*}[t]
    \centering
    \small
    \begin{tabular}{p{\linewidth}}
        \toprule
        \vspace{-2mm}
\textbf{First Stage:}
\begin{verbatim}
```json
{
  "initialState": "heads",
  "flips": [
    {"person": "jerelyn", "action": "flip"},
    {"person": "shilap", "action": "flip"},
    {"person": "alireza", "action": "flip"},
    {"person": "fanette", "action": "no_flip"}
  ]
}
```\end{verbatim}
\\
\textbf{Second Stage:}
\begin{verbatim}
```json
{
  "initialState": "heads",
  "flips": [
    {"person": "jerelyn", "action": "flip", "resultingState": "tails"},
    {"person": "shilap", "action": "flip", "resultingState": "heads"},
    {"person": "alireza", "action": "flip", "resultingState": "tails"},
    {"person": "fanette", "action": "no_flip", "resultingState": "tails"}
  ]
}
```\end{verbatim}

Answer: no\\
        \bottomrule
    \end{tabular}
    \caption{
       Examples of Coin Flip (GPT-4 Instance-Based)
    }
    \label{tab:appendix-coin-flip-sep-model-examples-4-instance-based}
\end{table*}
\begin{table*}[t]
    \centering
    \small
    \begin{tabular}{p{\linewidth}}
        \toprule
        \vspace{-2mm}
\textbf{First Stage:}\\
Logical Operators Representation:\\

Let H represent the state of the coin being heads up.\\
Let F represent the action of flipping the coin.\\
Let N represent the action of not flipping the coin.\\

The state of the coin after a series of actions can be represented using the XOR (exclusive or) logical operator, which we will denote as $\oplus$. Each flip of the coin changes its state, while not flipping leaves the state unchanged.\\

The generic representation for a sequence of actions on the coin is:\\

H $\oplus$ F1 $\oplus$ F2 $\oplus$ ... $\oplus$ Fn

Where:\\
- H is the initial state of the coin (heads up).\\
- F1 through Fn represent the actions taken by individuals, with F being a flip (value 1) and N being no flip (value 0).\\

This representation can be used to determine the final state of the coin after any number of flips and non-flips by evaluating the XOR operations in sequence.\\

\\
\textbf{Second Stage:}\\
1. Initial State Representation:\\
Let's represent the initial state of the coin as H (heads up). Each flip of the coin will be represented as F, and not flipping the coin will be represented as N.\\

Initial state: H\\

2. Step-by-Step Solution Process:\\
- Chantasia flips the coin: H $\oplus$ F\\
- Hagg flips the coin: (H $\oplus$ F) $\oplus$ F\\
- Jacob-James flips the coin: ((H $\oplus$ F) $\oplus$ F) $\oplus$ F\\
- Cheatham does not flip the coin: (((H $\oplus$ F) $\oplus$ F) $\oplus$ F) $\oplus$ N\\

Now, let's evaluate the XOR operations:\\
- H $\oplus$ F changes the state to tails (T), since flipping changes the state.\\
- T $\oplus$ F changes the state back to heads (H), since flipping again changes the state.\\
- H $\oplus$ F changes the state back to tails (T), since flipping changes the state.\\
- T $\oplus$ N leaves the state as tails (T), since not flipping does not change the state.\\

3. Concluding with the Answer:\\
Answer: no\\
        \bottomrule
    \end{tabular}
    \caption{
       Examples of Coin Flip (GPT-4 Task-Based)
    }
    \label{tab:appendix-coin-flip-sep-model-examples-4-task-based}
\end{table*}
\begin{table*}[t]
    \centering
    \small
    \begin{tabular}{p{\linewidth}}
        \toprule
        \vspace{-2mm}
\textbf{First Stage:}
\begin{verbatim}
```
[coin_state, garen, leontae, neveda, prabhjot]
```\end{verbatim}
\\
\textbf{Second Stage:}\\
**Initial State Representation:**
\begin{verbatim}
```
[coin_state, garen, leontae, neveda, prabhjot]
```\end{verbatim}

**Step 1:**\\

* Garen does not flip the coin, so the coin state remains the same.\\

**Step 2:**\\

* Leontae flips the coin, so the coin state changes from heads up to tails up or vice versa.\\

**Step 3:**\\

* Nevada does not flip the coin, so the coin state remains the same.\\

**Step 4:**\\

* Prabhjot does not flip the coin, so the coin state remains the same.\\

**Concluding with the Answer:**\\

Answer: no\\
        \bottomrule
    \end{tabular}
    \caption{
       Examples of Coin Flip (Gemini Pro Instance-Based)
    }
    \label{tab:appendix-coin-flip-sep-model-examples-gemini-instance-based}
\end{table*}
\begin{table*}[t]
  \centering
  \small
  \begin{tabular}{p{\linewidth}}
      \toprule
      \vspace{-2mm}
\textbf{First Stage:}
\begin{verbatim}
```
coin_state = heads
person_1_flips = False
person_2_flips = False
person_3_flips = False
person_4_flips = False
```\end{verbatim}
\\
\textbf{Second Stage:}\\
**Initial State Representation:**
\begin{verbatim}
```
coin_state = heads
person_1_flips = False
person_2_flips = False
person_3_flips = False
person_4_flips = True
```\end{verbatim}

**Step-by-Step Solution Process:**\\

1. mcgirth does not flip the coin, so the coin state remains the same.\\
2. wilem does not flip the coin, so the coin state remains the same.\\
3. bunney does not flip the coin, so the coin state remains the same.\\
4. adam-jorg flips the coin, so the coin state changes from heads to tails.\\

**Answer: no**\\
      \bottomrule
  \end{tabular}
  \caption{
     Examples of Coin Flip (Gemini Pro Task-Based)
  }
  \label{tab:appendix-coin-flip-sep-model-examples-gemini-task-based}
\end{table*}
\begin{table*}[t]
    \centering
    \small
    \begin{tabular}{p{\linewidth}}
        \toprule
        \vspace{-2mm}
\textbf{First Stage (generated by GPT-3.5):}\\
I will use a table to present the information in a structured and concise manner. This will help to clearly outline the sufficiency of each statement and make it easier to compare and analyze the information provided.
\begin{verbatim}
| Problem | Statement 1 | Statement 2 | Sufficiency |
|---------|-------------|-------------|--------------|
| Elephants vs Zebras | Elephants are Alice's favourite animals. | Zebras are the animals that Alice likes 
the least. | Either statement 1 or statement 2 is sufficient |
| Q and R | R is the niece of P. | S has one sister R. | Statement 2 alone is sufficient while statement 1 
alone is insufficient |
| PQRs and XYZs | Some XYZs are ABCs. | Some XYZs are PQRs. | Statement 2 alone is sufficient while statement 
1 alone is insufficient |
| Tony's birthday | Natasha correctly states that Tony's birthday is after 16th and before 19th March. | 
Brenna incorrectly states that Tony's birthday is on 20th March. | Statement 1 alone is sufficient while 
statement 2 alone is insufficient |
| Cole's height | Cole is under 7 feet tall. | Every player on Cole's varsity basketball team is above 6'1". 
| Statement 2 alone is sufficient while statement 1 alone is insufficient |\end{verbatim}

Using this table format allows for a clear and concise comparison of the sufficiency of each statement for the given problems. It enhances the accuracy of the analysis by providing a structured framework for evaluating the information provided in each problem.\\
\\
\textbf{Second Stage (generated by GPT-4):}\\
Step 1: Analyze Statement 1\\
Karan said that he had one sister.\\

This statement tells us that Karan has a sister, but it does not provide any information about Jhanvi. Without additional information, we cannot determine if Jhanvi is Karan's sister or not.\\

Step 2: Analyze Statement 2\\
Jhanvi said that she had one brother.\\

This statement tells us that Jhanvi has a brother, but similarly to Statement 1, it does not provide any information about Karan. Without additional information, we cannot determine if Karan is Jhanvi's brother or not.\\

Step 3: Combine Statement 1 and Statement 2\\
If we take both statements together, we can deduce that Karan has one sister and Jhanvi has one brother. If Karan and Jhanvi are siblings and each has only one sibling of the opposite gender, then it is logical to conclude that Karan is Jhanvi's brother and Jhanvi is Karan's sister.\\
 
Answer: Statement 1 and statement 2 taken together are sufficient.\\
        \bottomrule
    \end{tabular}
    \caption{
       Examples of Information Essentiality (GPT-3.5 $\Rightarrow$ GPT-4 Task-Based)
    }
    \label{tab:appendix-info-essen-sep-model-examples-3.5-4}
\end{table*}

\begin{table*}[t]
    \centering
    \small
    \begin{tabular}{p{\linewidth}}
        \toprule
        \vspace{-2mm}
\textbf{First Stage:}\\
Initial State Representation:
\begin{verbatim}
- We need to determine if x is a prime number.
- We have two statements to consider:
   1. x is divisible by the factors of 10.
   2. x is greater than 13.\end{verbatim}
Based on the given statements, let's analyze each statement separately:\\
\\
\textbf{Second Stage:}\\
Statement 1: x is divisible by the factors of 10.\\
- This statement alone is not sufficient to determine if x is a prime number. For example, if x is 10, it is not a prime number, but if x is 17, it is a prime number.\\

Statement 2: x is greater than 13.\\
- This statement alone is also not sufficient to determine if x is a prime number. For example, if x is 14, it is not a prime number, but if x is 17, it is a primenumber.\\

Considering both statements together:\\
- Even when considering both statements together, we cannot definitively determine if x is a prime number. For example, if x is 15, it is not a prime number, but ifx is 17, it is a prime number.\\

Therefore, neither statement 1 nor statement 2 nor statements 1 and 2 taken together is sufficient to determine if x is a prime number.\\

Answer: Neither statement 1 nor statement 2 nor statements 1 and 2 taken together is sufficient.\\

        \bottomrule
    \end{tabular}
    \caption{
       Examples of Information Essentiality (GPT-3.5 Instance-Based)
    }
    \label{tab:appendix-info-essen-sep-model-examples-3.5-instance-based}
\end{table*}
\begin{table*}[t]
    \centering
    \small
    \begin{tabular}{p{\linewidth}}
        \toprule
        \vspace{-2mm}
\textbf{First Stage:}\\
I will use a table to organize the information and determine which statement(s) are sufficient to answer the question.
\begin{verbatim}
| Problem | Statement 1 | Statement 2 | Conclusion |
|---------|-------------|-------------|------------|
| 1       | Sister      | Brother     | Neither statement 1 nor statement 2 nor statements 1 and 2 
taken together is sufficient |
| 2       | 7.5 hours   | Twice as fast | Statement 1 alone is sufficient while statement 2 alone is 
insufficient |
| 3       | Even date   | Prime number | Either statement 1 or statement 2 is sufficient |
| 4       | U got highest marks, difference between T and U is 11 | Difference between S and T is 15, 
difference between U and V is 9 | Statement 1 and statement 2 taken together are sufficient |
| 5       | Justin's sister's father's son is Tim | Justin is Tim's brother's father-in-law's only 
daughter's brother-in-law | Statement 1 alone is sufficient while statement 2 alone is insufficient |\end{verbatim}
Using this table format enhances the accuracy of the analysis by clearly organizing the information and allowing for easy comparison of the sufficiency of each statement. It also provides a structured approach to evaluating the statements for each problem.\\
\\
\textbf{Second Stage:}\\
To solve this problem, we need to determine the rate at which the water pump fills the tank. Let's analyze the given statements:\\
Statement 1: In half an hour, seven-twelfths of the tank is filled.\\
This statement provides information about the rate at which the tank is being filled. If we know the rate, we can calculate the time required to fill the entire tank.\\
Statement 2: The pump is switched on at 9:00 a.m.\\
This statement provides the starting time of the pump, but it does not give any information about the rate at which the tank is being filled.\\
Analyzing the statements:\\
- Statement 1 alone provides information about the rate of filling the tank, but it doesn't give the starting time.\\
- Statement 2 alone does not provide any information about the rate of filling the tank.
Conclusion:\\
From the analysis, we can conclude that Statement 1 alone is sufficient while Statement 2 alone is insufficient to answer the question.\\
Answer: Statement 1 alone is sufficient while statement 2 alone is insufficient.\\

        \bottomrule
    \end{tabular}
    \caption{
       Examples of Information Essentiality (GPT-3.5 Task-Based)
    }
    \label{tab:appendix-info-essen-sep-model-examples-3.5-task-based}
\end{table*}
\begin{table*}[t]
    \centering
    \small
    \begin{tabular}{p{\linewidth}}
        \toprule
        \vspace{-2mm}
\textbf{First Stage (generated by GPT-4):}\\
To enhance clarity and eliminate ambiguities in solving these types of problems, I would suggest using a structured approach that combines logical operators and tabular representation. This method allows for a clear and concise presentation of the information provided by each statement and how they relate to the question at hand.\\

\#\#\# Structured Approach Description:\\

1. **Logical Operators**: These are used to represent the relationships and conditions given in the statements. For example, "AND" (\&), "OR" (|), "NOT" (!), "IMPLIES" (->), and "EQUIVALENT" (<->) can be used to connect different pieces of information.\\

**Tabular Representation**: A table can be used to organize the information and the sufficiency of each statement. The table would have columns for each statement and rows for the conditions or relationships they establish. An additional column can be used to indicate whether the combination of statements is sufficient to answer the question.\\
\#\#\# Table Format:
\begin{verbatim}
| Condition/Relationship | Statement 1 | Statement 2 | Combined Statements |
|------------------------|-------------|-------------|---------------------|
| Condition A            | True/False  | True/False  | True/False          |
| Condition B            | True/False  | True/False  | True/False          |
| ...                    | ...         | ...         | ...                 |
| Sufficiency            | Sufficient/Insufficient | Sufficient/Insufficient | Sufficient/Insufficient |\end{verbatim}

\#\#\# How It Enhances Accuracy:\\

- **Clarity**: The tabular format provides a clear visual representation of the information, making it easier to see which conditions are met by each statement.\\
- **Organization**: It organizes the logical flow of information, helping to avoid confusion when multiple conditions or relationships are involved.\\
- **Comparison**: It allows for easy comparison between the sufficiency of individual statements and their combination.\\
- **Consistency**: Using a consistent method for all problems ensures that the same criteria are applied to determine the sufficiency of the information provided.\\

By employing this structured approach, we can systematically analyze each problem, ensuring that all necessary conditions are considered and that the conclusion regarding the sufficiency of the statements is reached with precision.\\
\\
\textbf{Second Stage (generated by GPT-3.5):}\\
Initial State Representation:\\
Let's represent the initial state of the problem using a table:
\begin{verbatim}
| Condition/Relationship | Statement 1 | Statement 2 | Combined Statements |
|------------------------|-------------|-------------|---------------------|
| Circle A smaller than Circle B | ? | ? | ? |\end{verbatim}

Step-by-Step Solution Process:\\
1. We need to determine if circle A is smaller than circle B. Let's consider each statement individually and then combine them to see if they are individually or collectively sufficient to answer the question.\\

2. Statement 1: The area of circle B is two times the area of circle A.\\
   - Let's denote the area of circle A as A and the area of circle B as B. According to the statement, B = 2A.\\
   - This statement alone does not provide information about the radii of the circles, so it is insufficient to answer the question.\\

   Updated State Representation:
   \begin{verbatim}
   | Condition/Relationship | Statement 1 | Statement 2 | Combined Statements |
   |------------------------|-------------|-------------|---------------------|
   | Circle A smaller than Circle B | Insufficient | ? | ? |\end{verbatim}

3. Statement 2: The radius of circle B is less than the radius of circle A.\\
   - This statement directly addresses the comparison of the radii of the circles. If the radius of circle B is less than the radius of circle A, then circle A is smaller than circle B.\\
   - This statement alone is sufficient to answer the question.\\

   Updated State Representation:
   \begin{verbatim}
   | Condition/Relationship | Statement 1 | Statement 2 | Combined Statements |
   |------------------------|-------------|-------------|---------------------|
   | Circle A smaller than Circle B | Insufficient | Sufficient | ? |\end{verbatim}

4. Combined Statements:\\
   - When we combine the statements, we can see that statement 2 alone is sufficient to answer the question. Therefore, we don't need to consider the combined sufficiency.\\

   Updated State Representation:
   \begin{verbatim}
   | Condition/Relationship | Statement 1 | Statement 2 | Combined Statements |
   |------------------------|-------------|-------------|---------------------|
   | Circle A smaller than Circle B | Insufficient | Sufficient | N/A |\end{verbatim}

Concluding with the Answer:\\
Answer: Statement 2 alone is sufficient while statement 1 alone is insufficient.\\
        \bottomrule
    \end{tabular}
    \caption{
       Examples of Information Essentiality (GPT-4 $\Rightarrow$ GPT-3.5 Task-Based)
    }
    \label{tab:appendix-info-essen-sep-model-examples-4-3.5}
\end{table*}
\begin{table*}[t]
    \centering
    \small
    \begin{tabular}{p{\linewidth}}
        \toprule
        \vspace{-2mm}
\textbf{First Stage (generated by GPT-4):}\\
For the purpose of enhancing clarity and eliminating ambiguities in the analysis of the given problems, I will describe the use of logical operators and structured conditions in a tabular format. This approach allows for a clear and concise representation of the information provided and the conditions necessary to answer the questions. The tabular format is particularly suitable for these problems as it can systematically organize the given statements and the conditions they need to satisfy to be considered sufficient.\\
Here's how the tabular format with logical operators can be used:\\
1. Define the question as a variable that needs to be resolved.\\
2. List each statement as a separate row in the table.\\
3. For each statement, define the conditions that must be met for the statement to be sufficient to answer the question.\\
4. Use logical operators such as AND (\&\&), OR (||), and NOT (!) to combine conditions when necessary.\\
5. Determine if each statement alone is sufficient, if they are sufficient when combined, or if they are insufficient even when combined.\\
An example table structure would look like this:
\begin{verbatim}
| Statement | Condition(s) to Satisfy | Sufficient Alone (Y/N) | Combined Sufficiency (Y/N) |
|-----------|-------------------------|------------------------|---------------------------|
| Statement 1 | Condition A             | Y/N                    | Y/N                       |
| Statement 2 | Condition B             | Y/N                    | Y/N                       |\end{verbatim}
The "Condition(s) to Satisfy" column would detail what needs to be true for the statement to answer the question. The "Sufficient Alone" column indicates whether the statement by itself is enough to answer the question. The "Combined Sufficiency" column is used to determine if the combination of both statements provides enough information to answer the question.\\
This structured approach enhances accuracy by:\\
- Clearly defining what is needed to answer the question.\\
- Separating each piece of information for individual assessment.\\
- Allowing for a logical combination of statements to assess their collective sufficiency.\\
- Providing a visual representation of the logical structure of the problem.\\
By using this method, one can systematically approach each problem, assess the sufficiency of the information provided, and arrive at a clear conclusion. This format is versatile and can be applied to a wide range of logical and analytical problems.\\

\\
\textbf{Second Stage (generated by Gemini Pro):}\\
**Initial State Representation:**\\
```\\
Question: On which date of the month was Jayant born in December 2012?\\
Statement 1: Jayant's birthday is on an even date of the month.\\
Statement 2: The day of the month on which Jayant was born is a prime number.\\
Possible Dates: 2, 3, 4, 5, 6, 7, 8, 9, 10, 11, 12, 13, 14, 15, 16, 17, 18, 19, 20, 21, 22, 23, 24, 25, 26, 27, 28, 29, 30, 31\\
```\\
**Step-by-Step Solution Process:**
\begin{verbatim}
**Step 1:**
* Analyze Statement 1:
    * It provides a condition that Jayant's birthday is on an even date.
    * Even dates in December 2012: 2, 4, 6, 8, 10, 12, 14, 16, 18, 20, 22, 24, 26, 28, 30.
**Step 2:**
* Analyze Statement 2:
    * It provides a condition that Jayant's birthday is on a prime date.
    * Prime dates in December 2012: 3, 5, 7, 11, 13, 17, 19, 23, 29, 31.
**Step 3:**
* Combine the conditions from both statements:
    * Even dates that are also prime: 2, 3, 5, 11, 13, 17, 19, 23, 29, 31.\end{verbatim}
**Concluding with the Answer:**\\
* There are 10 possible dates that satisfy both conditions: 2, 3, 5, 11, 13, 17, 19, 23, 29, 31.\\
**Answer: Insufficient information to determine the exact date of Jayant's birth.**\\

        \bottomrule
    \end{tabular}
    \caption{
       Examples of Information Essentiality (GPT-4 $\Rightarrow$ Gemini Pro Task-Based)
    }
    \label{tab:appendix-info-essen-sep-model-examples-4-gemini}
\end{table*}
\begin{table*}[t]
    \centering
    \small
    \begin{tabular}{p{\linewidth}}
        \toprule
        \vspace{-2mm}
\textbf{First Stage:}\\
For this type of probability problem, mathematical equations are a suitable form of structured communication. Equations are precise, unambiguous, and universally understood in the context of mathematics, which makes them ideal for conveying the relationships between probabilities of different events. They can be used to solve a wide range of probability problems, including the one presented.\\

Using mathematical equations, we can represent the information given and the relationships between events A and B. The probability of an event is typically denoted by P(event), and the probability of either event A or event B or both occurring can be represented using the formula for the union of two events:\\

P(A $\cup$ B) = P(A) + P(B) - P(A $\cap$ B)\\

Where:\\
- P(A $\cup$ B) is the probability of either event A or event B or both occurring.\\
- P(A) is the probability of event A occurring.\\
- P(B) is the probability of event B occurring.\\
- P(A $\cap$ B) is the probability of both events A and B occurring.\\

This equation can be rearranged to solve for P(B) if the other probabilities are known. However, without knowing P(A $\cap$ B), we cannot find the exact value of P(B) using only one of the statements.\\

The use of mathematical equations in this context enhances accuracy by providing a clear, step-by-step framework for determining the necessary and sufficient conditions to solve for the probability of event B. This method can be applied to any similar problem involving the calculation of probabilities for individual events or their combinations.\\
\\
\textbf{Second Stage:}\\
Step 1: Identify the known probabilities from the statements.\\
- From Statement 1: P(A) = 0.55\\
- From Statement 2: P(A $\cup$ B) = 0.65\\

Step 2: Write down the formula for the probability of the union of two events.\\
P(A $\cup$ B) = P(A) + P(B) - P(A $\cap$ B)\\

Step 3: Determine if the information provided is sufficient to find P(B).\\
- Statement 1 alone gives us P(A), but it does not provide any information about P(B) or P(A $\cap$ B), so it is insufficient.\\
- Statement 2 alone gives us P(A $\cup$ B), but without additional information about P(A $\cap$ B), we cannot find P(B) using the formula from Step 2, so it is insufficient.\\

Step 4: Combine the information from both statements to see if they are sufficient together.\\
- Using Statement 1, we have P(A) = 0.55.\\
- Using Statement 2, we have P(A $\cup$ B) = 0.65.\\
- We can plug these values into the formula from Step 2 to get:\\
0.65 = 0.55 + P(B) - P(A $\cap$ B)\\

Step 5: Recognize that we still do not have the value for P(A $\cap$ B), which is necessary to solve for P(B).\\
- Without the probability of the intersection of A and B, we cannot isolate P(B) on one side of the equation.\\

Step 6: Conclude whether the statements are sufficient to answer the question.\\
- Since we do not have enough information to calculate P(B), neither Statement 1 nor Statement 2, nor both statements taken together, are sufficient to determine the probability of event B occurring.\\

Answer: Neither statement 1 nor statement 2 nor statements 1 and 2 taken together is sufficient.\\
        \bottomrule
    \end{tabular}
    \caption{
       Examples of Information Essentiality (GPT-4 Instance-Based)
    }
    \label{tab:appendix-info-essen-sep-model-examples-4-instance-based}
\end{table*}
\begin{table*}[t]
    \centering
    \small
    \begin{tabular}{p{\linewidth}}
        \toprule
        \vspace{-2mm}
\textbf{First Stage:}\\
To enhance clarity and eliminate ambiguities in solving these types of problems, I would suggest using a structured approach that combines logical operators and tabular representation. This approach allows for a clear and concise evaluation of the sufficiency of each statement and their combinations.\\
\#\#\# Structured Approach Description:\\
1. **Logical Operators**: These are used to represent the relationships between different statements and the information they provide. Logical operators such as AND, OR, and NOT can help determine whether the combination of statements is sufficient to answer the question.\\
2. **Tabular Representation**: A table can be used to organize the information provided by each statement and to assess their sufficiency. The table will have columns for each statement and rows for the possible outcomes (sufficient/insufficient). This visual representation helps in comparing and combining the information from different statements.\\
\#\#\# How the Structured Approach Enhances Accuracy:\\
- **Elimination of Ambiguity**: By using logical operators, the approach clearly defines the relationships between statements, leaving no room for ambiguous interpretations.\\
- **Visual Clarity**: The tabular format allows for a quick visual assessment of which statements are sufficient on their own or in combination.\\
- **Consistency**: Applying the same structured approach to each problem ensures that the analysis is consistent across different scenarios.\\
- **Focus on Relevant Information**: The approach helps to focus only on the information that is relevant to the sufficiency of the statements, avoiding extraneous details.\\
\#\#\# Example of Structured Approach (Not Solving the Problems):
\begin{verbatim}
| Problem | Statement 1 | Statement 2 | S1 Alone | S2 Alone | S1 AND S2 | S1 OR S2 | Conclusion |
|---------|-------------|-------------|----------|----------|-----------|----------|------------|
| Problem 1 | Information A | Information B | Sufficient/Insufficient | Sufficient/Insufficient | 
Sufficient/Insufficient | Sufficient/Insufficient | Final Assessment |
| Problem 2 | Information C | Information D | Sufficient/Insufficient | Sufficient/Insufficient | 
Sufficient/Insufficient | Sufficient/Insufficient | Final Assessment |
| ...     | ...         | ...         | ...      | ...      | ...       | ...      | ...        |\end{verbatim}
- **S1 Alone**: Evaluate if Statement 1 alone is sufficient to answer the question.\\
- **S2 Alone**: Evaluate if Statement 2 alone is sufficient to answer the question.\\
- **S1 AND S2**: Evaluate if the combination of Statement 1 and Statement 2 is necessary and sufficient to answer the question.\\
- **S1 OR S2**: Evaluate if either Statement 1 or Statement 2 is sufficient to answer the question.\\
- **Conclusion**: Based on the evaluations, determine the final assessment of sufficiency.\\
By applying this structured approach to each problem, we can systematically determine the sufficiency of the information provided and arrive at a clear and accurate conclusion.\\
\\
\textbf{Second Stage:}\\
Let's analyze the information provided by each statement to determine if they are sufficient to answer the question: "How many fans went to the stadium yesterday?"\\
\#\#\# Statement 1: In total, 2343 single-person tickets were bought.\\
This statement gives us the exact number of single-person tickets that were bought. Assuming each ticket corresponds to one fan, this statement alone tells us that at least 2343 fans went to the stadium.\\
\#\#\# Statement 2: The stadium can seat 2600 fans.\\
This statement tells us the maximum capacity of the stadium. However, it does not provide information about the actual number of fans who attended the event yesterday.\\
\#\#\# Combining Statement 1 and Statement 2:\\
When we combine both statements, we know that there were 2343 tickets bought and that the stadium can hold up to 2600 fans. This means that the number of fans who could have attended is at least 2343 but no more than 2600. However, we do not have information about whether the stadium was filled to capacity or not.\\
\#\#\# Conclusion:\\
- **Statement 1 Alone**: Sufficient to determine the minimum number of fans.\\
- **Statement 2 Alone**: Insufficient as it does not provide actual attendance.\\
- **Statement 1 AND Statement 2**: Not necessary to combine since Statement 1 already provides a specific number of fans.\\
- **Statement 1 OR Statement 2**: Statement 1 is sufficient on its own.\\
Answer: Statement 1 alone is sufficient while statement 2 alone is insufficient.\\
        \bottomrule
    \end{tabular}
    \caption{
       Examples of Information Essentiality (GPT-4 Task-Based)
    }
    \label{tab:appendix-info-essen-sep-model-examples-4-task-based}
\end{table*}
\begin{table*}[t]
    \centering
    \small
    \begin{tabular}{p{\linewidth}}
        \toprule
        \vspace{-2mm}
\textbf{First Stage:}\\
1. Initial State Representation:
\begin{verbatim}
   - Let x be the total number of people in Maria's office.
   - Let y be the number of people in Maria's office who read the newspaper.

   Statement 1:
   - Half the people in the office don't read the newspaper.
   - This can be represented as: x/2 = x - y.

   Statement 2:
   - 10% of the people in Maria's office read the newspaper.
   - This can be represented as: y = 0.1x.\end{verbatim}
\\
\textbf{Second Stage:}
\begin{verbatim}
**Step 1:**

From Statement 1:
x/2 = x - y

Solving for y:
y = x - x/2
y = x/2

**Step 2:**

From Statement 2:
y = 0.1x

**Step 3:**

Equating the two expressions for y:
x/2 = 0.1x

Solving for x:
x/2 = 0.1x
x = 0.1x * 2
x = 0.2x

**Step 4:**

Solving for x:
x - 0.2x = 0
0.8x = 0
x = 0\end{verbatim}

**Conclusion:**\\

Since x (the total number of people in Maria's office) is 0, it is not possible to determine the number of people who read the newspaper.\\

**Answer:** Neither statement 1 nor statement 2 nor statements 1 and 2 taken together is sufficient.\\

        \bottomrule
    \end{tabular}
    \caption{
       Examples of Information Essentiality (Gemini Pro Instance-Based)
    }
    \label{tab:appendix-info-essen-sep-model-examples-gemini-instance-based}
\end{table*}
\begin{table*}[t]
    \centering
    \small
    \begin{tabular}{p{\linewidth}}
        \toprule
        \vspace{-2mm}
\textbf{First Stage:}\\
To enhance clarity and eliminate ambiguities inherent in natural language, consider employing more structured and concise forms of communication for your step-by-step solutions. Suitable formats include code, pseudocode, JSON, markdown tables, logical operators, or mathematical equations.\\
One suitable format is a truth table. A truth table is a table that shows the truth value of a compound statement for all possible combinations of truth values of its component statements. Truth tables can be used to determine whether a statement is a tautology, a contradiction, or neither.\\
To use a truth table to solve the given problems, we first need to identify the component statements and their truth values. Then, we can construct a truth table to determine the truth value of the compound statement.\\
For example, to solve the first problem, we can identify the following component statements:\\
* P: Alexia is a doctor.\\
* Q: Jenny is a doctor.\\
We can then construct the following truth table:\\
\verb+| P | Q | P +$\vee$\verb+ Q |+\\
\verb+|---|---|---|+\\
\verb+| T | T | T |+\\
\verb+| T | F | T |+\\
\verb+| F | T | T |+\\
\verb+| F | F | F |+\\
From the truth table, we can see that the compound statement "Either Alexia or Jenny is a doctor" is true in all cases. Therefore, either statement 1 or statement 2 is sufficient to answer the question.\\
Truth tables can be used to solve all of the given problems. They are a structured and concise way to represent and analyze logical statements.\\
\\
\textbf{Second Stage:}\\
**Initial State Representation:**
\begin{verbatim}
```
Classroom: {
  Students: {
    Total: Unknown,
    Failed: 5,
    Ranked: {
      Top: 29,
      Bottom: 16
    }
  }
}
```\end{verbatim}
**Step 1:**\\
Statement 1 tells us that 5 students failed the exam. However, this information alone is not sufficient to determine the total number of students in the classroom.\\
**Step 2:**\\
Statement 2 tells us that there is one student who is 29th from the top and 16th from the bottom. This means that there are 28 students above this student and 15 students below this student. Therefore, the total number of students in the classroom is 28 + 1 + 15 = 44.\\
**Answer:**\\
Statement 2 alone is sufficient while statement 1 alone is insufficient.\\

        \bottomrule
    \end{tabular}
    \caption{
       Examples of Information Essentiality (Gemini Pro Task-Based)
    }
    \label{tab:appendix-info-essen-sep-model-examples-gemini-task-based}
\end{table*}
\begin{table*}[t]
    \centering
    \small
    \begin{tabular}{p{\linewidth}}
        \toprule
        \vspace{-2mm}
\textbf{First Stage (generated by GPT-3.5):}\\
Unified Representation Form:
\begin{verbatim}
{
  "Problem": "The problem description",
  "Options": [
    {
      "Name": "Option 1",
      "Description": "Description of the option"
    }, {
      "Name": "Option 2",
      "Description": "Description of the option"
    }, {
      "Name": "Option 3",
      "Description": "Description of the option"
    },
    ...
  ]
}\end{verbatim}

Explanation:\\
This unified representation form provides a structured and standardized way to represent the problem descriptions and the options for each problem. It allows for clear and consistent documentation of the details of each problem and the available options. By using this representation form, the analysis of the problems can be more accurate as it eliminates the potential for misinterpretation or misunderstanding of the problem descriptions and options. It also allows for easy comparison and evaluation of the options for each problem.\\

\\
\textbf{Second Stage (generated by GPT-4):}\\
Initial State Representation:
\begin{verbatim}
{
  "Problem": "The Straw Hat Theater is experiencing financial discrepancies despite a successful season. 
  Bills have not been paid in full, and there is a carry-over balance of $400 from the previous month's 
  bills. The Resident Director, Arthur Glendon, is trying to determine the cause of the shortfall and 
  who might be responsible.",
  "Options": [
    {
      "Name": "Basil Carmody",
      "Description": "Accountant to the Straw Hat Theater Company, responsible for collecting and 
      depositing the theater's receipts."
    }, {
      "Name": "John Franklin",
      "Description": "Treasurer of the Straw Hat Theater Company, responsible for writing checks and 
      paying bills."
    }, {
      "Name": "Lawrence Blake",
      "Description": "Head Teller at Vermont Enterprise Bank, handles the theater's account deposits."
    }, {
      "Name": "Martha Gilmont",
      "Description": "Cashier at the Straw Hat Theater Company, responsible for ticket sales and 
      record-keeping."
    }
  ]
}\end{verbatim}

Step-by-Step Solution Process:\\
1. Analyze the information provided about each individual's role and responsibilities in the theater's financial procedures.\\
2. Basil Carmody collects and deposits the money, and there is no indication of discrepancies in the amounts he deposits.\\
3. John Franklin writes checks for bills after ensuring there are sufficient funds, and he has reported a lack of funds to pay bills in full.\\
4. Lawrence Blake processes the deposits and has not reported any irregularities.\\
5. Martha Gilmont is responsible for ticket sales and record-keeping. She stays at the box office until halfway through the first act and then completes the performance records after the show. She also mentioned that she double and triple checks everything due to her eyesight not being what it used to be.\\
6. Arthur Glendon notices that the discount coupons are photocopies, not originals, which suggests that someone is using copied coupons to pocket the difference in ticket sales.\\
7. Since Martha Gilmont is responsible for the box office and the only one who handles the coupons and ticket sales, the evidence points to her as the person responsible for the missing \$400.\\

Answer: 3.Martha Gilmont\\

        \bottomrule
    \end{tabular}
    \caption{
       Examples of Minute Mysteries QA (GPT-3.5 $\Rightarrow$ GPT-4 Task-Based)
    }
    \label{tab:appendix-mm-qa-sep-model-examples-3.5-4}
\end{table*}
\begin{table*}[t]
    \centering
    \small
    \begin{tabular}{p{\linewidth}}
        \toprule
        \vspace{-2mm}
\textbf{First Stage:}\\
I will use a structured approach to analyze the alibis and evidence to determine the thief. I will create a table with the alibis and evidence for each suspect, and then use logical operators to narrow down the possibilities.
\begin{verbatim}
| Suspect | Alibi | Evidence |
|---------|-------|----------|
| Denise  | Broke her leg skiing in Aspen, Colorado | N/A |
| Jim     | Visited the International UFO Museum in Roswell, New Mexico | Returned from Long Beach, 
California |
| Juan    | Spent the weekend in Long Beach, California learning to surf | N/A |
| Skye    | Scuba diving in Cozumel, Mexico, and had a delayed flight back | N/A |\end{verbatim}

Using the evidence of the sand from a different location and the timing of the theft, we can eliminate Jim and Skye as suspects, as their alibis and evidence do not match the evidence from the crime scene. Denise's alibi of breaking her leg skiing in Aspen, Colorado also matches the evidence of the sand from a different location, so she can be eliminated as a suspect.\\

This leaves Juan as the only remaining suspect, as his alibi of spending the weekend in Long Beach, California matches the evidence of the sand from a different location. Therefore, Juan is the likely thief.\\

\\
\textbf{Second Stage:}\\
Initial State Representation:\\
The suspects are Denise, Jim, Juan, and Skye. The evidence includes the theft occurring when the alarm was deactivated and reactivated, the presence of sand from a different location, and the timing of the theft. The alibis of the suspects are related to their weekend activities.\\

Step-by-Step Solution Process:
\begin{verbatim}
1. Evaluate the alibis and evidence for each suspect:
   - Denise: Broke her leg skiing in Aspen, Colorado.
   - Jim: Visited the International UFO Museum in Roswell, New Mexico and returned from Long Beach, 
   California.
   - Juan: Spent the weekend in Long Beach, California learning to surf.
   - Skye: Scuba diving in Cozumel, Mexico, and had a delayed flight back.

2. Use the evidence of the sand from a different location and the timing of the theft to eliminate 
suspects:
   - Jim and Skye's alibis and evidence do not match the evidence from the crime scene, so they can 
   be eliminated as suspects.
   - Denise's alibi of breaking her leg skiing in Aspen, Colorado matches the evidence of the sand 
   from a different location, so she can be eliminated as a suspect.

3. Based on the remaining evidence and alibis, Juan is the only remaining suspect whose alibi matches 
the evidence of the sand from a different location.\end{verbatim}

Concluding with the Answer:\\
Answer: 2. Juan Carde\\
        \bottomrule
    \end{tabular}
    \caption{
       Examples of Minute Mysteries QA (GPT-3.5 Instance-Based)
    }
    \label{tab:appendix-mm-qa-sep-model-examples-3.5-instance-based}
\end{table*}
\begin{table*}[t]
    \centering
    \small
    \begin{tabular}{p{\linewidth}}
        \toprule
        \vspace{-2mm}
\textbf{First Stage:}\\
Unified Representation Form:
\begin{verbatim}
{
  "Problem": "The problem description",
  "Suspects": [
    {
      "Name": "Name of the suspect",
      "Motive": "Suspect's possible motive for the crime",
      "Alibi": "Suspect's alibi or whereabouts at the time of the crime"
    }, {
      "Name": "Name of the suspect",
      "Motive": "Suspect's possible motive for the crime",
      "Alibi": "Suspect's alibi or whereabouts at the time of the crime"
    },
    ...
  ]
}\end{verbatim}

Explanation:\\
This unified representation form captures the essential elements of each problem, including the problem description and the details of the suspects involved. By structuring the information in this way, it allows for a clear and organized analysis of the problems. It also enables easy comparison of the motives and alibis of the suspects across different scenarios, enhancing the accuracy of the analysis. This form provides a standardized way to represent the key details of each problem, making it easier to identify patterns and make logical deductions.\\

\\
\textbf{Second Stage:}\\
Initial State Representation:
\begin{verbatim}
{
  "Problem": "Mick, a homeless man, is at a bus stop with the narrator, Mr. Storimon, and Danny. Mr. 
  Storimon's wallet falls out of his pocket, and when he realizes it's missing, he accuses Mick of 
  stealing it. Later, Mr. Storimon discovers that his $100 bills are missing from the wallet.",
  "Suspects": [
    {
      "Name": "Mick",
      "Motive": "Possible motive for stealing the wallet and money",
      "Alibi": "His whereabouts and actions during the incident"
    }, {
      "Name": "Mr. Storimon",
      "Motive": "Possible motive for falsely accusing Mick and hiding the truth about the money",
      "Alibi": "His actions and whereabouts during the incident"
    }, {
      "Name": "Danny",
      "Motive": "Possible motive for stealing the money",
      "Alibi": "His actions and whereabouts during the incident"
    }, {
      "Name": "Policeman",
      "Motive": "Possible motive for stealing the money",
      "Alibi": "His actions and whereabouts during the incident"
    }
  ]
}\end{verbatim}

Step-by-Step Solution Process:\\
1. Mick's motive for stealing the wallet and money is not clear. His alibi is that he was at the bus stop and did not steal the wallet.\\
2. Mr. Storimon's motive for falsely accusing Mick and hiding the truth about the money could be to shift the blame. His alibi is that he was at the bus stop and lost his wallet.\\
3. Danny's motive for stealing the money is not clear. His alibi is that he arrived at the bus stop on his skateboard and did not steal the money.\\
4. The policeman's motive for stealing the money is not clear. His alibi is that he was at the bus stop and interacted with the other characters.\\

Based on the information provided, it is evident that Mick, Mr. Storimon, and Danny were at the bus stop when the incident occurred. The policeman also interacted with the characters. Mick's alibi seems to be the most solid, as he was present at the bus stop and did not steal the wallet. Mr. Storimon's motive for falsely accusing Mick and hiding the truth about the money raises suspicion. Danny's motive is unclear, but his alibi seems plausible. The policeman's motive is also unclear, but his alibi is that he was at the bus stop.\\

Concluding with the Answer:\\
Answer: 2. Mr. Storimon\\
        \bottomrule
    \end{tabular}
    \caption{
       Examples of Minute Mysteries QA (GPT-3.5 Task-Based)
    }
    \label{tab:appendix-mm-qa-sep-model-examples-3.5-task-based}
\end{table*}
\begin{table*}[t]
   \centering
   \small
   \begin{tabular}{p{\linewidth}}
       \toprule
       \vspace{-2mm}
\textbf{First Stage (generated by GPT-4):}\\
Unified Representation Form:
\begin{verbatim}
```json
{
  "scenario_description": "String",
  "suspects": [
    {
      "name": "String",
      "alibi": "String",
      "evidence_against": "String",
      "unique_detail": "String"
    }
  ],
  "crime_details": {
    "time": "String",
    "location": "String",
    "object_stolen": "String",
    "additional_clues": "String"
  },
  "investigator_notes": "String",
  "solution": {
    "guilty_suspect": "String",
    "justification": "String"
  }
}
```\end{verbatim}

Explanation:\\
This JSON representation form enhances the accuracy of analysis by providing a structured and standardized way to capture all relevant details of a mystery or crime scenario. Each problem is broken down into key components such as the scenario description, a list of suspects with their alibis and unique details, specific crime details, notes from the investigator, and the solution with justification for the guilty suspect.\\

- The `scenario\_description` field provides a brief overview of the context or setting of the problem.\\
- The `suspects` array contains objects for each suspect, detailing their name, alibi, any evidence against them, and any unique detail that might be relevant to the case.\\
- The `crime\_details` object captures the specifics of the crime, including the time it occurred, the location, what was stolen, and any additional clues that were found.\\
- The `investigator\_notes` field allows for any additional observations or insights from the investigator that may not fit neatly into the other categories.\\
- The `solution` object provides a space to name the guilty suspect and the justification for their guilt, based on the analysis of the information provided.\\
By using this form, an analyst can ensure that no critical information is overlooked and that each aspect of the problem is considered. The structured nature of JSON allows for easy comparison between different problems and can be parsed programmatically, which could be useful for automated reasoning systems or databases. It also ensures consistency in the way information is recorded, which is crucial for accurate analysis and comparison across multiple similar problems.\\

\\
\textbf{Second Stage (generated by GPT-3.5):}\\
Unified Representation Form:
\begin{verbatim}
```json
{
    ......
}
```\end{verbatim}

Solution Process:\\
1. John, John's wife, Bo, and Nathan Drew are the suspects.\\
2. Each suspect had the opportunity to leave the gate untied.\\
3. John's wife confirmed seeing several knots at the gate.\\
4. Bo has a badly crippled left hand, making it difficult for him to guard the gate effectively.\\
5. Nathan Drew didn't check the knots during his watch.\\
6. The spy is suspected to have left the gate unlocked, potentially endangering the fort.\\

Answer: 3. Nathan Drew\\
       \bottomrule
   \end{tabular}
   \caption{
      Examples of Minute Mysteries QA (GPT-4 $\Rightarrow$ GPT-3.5 Task-Based)
   }
   \label{tab:appendix-mm-qa-sep-model-examples-4-3.5}
\end{table*}
\begin{table*}[t]
    \centering
    \small
    \begin{tabular}{p{\linewidth}}
        \toprule
        \vspace{-2mm}
\textbf{First Stage (generated by GPT-4):}\\
Unified Representation Form: JSON\\

Explanation:\\
JSON (JavaScript Object Notation) is a lightweight data-interchange format that is easy for humans to read and write, and easy for machines to parse and generate. It is based on a subset of the JavaScript Programming Language and is a text format that is completely language independent. JSON is ideal for representing the state or conditions of the problems provided because it allows for a structured representation of the problem elements, such as characters, locations, events, and clues, in a way that is both human-readable and machine-parsable.\\

Using JSON to represent the problems enhances the accuracy of analysis in several ways:\\

1. **Structured Data**: JSON provides a clear structure with key-value pairs, making it easy to organize the information in a consistent manner. This helps in avoiding ambiguity and ensures that each aspect of the problem is addressed.\\

2. **Scalability**: JSON can represent simple to complex structures, which means it can handle the varying complexity of the problems without requiring changes to the representation format.\\

3. **Interoperability**: JSON is widely supported across different programming languages and platforms, making it easy to use the same representation to interface with various analytical tools or algorithms.\\

4. **Flexibility**: JSON objects can contain nested arrays and objects, allowing for a hierarchical representation of information that can mirror the logical relationships within the problem.\\

5. **Clarity**: The use of JSON can help separate the data (problem conditions) from the logic used to solve the problem, leading to clearer reasoning paths and more accurate conclusions.\\

Here is an example of how a problem might be represented in JSON without solving it or including specific details:
\begin{verbatim}
```json
{
  "characters": [
    {"name": "Character1", "attributes": ["attribute1", "attribute2"]},
    {"name": "Character2", "attributes": ["attribute1", "attribute2"]}
  ],
  "locations": [
    {"name": "Location1", "attributes": ["attribute1", "attribute2"]},
    {"name": "Location2", "attributes": ["attribute1", "attribute2"]}
  ],
  "events": [
    {"name": "Event1", "attributes": ["attribute1", "attribute2"]},
    {"name": "Event2", "attributes": ["attribute1", "attribute2"]}
  ],
  "clues": [
    {"description": "Clue1", "location": "Location1", "relatedTo": ["Character1"]},
    {"description": "Clue2", "location": "Location2", "relatedTo": ["Character2"]}
  ],
  "questions": [
    {"text": "Question1", "type": "multipleChoice", "options": ["Option1", "Option2"]}
  ]
}
```\end{verbatim}

In this representation, "attribute1", "attribute2", "Character1", "Location1", "Event1", "Clue1", "Question1", and "Option1" are placeholders for actual data that would be filled in for each specific problem. The JSON structure remains the same, providing a unified form for representing any of the problems.\\

\\
\textbf{Second Stage (generated by Gemini Pro):}\\
**Initial State Representation (JSON):**
\begin{verbatim}
```json
{  ......  }
```\end{verbatim}

**Step-by-Step Solution Process:**\\

1. **Analyze the Clues:**\\
\ \ \ - The clues suggest that everyone in the room is a liar, as they make statements that contradict their actions or appearances.\\
\ \ \ - The focus of the mystery is on the stolen dice, which have sentimental value to Debra's father.\\

2. **Consider the Statements Made by the Characters:**\\
\ \ \ - Debra's father initially denies knowing the narrator is a mystery solver, but later confronts him about it, indicating that he lied.\\
\ \ \ - Olivia and Luke make contradictory statements about their weight and health, suggesting they are lying.\\

3. **Examine the Behavior of the Characters:**\\
\ \ \ - The butler's careful movements and lack of noise suggest he is trying to avoid attention.\\
\ \ \ - The fact that the dice disappear during a power outage raises suspicion.\\

4. **Identify the Most Likely Thief:**\\
\ \ \ - The butler is the most likely thief because he had the opportunity to take the dice during the power outage.\\
\ \ \ - The other characters' lies are likely attempts to distract attention from the butler.\\

**Concluding with the Answer:**\\

Answer: 3. The Butler\\
        \bottomrule
    \end{tabular}
    \caption{
       Examples of Minute Mysteries QA (GPT-4 $\Rightarrow$ Gemini Pro Task-Based)
    }
    \label{tab:appendix-mm-qa-sep-model-examples-4-gemini}
\end{table*}
\begin{table*}[t]
    \centering
    \small
    \begin{tabular}{p{\linewidth}}
        \toprule
        \vspace{-2mm}
\textbf{First Stage:}\\
To enhance clarity and eliminate ambiguities in solving the problem of who stole the chess pieces, I will describe the use of logical operators and a markdown table. These tools can help structure the information given in the narrative and allow for a systematic analysis of the facts to deduce the culprit.\\

Logical Operators:\\
Logical operators such as AND, OR, and NOT can be used to combine different conditions that must be met for a suspect to be considered the thief. For example, if a suspect had the opportunity to steal the pieces (Opportunity = True) AND had a motive to stop the chess boxing fights (Motive = True), then they could be considered a potential thief (Suspect = True).\\

Markdown Table:\\
A markdown table can be used to organize information about each character in a structured way. Each row represents a character, and each column represents a key piece of information relevant to the mystery (e.g., Opportunity, Motive, Alibi, Actions). By filling in the table with information from the narrative, patterns may emerge that point to the likely thief.\\

Here's an example of how the markdown table might be structured:
\begin{verbatim}
| Character | Opportunity | Motive | Alibi | Actions | Suspect |
|-----------|-------------|--------|-------|---------|---------|
| Ernie     | ?           | ?      | ?     | ?       | ?       |
| Gordon    | ?           | ?      | ?     | ?       | ?       |
| Jesse     | ?           | ?      | ?     | ?       | ?       |
| Mac       | ?           | ?      | ?     | ?       | ?       |\end{verbatim}

By filling in the table with details from the story, we can systematically evaluate each character's likelihood of being the thief. The markdown table provides a clear visual representation of the information, making it easier to compare suspects and identify inconsistencies or patterns that could lead to a solution.\\

Using these tools, the analysis can be conducted in a step-by-step manner, ensuring that each piece of evidence is considered and that the conclusion is based on a logical deduction rather than speculation or assumption.\\

\\
\textbf{Second Stage:}\\
Initial State Representation:\\

Let's organize the information we have about each character in the markdown table format:
\begin{verbatim}
| Character | Opportunity | Motive | Alibi | Actions | Suspect |
|-----------|-------------|--------|-------|---------|---------|
| Ernie     | Yes         | Yes    | No    | Studying chess openings | ? |
| Gordon    | Yes         | No     | Yes   | Joking, training Ernie | ? |
| Jesse     | Yes         | No     | No    | Studying chess openings | ? |
| Mac       | Yes         | No     | No    | Boxing, not talking | ? |\end{verbatim}

Opportunity: All characters were present at the event and could have had the opportunity to steal the pawns.\\
Motive: Ernie has a motive to avoid boxing Mac, but it's not clear if others have a motive.\\
Alibi: Gordon was watching the door, so he has an alibi.\\
Actions: Ernie and Jesse were studying chess openings, Mac was boxing, and Gordon was joking and training Ernie.\\
Suspect: We need to determine who is most likely to have stolen the pawns based on the information provided.\\

Step-by-Step Solution Process:\\

1. The referee was sure that Gordon did not enter the room with the chessboard through the door, which gives Gordon an alibi for the time the pawns were stolen.\\
2. Ernie was busy studying chess openings and preparing for his match, which would make it unlikely for him to risk stealing the pawns and jeopardizing his chance to compete.\\
3. Mac, the real boxer, was involved in the boxing match and had just finished a fight, making it unlikely for him to have the opportunity to steal the pawns without being noticed.\\
4. Jesse was the one who discovered the pawns were missing, which could either mean he is innocent and just reporting the fact, or he is using the discovery as a way to deflect suspicion from himself.\\

Considering the above points, Jesse had the opportunity to steal the pawns when he went to get the chessboard. He had no alibi since he was not being watched at that moment. The other characters either had an alibi or were engaged in activities that made it unlikely for them to steal the pawns.\\

Concluding with the Answer:\\

Answer: 2.Jesse\\
        \bottomrule
    \end{tabular}
    \caption{
       Examples of Minute Mysteries QA (GPT-4 Instance-Based)
    }
    \label{tab:appendix-mm-qa-sep-model-examples-4-instance-based}
\end{table*}
\begin{table*}[t]
    \centering
    \small
    \begin{tabular}{p{\linewidth}}
        \toprule
        \vspace{-2mm}
\textbf{First Stage:}\\
Unified Representation Form: JSON (JavaScript Object Notation)\\

Explanation:\\
JSON is a lightweight data-interchange format that is easy for humans to read and write, and easy for machines to parse and generate. It is based on a subset of the JavaScript Programming Language and is a text format that is completely language independent. JSON is ideal for representing the state or conditions of the problems described because it can structure data in a hierarchical manner, which is particularly useful for representing complex scenarios with multiple entities and attributes.\\

Using JSON to represent the problems enhances accuracy in the following ways:\\

- **Standardization**: JSON provides a consistent format for representing the various elements of each problem, such as characters, locations, items, and events. This standardization reduces ambiguity and ensures that each aspect of the problem is clearly defined.\\
- **Hierarchy and Nesting**: JSON's ability to nest objects within other objects allows for a clear representation of the relationships between different elements of the problem, such as the location hierarchy (house > room > closet) or the sequence of events.\\
- **Key-Value Pairs**: JSON's key-value pairs enable precise labeling of each piece of information, which helps in accurately identifying and retrieving data for analysis.\\
- **Flexibility**: JSON can represent a wide range of data types, including strings, numbers, arrays, and booleans, which is useful for capturing the diverse information presented in the problems.\\
- **Machine Readability**: JSON is easily parsed by most programming languages, which allows for automated processing and analysis of the problem data, potentially leading to more accurate and efficient problem-solving.\\

Here is an example of how one of the problems might be represented in JSON:
\begin{verbatim}
```json
{
  "problem": {
    "title": "Mystery of the Hidden Brother",
    "characters": [
      {"name": "Andy", "age": 5.5, "attributes": ["expressive", "knows Matthew's location"]},
      {"name": "Matthew", "age": 3, "attributes": ["hiding", "younger brother"]},
      ......
    ],
    "locations": [
      {"name": "Matthew's room", "attributes": ["no door on closet", "crib with books underneath"]},
      {"name": "Andy's room", "attributes": ["water spill at entrance"]},
      ......
    ],
    "events": [
      {"description": "Andy tells Dad that Matthew is hiding", "location": "home", "time": "evening"},
      {"description": "Andy is excited for Dad to find Matthew", "location": "den"},
      ......
    ],
    "hints": [
      {"description": "Grass stains on Andy's feet", "implication": "recently outside"},
      {"description": "Chocolate remnants around Andy's lips", "implication": "ate chocolate"},
      ......
    ],
    "question": "Where is Matthew?"
  }
}
```\end{verbatim}

This JSON representation captures the essential details of the problem, organizing them into categories such as characters, locations, events, and hints. Each category is an array of objects, where each object represents a specific element with its own set of attributes. This structured format allows for a clear and detailed representation of the problem, which can be easily translated into a solution algorithm.\\

\\
\textbf{Second Stage:}\\
1. Initial State Representation:
\begin{verbatim}
```json
{  ......  }
```\end{verbatim}

2. Step-by-Step Solution Process:\\

- The grandfather clock stopped at 10:10 p.m., which is shortly after Ed left the shop. It is unlikely that the thief broke in immediately after Ed left, as it would be too risky.\\
- ......

Given that the mantle clock was left on the edge of the counter by Ed and was found on the floor, it is likely that the intruder knocked it over, causing it to stop at 10:59 p.m. This suggests that the break-in occurred around this time.\\

3. Concluding with the Answer:\\

Answer: 1. The mantle clock (stopped at 10:59 p.m.)\\
        \bottomrule
    \end{tabular}
    \caption{
       Examples of Minute Mysteries QA (GPT-4 Task-Based)
    }
    \label{tab:appendix-mm-qa-sep-model-examples-4-task-based}
\end{table*}
\begin{table*}[t]
    \centering
    \small
    \begin{tabular}{p{\linewidth}}
        \toprule
        \vspace{-2mm}
\textbf{First Stage:}\\
To enhance the clarity and accuracy of the analysis, a structured and concise form of communication can be employed. One suitable format is a markdown table, which can be used to organize and present the relevant information in a clear and concise manner. The table can include columns for each of the five friends, their departure times, and any other relevant information. This format would allow for easy comparison and identification of the missing person.\\

For example, the table could be structured as follows:
\begin{verbatim}
| Friend | Departure Time | Additional Information |
|---|---|---|
| Billy Friend | 9:50 PM | Drove Susan Allen home |
| Diana Scott | 11:00 PM | Drove home in her father's Cadillac |
| Harrell Garner | 9:30 PM | Called parents to inform them he was on his way home |
| Susan Allen | 9:50 PM | Rode with Billy Friend to her dorm |
| Steve Collier | N/A | Host of the study session |\end{verbatim}

By organizing the information in this way, it becomes easier to identify the missing person. Since all of the friends are accounted for except for one, it is clear that the missing person is the one who left last, which is Diana Scott.\\

Using a structured and concise form of communication, such as a markdown table, can help to improve the accuracy of the analysis by providing a clear and organized way to present the relevant information. This can make it easier to identify patterns and inconsistencies, and to draw accurate conclusions.\\

\\
\textbf{Second Stage:}\\
**Initial State Representation:**\\

* Five friends (Billy, Diana, Harrell, Susan, and Steve) gathered at Steve's apartment on Thursday night, November 21, 1963, to study for upcoming exams.\\
* Harrell left first at 9:30 PM and drove home, which usually took about 20 minutes.\\
* Billy and Susan left next at 9:50 PM. Billy drove Susan to her dorm and then headed to his own home.\\
* Diana left last at 11:00 PM and drove home in her father's Cadillac.\\
* Steve stayed at his apartment and went to bed.\\

**Step-by-Step Solution Process:**\\

1. We know that one of the friends is missing, but we don't know who it is.\\
2. We can eliminate Harrell as a possibility because he arrived home safely.\\
3. We can also eliminate Billy and Susan because they were together and there is no indication that anything happened to them.\\
4. This leaves Diana as the only remaining friend who has not been accounted for.\\

**Concluding with the Answer:**\\

Answer: 1. Diana Scott\\
        \bottomrule
    \end{tabular}
    \caption{
       Examples of Minute Mysteries QA (Gemini Pro Instance-Based)
    }
    \label{tab:appendix-mm-qa-sep-model-examples-gemini-instance-based}
\end{table*}
\begin{table*}[t]
  \centering
  \small
  \begin{tabular}{p{\linewidth}}
      \toprule
      \vspace{-2mm}
\textbf{First Stage:}
\begin{verbatim}
```
[room_of_murder, murder_weapon, suspect, motive]
```\end{verbatim}

This representation form can enhance the accuracy of analysis by providing a structured and organized way to capture the key elements of each similar problem. By breaking down the problems into these four categories, it becomes easier to identify patterns and relationships between the different elements. This can help to identify potential suspects, motives, and murder weapons, as well as to eliminate unlikely possibilities. Additionally, this representation form can help to ensure that all of the relevant information is considered when analyzing the problems, which can lead to more accurate and reliable conclusions.\\

For example, consider the following two problems:\\

**Problem 1:**\\

* A man is found dead in his study, stabbed in the back.\\
* The study is locked from the inside.\\
* The only other person in the house is the man's wife, who claims she was asleep at the time of the murder.\\

**Problem 2:**\\

* A woman is found dead in her bedroom, strangled.\\
* The bedroom window is open, and there are footprints in the flower bed below.\\
* The only other person in the house is the woman's husband, who claims he was out of town at the time of the murder.\\

Using the representation form above, we can represent these two problems as follows:
\begin{verbatim}
```
Problem 1: [study, knife, wife, unknown]
Problem 2: [bedroom, hands, husband, unknown]
```\end{verbatim}

By comparing these two representations, we can see that there are some similarities between the two problems. For example, in both cases, the murder took place in a private room, and the only other person in the house is the victim's spouse. This suggests that the spouse may be a potential suspect in both cases.\\

Additionally, by considering the different elements of each problem, we can start to generate hypotheses about what may have happened. For example, in Problem 1, the fact that the study is locked from the inside suggests that the wife may have been involved in the murder. In Problem 2, the fact that the bedroom window is open and there are footprints in the flower bed below suggests that the husband may have entered the house through the window and killed the wife.\\

Of course, these are just hypotheses at this point, and more investigation would be needed to determine what actually happened in each case. However, the representation form above provides a useful starting point for analysis, and it can help to ensure that all of the relevant information is considered.\\

\\
\textbf{Second Stage:}\\
**Initial State Representation:**
\begin{verbatim}
```
[room_of_incident, object_involved, people_present, time_of_incident]
```\end{verbatim}

**Step 1:**\\

* **room\_of\_incident:** Music room\\
* **object\_involved:** Canary cage, canary named Rufus\\
* **people\_present:** Mr. Carstairs, Mrs. Carstairs (upstairs), maid Marjorie (an hour ago), children Jimmy and Lydia, neighbor's boy Alex Johnston\\
* **time\_of\_incident:** Approximately 20 minutes before Crusher's arrival\\

**Step 2:**\\

* The canary cage was knocked off its perch and fell to the floor, causing the door latch to come undone and the canary to escape.\\
* The window was open, suggesting that the cage may have been knocked off the perch by someone or something coming in through the window.\\

**Step 3:**\\

* The children were playing frisbee outside the house at the time of the incident.\\
* Alex admits to running to catch a frisbee and falling, causing the frisbee to go through the window.\\
* Jimmy also went into the house to retrieve the frisbee, but didn't notice anything about the birdcage.\\
* Lydia was drinking from her cup near the house and had a clear view of the house, but didn't see the canary escape.\\

**Step 4:**\\

* There are mud stains and grass strands near the French doors, suggesting that someone came in through the window with muddy shoes.\\
* The cat, Sarabelle, was also in the room at the time of the incident.\\

**Conclusion:**\\

Based on the evidence, it is most likely that Alex is responsible for the disappearance of the canary. He admits to running to catch a frisbee and falling, which could have caused the cage to be knocked off the perch. The mud stains and grass strands near the French doors also suggest that he came in through the window with muddy shoes.\\

**Answer: 0. Alex Johnston**\\
      \bottomrule
  \end{tabular}
  \caption{
     Examples of Minute Mysteries QA (Gemini Pro Task-Based)
  }
  \label{tab:appendix-mm-qa-sep-model-examples-gemini-task-based}
\end{table*}
\begin{table*}[t]
    \centering
    \small
    \begin{tabular}{p{\linewidth}}
        \toprule
        \vspace{-2mm}
\textbf{First Stage (generated by GPT-3.5):}\\
Problem Representation:\\
Let's represent the problem in the form of a table:
\begin{verbatim}
| Problem | Given | Unknown | Formula |
|---------|-------|---------|---------|
| 1       | Cost, Profit, Percentage | Selling Price | Selling Price = Cost + Profit |
| 2       | Number of steps | Number of ways | Fibonacci sequence |
| 3       | Vitamin E, Vitamin A | Grams of health food | (0.09 * x) + (0.08 * x) = 1 |
| 4       | Average microwave ovens sold | Average for 22 months | (Total ovens sold in 22 months) / 22 |
| 5       | Size of the cube, Thickness | Number of unpainted faces | (Size - 2)^3 |\end{verbatim}

Explanation:\\
By representing the problems in a unified table format, it becomes easier to identify the given information, the unknowns, and the relevant formulas or concepts needed to solve each problem. This allows for a more systematic and organized approach to problem-solving, enhancing the accuracy of the analysis. It also helps in identifying patterns and similarities between different problems, making it easier to apply the appropriate solution methods.\\

\\
\textbf{Second Stage (generated by GPT-4):}\\
Initial State Representation:
\begin{verbatim}
| Problem | Given | Unknown | Formula |
|---------|-------|---------|---------|
| Choosing Oranges | 10 oranges in total | Number of ways to choose 2 oranges | Combination formula: 
C(n, k) = n! / (k!(n - k)!) |\end{verbatim}

Step-by-Step Solution Process:\\

1. We are given 10 oranges and we need to find the number of ways to choose 2 oranges from these 10.\\
2. The order in which we choose the oranges does not matter, so we use the combination formula.\\
3. Using the combination formula C(n, k) = n! / (k!(n - k)!), where n is the total number of items, and k is the number of items to choose, we get:\\
\verb|   C(10, 2) = 10! / (2!(10 - 2)!)|\\
4. Simplifying the factorial expressions, we get:\\
\verb|   C(10, 2) = (10 × 9) / (2 × 1)|\\
5. Further simplifying, we get:\\
\verb|   C(10, 2) = 90 / 2|\\
6. Finally, we get:\\
\verb|   C(10, 2) = 45|\\

Concluding with the Answer:\\
Answer: A\\

        \bottomrule
    \end{tabular}
    \caption{
       Examples of AQuA (GPT-3.5 $\Rightarrow$ GPT-4 Task-Based)
    }
    \label{tab:appendix-aqua-sep-model-examples-3.5-4}
\end{table*}
\begin{table*}[t]
    \centering
    \small
    \begin{tabular}{p{\linewidth}}
        \toprule
        \vspace{-2mm}
\textbf{First Stage:}\\
Initial State Representation:\\
Let's represent the vertices of the triangle as follows:\\
Vertex 1: (a, 0)\\
Vertex 2: (b, 0)\\
Vertex 3: (x, y)\\

We are looking for the area of the triangle formed by these vertices.\\

\\
\textbf{Second Stage:}\\
Step-by-Step Solution Process:\\
The area of a triangle with vertices (x1, y1), (x2, y2), and (x3, y3) can be calculated using the formula:\\
Area = |(x1(y2 - y3) + x2(y3 - y1) + x3(y1 - y2))/2|\\

Given the vertices of the triangle:\\
Vertex 1: (a, 0)\\
Vertex 2: (b, 0)\\
Vertex 3: (x, y)\\

Using the formula, the area of the triangle is:\\
Area = |(a(0 - y) + b(y - 0) + x(0 - 0))/2|\\
Area = |(by - ay)/2|\\

Therefore, the area of the triangle is represented by:\\
C) (by - ay)/2\\

Answer: C\\

        \bottomrule
    \end{tabular}
    \caption{
       Examples of AQuA (GPT-3.5 Instance-Based)
    }
    \label{tab:appendix-aqua-sep-model-examples-3.5-instance-based}
\end{table*}
\begin{table*}[t]
    \centering
    \small
    \begin{tabular}{p{\linewidth}}
        \toprule
        \vspace{-2mm}
\textbf{First Stage:}\\
1. Let's represent the similar problems using a table:
\begin{verbatim}
| Problem | Variables/Conditions |
|---------|----------------------|
| 1       | Size of cube = 5cm, thickness = 1cm, size of smaller cubes = 1cm, outer surface painted |
| 2       | Q is younger than R, R+T=50 |
| 3       | Coin size = 2x15mm, weight = 30g, volume of aluminum = volume of silver, silver is twice as heavy 
as aluminum |
| 4       | 5 vowels, 20 consonants, middle letter must be vowel, first and third letter must be different 
and consonants |
| 5       | Discount of 5% and 7%, difference in profit = Rs. 15 |\end{verbatim}

This representation form can be used to solve each problem by plugging in the specific values and conditions.\\

\\
\textbf{Second Stage:}\\
Initial State Representation:\\
The given equation is (5 + ? × 19 - 15 - 7) / (13 × 13 - 156) = 6\\

Step-by-Step Solution Process:\\
1. First, simplify the expression inside the brackets:\\
\verb|   (5 + ? × 19 - 15 - 7) = (5 + 19? - 22)|\\

2. Next, simplify the denominator:\\
\verb|   (13 × 13 - 156) = (169 - 156) = 13|\\

3. Substitute the simplified expressions back into the original equation:\\
\verb|   (5 + 19? - 22) / 13 = 6|\\

4. Multiply both sides by 13 to clear the fraction:\\
\verb|   5 + 19? - 22 = 78|\\

5. Combine like terms:\\
\verb|   19? - 17 = 78|\\

6. Add 17 to both sides:\\
\verb|   19? = 95|\\

7. Divide both sides by 19:\\
\verb|   ? = 5|\\

Concluding with the Answer:\\
Answer: C) 5\\

        \bottomrule
    \end{tabular}
    \caption{
       Examples of AQuA (GPT-3.5 Task-Based)
    }
    \label{tab:appendix-aqua-sep-model-examples-3.5-task-based}
\end{table*}
\begin{table*}[t]
   \centering
   \small
   \begin{tabular}{p{\linewidth}}
       \toprule
       \vspace{-2mm}
\textbf{First Stage (generated by GPT-4):}\\
JSON (JavaScript Object Notation) is a lightweight data-interchange format that is easy for humans to read and write, and easy for machines to parse and generate. It can be used to represent the state or conditions of the problems provided. Below is a JSON structure that could represent each of the problems:
\begin{verbatim}
```json
{
  "problem": {
    "type": "",
    "data": {
      "values": [],
      "units": "",
      "additional_info": {}
    },
    "choices": {
      "A": "",
      "B": "",
      "C": "",
      "D": "",
      "E": ""
    }
  }
}
```\end{verbatim}

This JSON structure includes a `problem` object that contains:\\

- `type`: A string describing the type of problem (e.g., "speed\_time\_distance", "probability", "geometry", "relative\_motion", "number\_theory").\\
- `data`: An object that includes:\\
\ \ - `values`: An array of numerical values relevant to the problem.\\
\ \ - `units`: A string indicating the units of measurement used in the problem.\\
\ \ - `additional\_info`: An object that can hold any additional information specific to the problem type (e.g., speed of the ship, width of pickets, etc.).\\
- `choices`: An object that lists the possible answer choices, with each choice represented by a key-value pair where the key is the choice label (A, B, C, D, E) and the value is the corresponding answer in string format.\\

This JSON structure can be adapted to represent each of the problems provided by simply filling in the appropriate values for each key.

\\
\textbf{Second Stage (generated by GPT-3.5):}
\begin{verbatim}
```json
{
  "problem": {
    "type": "quadratic_equation",
    "data": {
      "values": [-5, "k"],
      "units": "",
      "additional_info": {}
    },
    "choices": {
      "A": "9",
      "B": "7",
      "C": "3",
      "D": "8",
      "E": "12"
    }
  }
}
```\end{verbatim}

To find the largest integral value of 'k' for which the quadratic equation x$^2$ - 5x + k = 0 will have two real and distinct roots, we can use the discriminant formula. The discriminant (D) of the quadratic equation ax$^2$ + bx + c = 0 is given by D = b$^2$ - 4ac. For the equation x$^2$ - 5x + k = 0, the discriminant must be greater than 0 for the equation to have two real and distinct roots.\\

The discriminant in this case is (-5)$^2$ - 4*1*k = 25 - 4k. For the equation to have two real and distinct roots, 25 - 4k must be greater than 0. Solving for k, we get k < 25/4.\\

The largest integral value of k that satisfies this condition is 6. Therefore, the largest integral value of k for which the quadratic equation x$^2$ - 5x + k = 0 will have two real and distinct roots is 6.\\

Answer: D\\
       \bottomrule
   \end{tabular}
   \caption{
      Examples of AQuA (GPT-4 $\Rightarrow$ GPT-3.5 Task-Based)
   }
   \label{tab:appendix-aqua-sep-model-examples-4-3.5}
\end{table*}
\begin{table*}[t]
    \centering
    \small
    \begin{tabular}{p{\linewidth}}
        \toprule
        \vspace{-2mm}
\textbf{First Stage (generated by GPT-4):}\\
Unified Representation Form: Algebraic Equations\\

Explanation:\\
Algebraic equations are a versatile and powerful tool for representing a wide variety of problems, including the ones listed above. By using variables to represent unknown quantities and forming equations based on the given conditions, we can create a system that captures the essence of each problem. This approach allows us to translate word problems into mathematical language, which can then be solved using algebraic methods.\\

The algebraic representation enhances the accuracy of analysis by providing a clear, unambiguous framework for the problem. It eliminates the potential for misinterpretation of the problem's conditions and ensures that all relationships between quantities are explicitly stated. Additionally, algebra allows for the systematic application of mathematical rules and operations to find the solution, reducing the likelihood of errors that might arise from ad-hoc or intuitive reasoning.\\

For example, the first problem can be represented by the following system of equations, where A represents Arjun's money and S represents Sajal's money:
\begin{verbatim}
1. S + 30 = 2(A - 30)
2. A + 10 = 3(S - 10)\end{verbatim}

Each of the other problems can similarly be translated into algebraic equations that reflect the given conditions. This unified representation form is capable of expressing the relationships and constraints of each problem, making it a suitable choice for a wide range of similar problems.\\

\\
\textbf{Second Stage (generated by Gemini Pro):}\\
**Initial State Representation:**\\

Let x be the total number of people who took the survey.\\

**Step 1:**\\

60\% of the people who took the survey were women. So, the number of women is 0.6x.\\

**Step 2:**\\

Of the men who were surveyed, 75\% preferred red cars over green cars. So, the number of men who preferred red cars is 0.75 * (x - 0.6x) = 0.25x.\\

**Step 3:**\\

10 men liked green cars more than red. So, the number of men who preferred green cars is 0.25x - 10.\\

**Step 4:**\\

The total number of people who took the survey is the sum of the number of women and the number of men. So, x = 0.6x + (0.25x - 10).\\

**Step 5:**\\

Solving for x, we get x = 80.\\

**Answer: E**\\

        \bottomrule
    \end{tabular}
    \caption{
       Examples of AQuA (GPT-4 $\Rightarrow$ Gemini Pro Task-Based)
    }
    \label{tab:appendix-aqua-sep-model-examples-4-gemini}
\end{table*}
\begin{table*}[t]
    \centering
    \small
    \begin{tabular}{p{\linewidth}}
        \toprule
        \vspace{-2mm}
\textbf{First Stage:}\\
Initial State Representation:\\

Let's use a math equation to represent the initial state of the problem.\\

Let x be the length of the longer piece of rope in meters. According to the problem, the shorter piece is 3 meters shorter than the longer piece, so its length would be x - 3 meters.\\

The total length of the rope is 20 meters. Therefore, the sum of the lengths of the two pieces must equal 20 meters.\\

The equation representing the relationship between the lengths of the two pieces of rope is:\\

x + (x - 3) = 20\\

This equation can be used to solve for the value of x, which represents the length of the longer piece of rope.\\

\\
\textbf{Second Stage:}\\
Initial State Representation:\\
We have the equation x + (x - 3) = 20, where x is the length of the longer piece of rope.\\

Step-by-Step Solution Process:\\
1. Combine like terms on the left side of the equation:\\
\verb|   2x - 3 = 20|\\

2. Add 3 to both sides to isolate the term with the variable x:\\
\verb|   2x - 3 + 3 = 20 + 3|\\
\verb|   2x = 23|\\

3. Divide both sides by 2 to solve for x:\\
\verb|   2x / 2 = 23 / 2|\\
\verb|   x = 11.5|\\

Concluding with the Answer:\\
The length of the longer piece of rope is 11.5 meters.\\

Answer: D\\
        \bottomrule
    \end{tabular}
    \caption{
       Examples of AQuA (GPT-4 Instance-Based)
    }
    \label{tab:appendix-aqua-sep-model-examples-4-instance-based}
\end{table*}
\begin{table*}[t]
    \centering
    \small
    \begin{tabular}{p{\linewidth}}
        \toprule
        \vspace{-2mm}
\textbf{First Stage:}\\
Representation Form: Pseudocode\\

Explanation:\\
Pseudocode is a high-level description of an algorithm that uses the structural conventions of programming languages but is intended for human reading rather than machine reading. It allows us to represent the logic of a problem without getting bogged down in the syntax of a particular programming language. By using pseudocode, we can create a generic template that can be adapted to solve a variety of similar problems. This enhances accuracy by ensuring that the logical steps are clearly outlined and can be followed systematically, reducing the risk of errors that might arise from misinterpreting the problem or from syntax issues in actual code.\\

Pseudocode Template:
\begin{verbatim}
```
INITIALIZE variables and ratios as needed

READ input values

CALCULATE based on given ratios or probabilities

IF there are conditions or multiple scenarios THEN
    HANDLE each scenario separately
    CALCULATE probabilities or outcomes for each scenario
    COMBINE results as per the problem's requirements
END IF

CONVERT units if necessary

OUTPUT the final result
```\end{verbatim}

This pseudocode template can be adapted to represent the logic of each problem presented. It starts with initializing variables and ratios, reads input values, performs calculations based on the given information, handles conditional logic if present, converts units if required, and finally outputs the result. By following this structured approach, we can ensure that each step of the problem-solving process is accounted for, which helps in maintaining accuracy across different problems.\\

\\
\textbf{Second Stage:}\\
Initial State Representation:
\begin{verbatim}
```
tank_length = 8 // in meters
tank_width = 6 // in meters
tank_height = 2.5 // in meters
capacity_in_liters = 0 // to be calculated
```\end{verbatim}

Step-by-Step Solution Process:\\
\begin{verbatim}
```
// Step 1: Calculate the volume of the tank in cubic meters
volume_in_cubic_meters = tank_length * tank_width * tank_height

// Step 2: Convert the volume from cubic meters to liters
// 1 cubic meter = 1000 liters
capacity_in_liters = volume_in_cubic_meters * 1000

// Step 3: Substitute the values and calculate the capacity
capacity_in_liters = 8 * 6 * 2.5 * 1000
capacity_in_liters = 120 * 1000
capacity_in_liters = 120000
```\end{verbatim}

Concluding with the Answer:\\
Answer: D\\
        \bottomrule
    \end{tabular}
    \caption{
       Examples of AQuA (GPT-4 Task-Based)
    }
    \label{tab:appendix-aqua-sep-model-examples-4-task-based}
\end{table*}
\begin{table*}[t]
    \centering
    \small
    \begin{tabular}{p{\linewidth}}
        \toprule
        \vspace{-2mm}
\textbf{First Stage:}\\
1. Initial State Representation:\\

Let x be the length of the longer piece of rope.\\

Length of the shorter piece of rope = x - 3\\

Total length of the rope = 20 meters\\

Equation: x + (x - 3) = 20\\

\\
\textbf{Second Stage:}\\
1. Initial State Representation:\\

Let x be the length of the longer piece of wire.\\

Length of the shorter piece of wire = x - 3\\

Total length of the wire = 20\\

Equation: x + (x - 3) = 20\\

2. Step-by-Step Solution Process:\\

- Combine like terms: 2x - 3 = 20\\

- Add 3 to both sides: 2x = 23\\

- Divide both sides by 2: x = 11.5\\

3. Concluding with the Answer:\\

Answer: D) 11.5\\

        \bottomrule
    \end{tabular}
    \caption{
       Examples of AQuA (Gemini Pro Instance-Based)
    }
    \label{tab:appendix-aqua-sep-model-examples-gemini-instance-based}
\end{table*}
\begin{table*}[t]
    \centering
    \small
    \begin{tabular}{p{\linewidth}}
        \toprule
        \vspace{-2mm}
\textbf{First Stage:}\\
1. **Representation Form:**
\begin{verbatim}
   We can represent each problem as a tuple containing the following fields:

   - **Problem Type:** A category that classifies the problem, such as "Percentage Gain," "Discount and 
   Price Increase," "Speed and Distance," "Alloy Ratio," or "Purchase Ratio."
   - **Given Values:** A dictionary containing the values provided in the problem statement, such as the 
   initial price, selling price, discount percentage, speed, time, alloy ratios, or purchase quantities.
   - **Unknown Values:** A list of the values that need to be calculated, such as the percentage gain, 
   new price, length, alloy ratio, or purchase ratio.

   For example, the first problem can be represented as:

   ```
   (
       "Problem Type": "Percentage Gain",
       "Given Values": {
           "Initial Price": 450,
           "Selling Price": 7.50,
           "Quantity": 90
       },
       "Unknown Values": ["Percentage Gain"]
   )
   ```\end{verbatim}

2. **Accuracy Enhancement:**
\begin{verbatim}
   Using this structured representation form enhances the accuracy of analysis by:

   - **Standardization:** It provides a consistent and standardized way to represent problems, making it 
   easier to compare and analyze them.
   - **Clarity:** It eliminates ambiguities and ensures that all relevant information is captured, 
   reducing the chances of misinterpretation.
   - **Automation:** This representation form can be easily processed by computer programs, enabling 
   automated analysis and solution generation.

3. **Representation Form Generality:**

   This representation form can represent every single similar problem because it captures the essential 
   elements common to all problems of a particular type. For example, all percentage gain problems 
   involve calculating the percentage difference between an initial value and a final value. By 
   representing these problems in a structured manner, we can develop generic algorithms or techniques 
   to solve them.\end{verbatim}

\\
\textbf{Second Stage:}\\
**Initial State Representation:**
\begin{verbatim}
```
(
    "Problem Type": "Discount and Price Increase",
    "Given Values": {
        "Discount 1": 5,
        "Discount 2": 7,
        "Profit Difference": 15
    },
    "Unknown Values": ["Marked Price"]
)
```\end{verbatim}

**Step-by-Step Solution Process:**\\

1. Let the marked price of the watch be x.\\

2. Selling price with a 5\% discount = x - (5/100) * x = 0.95x\\

3. Selling price with a 7\% discount = x - (7/100) * x = 0.93x\\

4. According to the problem, the profit difference between the two discounts is Rs. 15. So, we can set up the equation:\\
\verb|   0.95x - 0.93x = 15|\\

5. Solving for x:\\

\verb|   0.02x = 15|\\
\verb|   x = 15 / 0.02|\\
\verb|   x = 750|\\

**Answer: D**\\
      \bottomrule
  \end{tabular}
  \caption{
     Examples of AQuA (Gemini Pro Task-Based)
  }
  \label{tab:appendix-aqua-sep-model-examples-gemini-task-based}
\end{table*}
\clearpage
\begin{table*}[t]
    \centering
    \small
    \begin{tabular}{p{\linewidth}}
        \toprule
        \vspace{-2mm}
        \textbf{Elimy:}\\
        Context:\\
        - Clive Anderson is a British television and radio presenter, comedy writer, and former barrister. He is best known for being the host of the improvisation show "Whose Line Is It Anyway?".\\
        \textbf{Foina:}\\
        **Question**:\\
        Where was Clive Anderson born?\\
        
        **Shared Answer**:\\
        <A>London</A>\\
        \bottomrule
    \end{tabular}
    \caption{
       Examples of Wiki Hop (GPT-3.5)
    }
    \label{tab:appendix-multi-wiki-hop-qa-3.5}
\end{table*}
\begin{table*}[t]
    \centering
    \small
    \begin{tabular}{p{\linewidth}}
        \toprule
        \vspace{-2mm}
        \textbf{Elimy:}\\
       Hey Fiona, I have the question and the options. Do you have any relevant information that can help us answer the question? \\
    Relevant Information:\\
    - Yerim Seowon is a former seowon located in Miryang, Gyeongsangnam-do, South Korea.\\
    - It was built in 1567 to enshrine a native Confucian scholar of Miryang.\\
    - It was designated a tangible cultural treasure of Gyeongsangnam-do in 1974.\\
    
    Thoughts:\\
    - The information clearly states that Yerim Seowon is in Gyeongsangnam-do, which is a province in South Korea.\\
    - The question asks for the country where Yerim Seowon is located.\\
    
    Based on the evidence provided, the answer should be "South Korea".\\
    
    Answer: south korea\\
        \textbf{Foina:}\\
      Great, thanks for the information. Based on the evidence you provided, the answer is indeed "south korea".\\
        \bottomrule
    \end{tabular}
    \caption{
       Examples of Wiki Hop (GPT-3.5, GPT-4)
    }
    \label{tab:appendix-multi-wiki-hop-qa-3.5-4}
\end{table*}
\begin{table*}[t]
    \centering
    \small
    \begin{tabular}{p{\linewidth}}
        \toprule
        \vspace{-2mm}
        \textbf{Elimy:}\\
Evidence:\\
- Gymnophiona includes legless caecilians and all amphibians more closely related to them than to frogs or salamanders.\\
- Caecilians are a group of limbless, serpentine amphibians.\\
- All modern caecilians and their closest fossil relatives are grouped as a clade, Apoda, within the larger group Gymnophiona.  \\
- Eocaecilia is a caecilian.\\

My thoughts:\\
- Since Eocaecilia is a caecilian, and all modern caecilians are grouped within the clade Apoda, which is part of the larger group Gymnophiona, the parent taxon of Eocaecilia is Gymnophiona.\\
Structured data:
\begin{verbatim}
{
  "Eocaecilia": {
    "classification": {
      "group": "caecilian",
      "clade": "Apoda",
      "larger_group": "Gymnophiona"
    }
  }
}\end{verbatim}

Conclusion:\\
Answer: gymnophiona\\
\textbf{Fiona:} \\
Great analysis, Emily. I agree with your conclusion. Based on the evidence provided, it's clear that the parent taxon of Eocaecilia is Gymnophiona. Well done!\\
        \bottomrule
    \end{tabular}
    \caption{
       Examples of Wiki Hop (GPT-4, GPT-3.5)
    }
    \label{tab:appendix-multi-wiki-hop-qa-4-3.5}
\end{table*}
\begin{table*}[t]
    \centering
    \small
    \begin{tabular}{p{\linewidth}}
        \toprule
        \vspace{-2mm}
        \textsc{\textbf{case-1:}}\\
        \textbf{Elimy:}\\
       Evidence:\\
- Hans Loritz joined the SS in September 1930.\\
- Hans Loritz joined the NSDAP in August 1930.\\
- NSDAP is the Nazi Party.\\

My thoughts:\\
Based on the evidence, Hans Loritz was a member of the Nazi Party (NSDAP).\\

Answer: \{nazi party\}\\
        \textbf{Foina:}\\
       Evidence:\\
- The context provided mentions the "Schutzstaffel" (SS), which was a major paramilitary organization under Adolf Hitler and the Nazi Party (NSDAP).\\
- Heinrich Himmler was a leading member of the Nazi Party and head of the SS.\\
- The context also references the Nazi Party (NSDAP) as being active between 1920 and 1945 and practicing Nazism.\\
My thoughts:\\
Given the information about the SS and its connection to the Nazi Party, as well as the direct mention of the Nazi Party (NSDAP), it is logical to conclude that Hans Loritz, associated with the SS, would be a member of the Nazi Party.\\
Answer: {nazi party}\\
\midrule
\textsc{\textbf{case-2:}}\\
\textbf{Elimy:}\\@Fiona: Requesting information on the film "The Eighth Day" to determine its genre. Please provide relevant data.\\
\textbf{Foina:}\\
@Emily: The film "The Eighth Day" is described as a Belgian comedy-drama. It tells the story of a friendship between a businessman and a man with Down syndrome. The film includes dream sequences with magical realism elements. It is not specified as a tragicomedy, but the blend of comedy and drama suggests it could fit that category as well.\\

Evidence:\\
- Described as a comedy-drama film.\\
- Contains elements of magical realism.\\
- Not explicitly labeled as tragicomedy, but the mix of genres implies a potential fit.\\

My thoughts: Based on the evidence, the genre of "The Eighth Day" aligns with both "comedy" and "drama," and potentially "tragicomedy" due to the blend of these elements.\\

Recommendation: We should consider "comedy," "drama," and "tragicomedy" as possible answers. Please cross-verify and confirm.   \\
\textbf{Elimy:}\\
@Fiona: Confirming that "The Eighth Day" is a comedy-drama, which aligns with the genre "tragicomedy" due to its blend of comedy and drama elements.\\
\textbf{Fiona:}
@Emily:Evidence:\\
- Comedy-drama is a subgenre of tragicomedy.\\
- "The Eighth Day" is described as a comedy-drama.\\

My thoughts: Given the evidence, "tragicomedy" is the most accurate genre that encompasses the elements of "The Eighth Day."\\    

Conclusion: Answer: tragicomedy\\
        \bottomrule
    \end{tabular}
    \caption{
       Examples of Wiki Hop (GPT-4)
    }
    \label{tab:appendix-multi-wiki-hop-qa-4}
\end{table*}
\begin{table*}[t]
    \centering
    \small
    \begin{tabular}{p{\linewidth}}
        \toprule
        \vspace{-2mm}
        \textbf{Elimy:}\\
       **Context**
       \begin{verbatim}
```json
{
  "Ruggero Deodato": {
    "nationality": "Italian",
    "occupation": ["film director", "screenwriter", "actor"]
  },
  "Alejandro Springall": {
    "nationality": "Mexican",
    "occupation": ["film director"]
  }
}
```\end{verbatim}

**Question**
       \begin{verbatim}
```json
{
  "question": "Are Ruggero Deodato from Italy, and Mexican Alejandro Springall, both film directors?"
}\end{verbatim}
        \textbf{Foina:}\\
       **Answer:**
\verb|```json|\\
\verb|{|\\
\verb|  "response": "Yes, Ruggero Deodato from Italy and Alejandro Springall from Mexico are both film directors."|\\
\verb|}|\\

\midrule
 \textbf{Elimy:}\\
        **Context:**\\
- Pollokshields Burgh Hall\\
- President of the Queen's Privy Council for Canada\\
- Accession Council\\
- Great Hall of the People\\
- St James's Palace\\

**Question:**\\
What building is opposite the ceremonial meeting place of the Accession Council in the United Kingdom?\\
        \textbf{Foina:}\\
       **Answer:**\\
\verb|```json|\\
\verb|{|\\
\verb|  "Building": "Great Hall of the People"|\\
\verb|}|\\
\verb|```|\\
$[...]$\\
        \bottomrule
    \end{tabular}
    \caption{
       Examples of Hotpot QA (GPT-3.5)
    }
    \label{tab:appendix-multi-hotpot-qa-3.5}
\end{table*}
\begin{table*}[t]
    \centering
    \small
    \begin{tabular}{p{\linewidth}}
        \toprule
        \vspace{-2mm}
              \textbf{Elimy:}\\
       Hey Fiona, I think we can solve this question by counting the number of bands each musician has been a member of. We can use a simple list of bands for each musician and then compare the lengths of the lists to find the answer.\\
Pete Doherty: ["The Libertines", "Babyshambles", "Peter Doherty and the Puta Madres"]\\
Jonny Craig: ["Dance Gavin Dance", "Emarosa", "Ghost Runner on Third", "Slaves", "westerHALTS", "Isles \& Glaciers"]\\
$[...]$\\
        \bottomrule
    \end{tabular}
     \caption{
       Examples of Hotpot QA (GPT-3.5, GPT-4)
    }
    \label{tab:appendix-multi-hotpot-qa-3.5-4}
\end{table*}
\begin{table*}[t]
    \centering
    \small
    \begin{tabular}{p{\linewidth}}
        \toprule
        \vspace{-2mm}
        \textbf{Elimy:}\\
      @Fiona: Requesting information on the actress who played "Celene" in "Walk All over Me" and her roles in "Starcraft 2" trilogy. \\
    \textbf{Foina:}\\
    @Emily: Tricia Helfer played "Celene" in "Walk All over Me" and voiced Sarah Kerrigan in the "Starcraft 2" trilogy.\\
    \textbf{Elimy:}
    Answer: Sarah Kerrigan\\
        \bottomrule
    \end{tabular}
    \caption{
       Examples of Hotpot QA (GPT-4, GPT-3.5)
    }
    \label{tab:appendix-multi-hotpot-qa-4-3.5}
\end{table*}
\begin{table*}[t]
    \centering
    \small
    \begin{tabular}{p{\linewidth}}
        \toprule
        \vspace{-2mm}
\textbf{Elimy:}\\
@Fiona: Requesting information on the designer of a 1917 Progressive Era community.\\
\textbf{Fiona:}\\
@Emily: Context indicates Howard Van Doren Shaw designed Deerpath Hill Estates in 1926. Not applicable for 1917 community.  \\
\textbf{Elimy:}\\
@Fiona: Context provided states Howard Van Doren Shaw designed Marktown in 1917. This matches the query for a 1917 Progressive Era community.\\
Answer: Marktown\\
\textbf{Fiona:}\\
@Emily: Confirming receipt of information on Marktown designed by Howard Van Doren Shaw in 1917.\\
Answer: Marktown\\
\midrule
        \textbf{Elimy:}\\
       Based on the provided context:\\

Jonny Craig has been a member of the following bands:\\
- Dance Gavin Dance\\
- Emarosa\\
- Ghost Runner on Third\\
- Slaves\\
- westerHALTS\\

Total bands: 5\\

Fiona, please provide the number of bands Pete Doherty has been a member of according to your context.\\
        \textbf{Foina:}\\
   Based on the provided context:\\

Pete Doherty has been a member of the following bands:\\
- The Libertines\\
- Babyshambles\\
- Peter Doherty and the Puta Madres\\

Total bands: 3\\

Answer: Jonny Craig\\
        \bottomrule
    \end{tabular}
    \caption{
       Examples of Hotpot QA (GPT-4)
    }
    \label{tab:appendix-multi-hotpot-qa-4}
\end{table*}






\begin{table*}[t]
    \centering
    \small
    \begin{tabular}{p{\linewidth}}
        \toprule
        \vspace{-2mm}
        \textbf{Elimy:}\\
     **Context:**\\
- The context is a passage from the story of Peter Pan in Kensington Gardens by J. M. Barrie. The passage describes the adventures of Maimie, a little girl, in the magical world of fairies and the little house she encounters.\\

**Question:**\\
- What is the crow's name?\\

**Your message to Fiona:**\\
- "The crow's name is not mentioned in the given context. We need to look for more information to find the answer."\\
        \textbf{Foina:}\\
**Your message to Emily:**\\
- "Confirmed. The provided context does not contain the crow's name. No further information available in my context. Suggest checking your context for the crow's name."\\

        \bottomrule
    \end{tabular}
    \caption{
       Examples of Narrative QA (GPT-3.5, GPT-4)
    }
    \label{tab:appendix-narrative-qa-3.5-4}
\end{table*}
\begin{table*}[t]
    \centering
    \small
    \begin{tabular}{p{\linewidth}}
        \toprule
        \vspace{-2mm}
        \textbf{Elimy:}\\
      I have the context that includes information about Mr. Jackson and his appearance in Mrs. Tittlemouse's house. I will now extract the relevant data to answer the question about which of Mr. Jackson's articles of clothing is dripping water.\\

Extracting relevant data:\\
- Mr. Jackson is described as having water dripping off his coat tails.\\
- He is sitting with his feet on the fender, implying that his coat is the source of the dripping water.\\

Conclusion:\\
- The article of Mr. Jackson's clothing that drips water is his coat.\\

Communicating conclusion to Fiona:\\
- Article of clothing: Coat\\
- Action: Dripping water\\

Answer: {coat}\\
        \textbf{Foina:}\\
Got it. Based on the information you provided, the article of Mr. Jackson's clothing that drips water is his coat.\\

Answer: {coat}\\
        \bottomrule
    \end{tabular}
    \caption{
       Examples of Narrative QA (GPT-4, GPT-3.5)
    }
    \label{tab:appendix-narrative-qa-4-3.5}
\end{table*}
\begin{table*}[t]
    \centering
    \small
    \begin{tabular}{p{\linewidth}}
        \toprule
        \vspace{-2mm}
        \textbf{Elimy:}\\
       @Fiona: Requesting information on contents used to fill the sack after bunnies' escape.\\
        \textbf{Foina:}\\
@Emily: The provided context does not contain the information required to answer the question about the contents of the sack. Please provide the relevant details from your context.\\
\textbf{Elimy:}\\
@Fiona: The contents used to fill the sack after the bunnies escape are "three rotten vegetable marrows, an old blacking-brush and two decayed turnips." Please confirm.\\
<A>three rotten vegetable marrows, an old blacking-brush and two decayed turnips</A>\\
        \bottomrule
    \end{tabular}
    \caption{
       Examples of Narrative QA (GPT-4)
    }
    \label{tab:appendix-multi-narrative-qa-4}
\end{table*}
\begin{table*}[]
    \centering
    \small
       \begin{tabular}{p{\linewidth}}
        \toprule
        \vspace{-2mm}
 1. Initial State Representation:\\
The initial state is that the coin is heads up.\\

2. Step-by-Step Solution Process:\\
- oshan flips the coin: The coin will change from heads to tails.\\
- quentin flips the coin: The coin will change from tails to heads.\\
- hayaat flips the coin: The coin will change from heads to tails.\\
- polewarzyk flips the coin: The coin will change from tails to heads.\\

3. Concluding with the Answer:\\
The coin is still heads up.\\
Answer: Yes\\
\bottomrule
    \end{tabular}
    \caption{Example response of Coin Flip (GPT-3.5)}
    \label{tab:single-coin-flip-3.5-case}
\end{table*}
\begin{table*}[]
    \centering
    \small
       \begin{tabular}{p{\linewidth}}
        \toprule
        \vspace{-2mm}
Initial State Representation: The coin is initially heads up.\\

Step-by-Step Solution Process:\\
1. Itzamar flips the coin. State changes; coin: tails up.\\
2. Clione flips the coin. State changes; coin: heads up.\\
3. Alely does not flip the coin. State remains unchanged; coin: heads up.\\
4. Sieghardt flips the coin. State changes; coin: tails up.\\

Answer: no\\
\bottomrule
    \end{tabular}
    \caption{Example response of Coin Flip (GPT-4)}
    \label{tab:single-coin-flip-4-case}
\end{table*}
\begin{table*}[]
    \centering
    \small
       \begin{tabular}{p{\linewidth}}
        \toprule
        \vspace{-2mm}
$[...]$\\
House 1:\\
- Cannot be the pianist\\
- Cannot be the person eating oranges\\
- Cannot have a television\\

House 2:\\
- Cannot be the person eating oranges\\
- Cannot have a television\\

House 3:\\
- Cannot be the person eating kiwis\\
- Cannot be the person wearing boots\\
$[...]$\\
\bottomrule
    \end{tabular}
    \caption{Example response of Logic Grid (GPT-3.5)}
    \label{tab:single-logic-grid-3.5-case}
\end{table*}
\begin{table*}[]
    \centering
    \small
       \begin{tabular}{p{\linewidth}}
        \toprule
        \vspace{-2mm}
$[...]$\\
Now we can fill in the details:\\
- House 1: Minivan, Daffodils, History Book, (not cat lover)\\
- House 2: Bird Keeper, Carnations, Romance Book\\
- House 3: SUV, (not cat lover), (not grapes)\\
- House 4: Convertible, Grapes\\
$[...]$\\
\bottomrule
    \end{tabular}
    \caption{Example response of Logic Grid (GPT-4) }
    \label{tab:single-logic-grid-4-case}
\end{table*}
\begin{table*}[]
    \centering
    \small
       \begin{tabular}{p{\linewidth}}
        \toprule
        \vspace{-2mm}
[Olivia]: Speaking:\\
Structured language for efficient communication.\\

Initial State Representation:\\
- Psychologist's tires slashed between 10:30 AM and 11:00 AM.\\
- Four "red-zone" patients are suspects: Mrs. Roberts, Randy Peters, Mr. Carlson, Mr. Leamington.\\

Step-by-Step Solution Process:\\
- Mrs. Roberts shows frustration but lacks a clear motive or physical capability for tire slashing.\\
- Randy Peters is visually impaired, but his ability to pick up pens suggests he could perform the act. His dream indicates subconscious aggression towards the psychologist.\\
- Mr. Carlson has a germ phobia, which would likely prevent him from slashing tires, but he expressed a desire to take out his frustration on someone.\\
- Mr. Leamington shows impulsive behavior and anger issues, but no direct link to the tire slashing incident.\\

Re-evaluation of Evidence:\\
- Mrs. Roberts and Mr. Leamington lack a clear motive or opportunity.\\
- Randy Peters' dream and subconscious aggression towards the psychologist, combined with his ability to navigate without clear vision, make him a plausible suspect.\\
- Mr. Carlson's expressed frustration and ability to overcome his phobia to sit in the chair suggest he could have slashed the tires.\\

Conclusion:\\
- Randy Peters and Mr. Carlson are the most likely suspects based on the evidence provided.\\
- Randy Peters' dream and actions during the session suggest a subconscious link to aggression towards the psychologist, making him the most likely suspect.\\
$[...]$\\
\bottomrule
    \end{tabular}
    \caption{Example response of Minute Mysteries QA (GPT-4)}
    \label{tab:single-minute-qa-4-case}
\end{table*}
\begin{table*}[]
    \centering
    \small
       \begin{tabular}{p{\linewidth}}
        \toprule
        \vspace{-2mm}
**Initial State Representation:**\\
- Socks: P, Q, R, S, T, U\\
- Pairs: 3 pairs in total\\
- P pairs with neither Q nor R\\

**Step-by-Step Solution Process:**\\
1. Let's start by analyzing statement 1: "Neither S nor T pairs with either Q or R."\\
   - This statement provides information about the pairs involving S and T, but it doesn't directly address the pairs involving P, Q, and R. Therefore, statement 1 alone is insufficient to determine the pairs.\\

2. Now, let's consider statement 2: "Neither P nor Q pairs with either S or U."\\
   - This statement provides information about the pairs involving P and Q, but it also doesn't directly address the pairs involving S, T, R, and U. Therefore, statement 2 alone is also insufficient to determine the pairs.\\
3. Next, let's analyze both statements together:\\
   - When we combine both statements, we have information about the pairs involving P, Q, R, S, T, and U. However, the information is still not sufficient to uniquely determine the pairs.\\

**Conclusion:**\\
Neither statement 1 nor statement 2 nor statements 1 and 2 taken together are sufficient.\\
\bottomrule
    \end{tabular}
    \caption{Example response of Information Essentiality (GPT-3.5)}
    \label{tab:single-info-essen-3.5-case}
\end{table*}
\begin{table*}[]
    \centering
    \small
       \begin{tabular}{p{\linewidth}}
        \toprule
        \vspace{-2mm}
Representation after Statement 1:
\begin{verbatim}
```
  ?       ?
H   E   ?   ?
  ?       ?
```\end{verbatim}
Statement 1 alone is insufficient.\\

Statement 2: F is sitting opposite to H.\\
This statement tells us the exact positions of F and H. Since there are only four players, we can deduce that E and G must be sitting to the left and right of F and H, respectively. However, we do not know which one is to the left and which one is to the right.\\

Representation after Statement 2:
\begin{verbatim}
```
  E/G     G/E
F   ?   ?   H
  ?       ?
```\end{verbatim}
Statement 2 alone is insufficient.\\

Now, let's combine both statements:\\

Statement 1 and 2 together:\\
From Statement 1, we know E is to the right of H. From Statement 2, we know F is opposite H. Therefore, G must be to the left of H, and E must be opposite G.\\

Final Representation:
\begin{verbatim}
```
  E       G
F   H   E   H
  G       F
```\end{verbatim}
Combining statements 1 and 2, we can determine that G is sitting opposite to E.\\

\midrule
Statement 1: "Q is to the north of R, who is to the west of P."\\
- If Q is north of R, we can represent this as Q ↑ R.\\
- If R is west of P, we can represent this as R ← P.\\
- Combining these two pieces of information, we can deduce that P is northeast of R and southeast of Q.\\

Representation:\\
\verb|```|\\
\verb|P|\\
↑ $\nwarrow$\\
\verb|R ← Q|\\
\verb|```|\\

From this representation, we can conclude that P is to the southeast of Q.\\
$[...]$\\
\bottomrule
    \end{tabular}
    \caption{Example responses of Information Essentiality (GPT-4)}
    \label{tab:single-info-essen-4-case}
\end{table*}



\end{document}